\documentclass{article}

\usepackage[preprint]{neurips_2026}
\usepackage{wrapfig}

\usepackage[utf8]{inputenc} 
\usepackage[T1]{fontenc}    
\usepackage{hyperref}       
\usepackage{url}            
\usepackage{booktabs}       
\usepackage{amsfonts}       
\usepackage{nicefrac}       
\usepackage{microtype}      
\usepackage{xcolor}         
\usepackage{graphicx}
\usepackage{subcaption} 
\usepackage{amsmath} 
\usepackage{xcolor}
\usepackage{multirow}
\newcommand{\std}[1]{{\scriptsize $\pm$ #1}}

\definecolor{xycolor}{HTML}{B85441}

\title{Breaking the Quality–Privacy Tradeoff in Tabular Data Generation via In-Context Learning}

\author{
Xinyan Han$^{1,*}$ \quad
Yan Lu$^{1,*}$ \quad
Xiaoyu Lin$^{1}$ \quad
Yuanyuan Jiang$^{1}$ \\
\textbf{Yuanrui Wang$^{1}$ \quad
Xuanyue Li$^{2}$ \quad
Wenchao Zou$^{3}$ \quad
Xingxuan Zhang$^{1,\dagger}$} \\
$^{1}$Tsinghua University \quad
$^{2}$Southeast University \quad
$^{3}$Siemens China \\
\texttt{\{han-xy25,luyan22,amy\_jyy,wang-yr23\}@mails.tsinghua.edu.cn} \\
\texttt{clementinexiaoyu@gmail.com} \quad \texttt{Lixy\_ai@163.com} \\
\texttt{wenchao.zou@siemens.com} \quad \texttt{xingxuanzhang@hotmail.com}
}

\begin{document}

\maketitle

\begin{abstract}
Tabular data synthesis aims to generate high-quality data while preserving privacy. However, we find that existing tabular generative models exhibit a clear tradeoff in the small-data regime: improving data quality typically comes at the cost of increased memorization of training samples, thereby weakening privacy protection. This tradeoff arises because small training sets make it difficult for dataset-specific generative models to distinguish generalizable structure from sample-specific patterns. 
To address this, we propose \textbf{DiffICL}, which formulates tabular data generation as an in-context learning problem. Instead of fitting each dataset from scratch, DiffICL leverages pretrained structural priors learned from a large collection of datasets, enabling it to infer data distributions from limited context rather than memorizing individual samples.
We evaluate DiffICL on 14 real-world datasets. Results show that DiffICL improves both data quality and privacy, and generate synthetic data that provides effective data augmentation. Our findings suggest that the quality–privacy tradeoff can be improved through better training paradigms.
\end{abstract}

\section{Introduction}
\begin{wrapfigure}{r}{0.37\textwidth}
\vspace{-1.2em}
    \centering
    \includegraphics[width=0.36\textwidth]{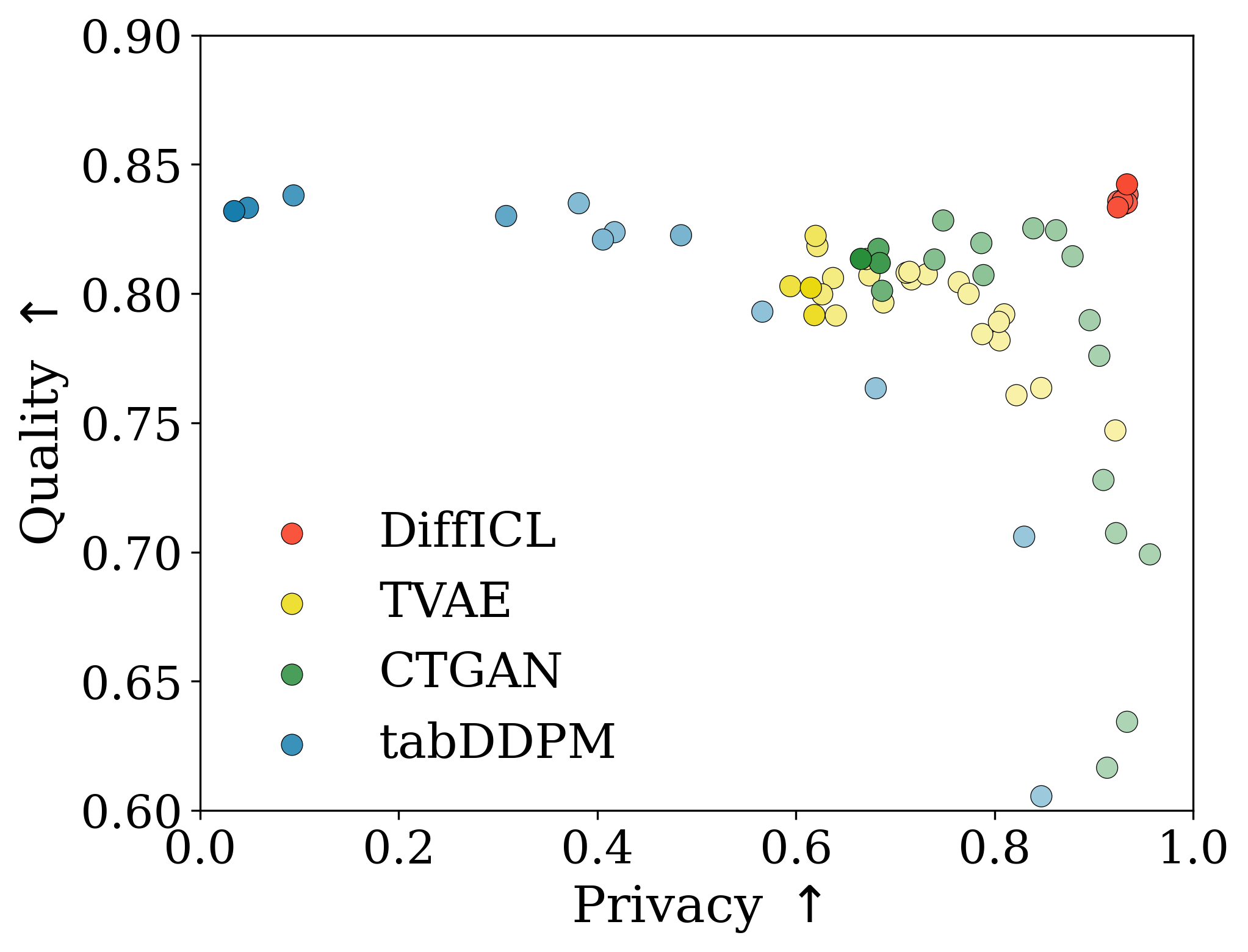}
    \vspace{-0.8em}
    \caption{Quality--privacy tradeoff frontiers.}
    \label{fig:pareto}
    \vspace{-1.4em}
\end{wrapfigure}
Tabular data is widely used across high-stakes domains such as healthcare, finance, and public administration, where data sharing is often hindered by privacy concerns and regulatory constraints~\cite{el2013anonymizing,de2013unique,oyewole2024data,de2015unique,narayanan2008robust}. This limitation creates a strong demand for privacy-preserving data sharing mechanisms, making tabular data synthesis a promising solution. Tabular data synthesis involves two key objectives: generating high-quality samples that faithfully capture the underlying data distribution to ensure strong downstream utility, and providing robust privacy guarantees to protect sensitive individual-level information~\cite{hernandez2025comprehensive,sarmin2025synthetic}. To achieve these goals, deep generative models have been widely adopted to model the complex and heterogeneous distributions of tabular data. Existing approaches can be broadly categorized into three main lines of work: GAN-based methods~\cite{xu2018synthesizing,park2018data,xu2019modeling,zhao2021ctab,zhao2024ctab}, VAE-based methods~\cite{xu2019modeling,liu2023goggle,apellaniz2024improved}, and the recently popular Diffusion-based methods~\cite{kim2022stasy,kotelnikov2023tabddpm,zhang2023mixed,lee2023codi,lin2024ctsyn,shi2024tabdiff}. 

Despite the recent progress in tabular data generation, achieving a favorable tradeoff between data quality and privacy remains particularly challenging in small-data regimes. To characterize this issue, we evaluate representative methods from different paradigms---tabDDPM~\cite{kotelnikov2023tabddpm}, TVAE~\cite{xu2019modeling}, and CTGAN~\cite{xu2019modeling}---on the Adult~\cite{kohavi1996scaling} dataset with  $N=200$ training samples. For each baseline, we vary the number of training steps to examine how generation quality and privacy protection change during optimization in Figure~\ref{fig:pareto}. Each point represents a model checkpoint obtained after a specific number of training steps. Darker points correspond to later checkpoints. Diffusion-based (\textcolor[HTML]{0A77A9}{blue}) models exhibit strong fitting capacity and can achieve high sample quality, yet this comes at the cost of severe overfitting: they generate samples that are overly close to the training records. This memorization can reveal the presence of individual training records or expose record-specific information, directly weakening privacy. In contrast, VAE-based (\textcolor[HTML]{EAD703}{yellow}) and GAN-based (\textcolor[HTML]{1D8830}{green}) methods introduce stronger architectural regularization that mitigates memorization, but their limited expressiveness yields substantially inferior generation quality. Under limited data, existing approaches fail to simultaneously achieve high fidelity and strong privacy protection, instead trading one for the other along a suboptimal Pareto front. 
This tradeoff arises because limited samples provide only a noisy and incomplete view of the underlying distribution. Due to the curse of dimensionality, small datasets provide only sparse coverage of the data space, making reliable inference of global distribution difficult.
\begin{figure}[t]
    \centering
    \begin{subfigure}{0.98\textwidth}
        \centering
        \includegraphics[width=\linewidth]{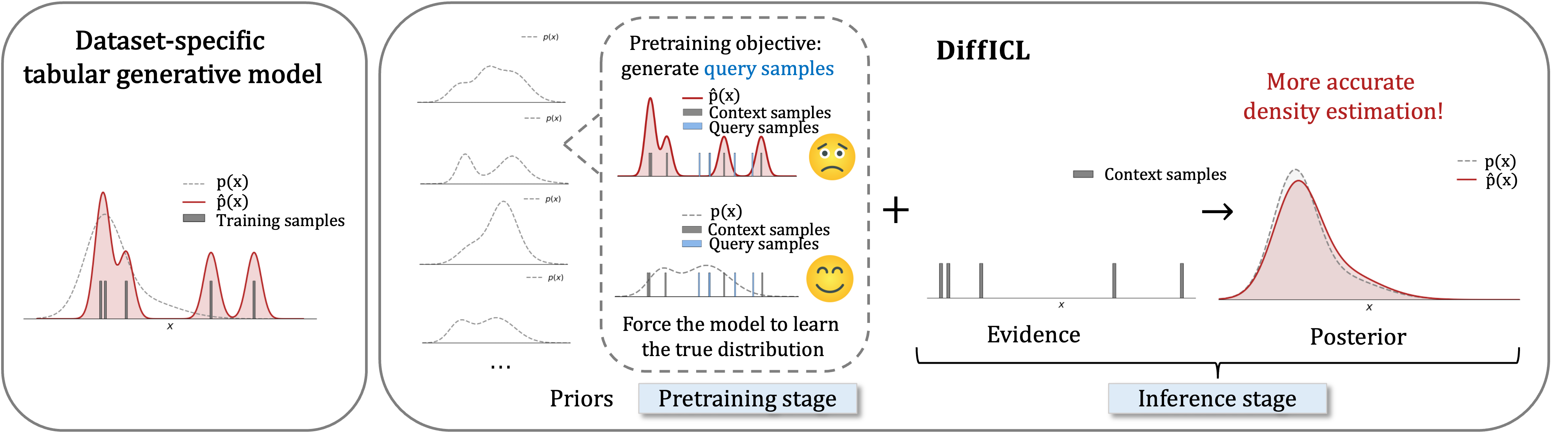}
    \end{subfigure}
    \caption{ICL pretraining enables more accurate density estimation from limited data by learning transferable structural priors. \small{\textbf{Left}: standard dataset-specific generative model overfits sparse observations and yields a poor estimate of the true distribution. \textbf{Right}: at the pretraining stage, the model learns to generate query samples $D_{\text{qry}}$ conditioned on context samples $D_{\text{ctx}}$, which requires inferring the underlying distribution instead of copying the context; the diverse distributions encountered during pretraining provide the priors. At the inference stage, the model treats the real dataset as context, infers the target distribution from $D_{\text{ctx}}$ and generates new synthetic samples.}}
    \vspace{-0.8em}
    \label{fig:background}
    \vspace{-0.8em}
\end{figure}

To address this limitation,  we turn to in-context learning (ICL) pretraining. In predictive tasks, tabular foundation models \cite{hollmann2025accurate,qu2025tabicl,ma2024tabdpt,garg2025real,zhang2025mitra,zhang2025limix} have demonstrated that ICL pretraining encodes rich structural priors into the network, enabling strong performance on small context sets without task-specific retraining. We argue that leveraging such pretrained priors is the key to solving the memorization problem of generative models in data scarce scenarios: Standard generative models optimize directly on the empirical data; with small sample sizes, deceptively low loss can be easily achieved through simple memorization. ICL pretraining removes the memorization incentive: each dataset is randomly partitioned into a context set $D_\text{ctx}$ and a disjoint query set $D_\text{qry}$, and the model is trained to generate $D_\text{qry}$ conditioned on $D_\text{ctx}$. Because $D_\text{ctx}$ and $D_\text{qry}$ are i.i.d., memorizing $D_\text{ctx}$ cannot yield a lower pretraining loss than genuinely learning the underlying true distribution. The model is thus driven to infer the true distribution rather than memorize the context, acquiring a prior for "distributions given a context." At inference, the scarce real dataset is fed as $D_\text{ctx}$, enabling generation that is faithful to the context distribution yet distinct from individual records, thereby improving privacy. We implement this idea in DiffICL. To handle the heterogeneity of tabular data—where tables differ in size, column semantics, and types—DiffICL projects raw features into a unified latent space via a tabular foundation model, then trains a dual-axis attention Transformer to perform conditional generation over these latent embeddings. As shown in Figure~\ref{fig:pareto} (\textcolor[HTML]{F84830}{red}), DiffICL establishes a superior Pareto frontier over existing paradigms. We summarize our main contributions as follows: 

\begin{enumerate}
    \vspace{-0.3em}
    \item To the best of our knowledge, we are the first to formulate tabular data generation as an in-context learning problem. We propose DiffICL, which avoids the memorization incentive that arises in standard generative training by optimizing an ICL pretraining objective (Section~\ref{sec:method}). 
    \vspace{-0.3em}
    \item  We evaluate DiffICL on 14 real-world datasets covering both classification and regression tasks. DiffICL achieves markedly better data quality and privacy protection than representative methods from each major generative paradigm on average (Section~\ref{sec:exp2}). 
    \vspace{-0.3em}
    \item DiffICL generates synthetic data that serves as effective data augmentation, improving downstream task performance when added to real training data (Section~\ref{sec:exp3}). Ablation results further show that this benefit depends on the structural priors acquired during pretraining, which guide the generation process at inference time and produce more informative augmentation samples (Section~\ref{sec:exp4}).
\end{enumerate}

\vspace{-0.3em}
\section{Quality–Privacy Tradeoff in Tabular Generative Models}
\vspace{-0.3em}
\label{sec:tradeoff}

\begin{figure}[t]
    \centering
    \begin{subfigure}{0.9\textwidth}
        \centering
        \includegraphics[width=\linewidth]{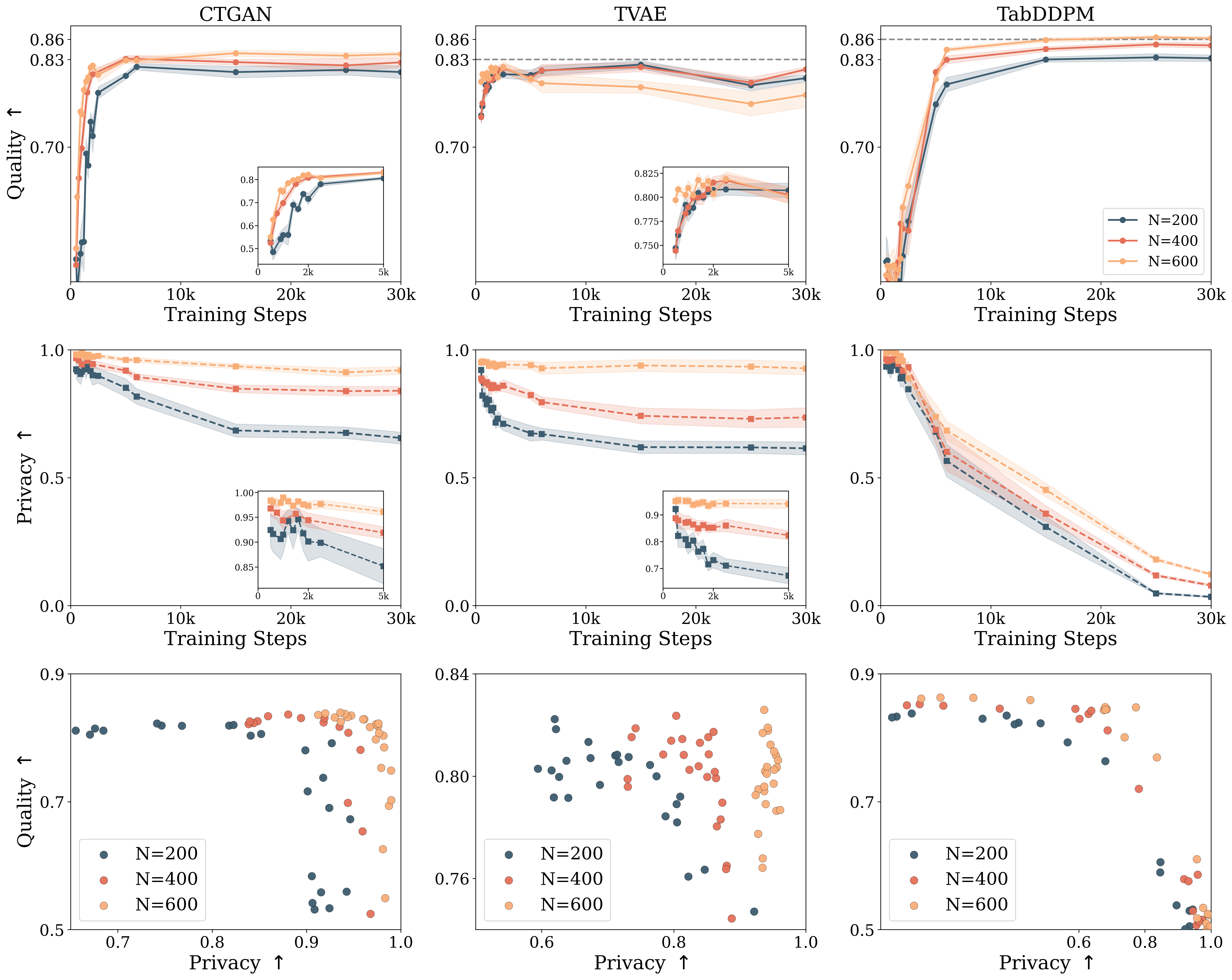}
    \end{subfigure}
    \vspace{-0.5em}
    \caption{Training dynamics and quality–privacy tradeoffs across dataset sizes ($N=200, 400, 600$) on the Adult dataset. \small{\textbf{Top} and \textbf{middle} rows show the evolution of generation quality (AUC) and privacy (DCROverfit) over training steps for CTGAN, TVAE, and tabDDPM. \textbf{Bottom} row plots the resulting quality–privacy frontiers, where each point corresponds to a training checkpoint.  }
}
\vspace{-0.8em}
    \label{fig:training_dynamic}
\end{figure}
In this section, we examine the tradeoff between generation quality and privacy risk on the Adult dataset with $N \in \{200, 400, 600\}$. We evaluate multiple generative models at different stages of training. Generation quality is measured by the AUC of an XGBoost classifier trained on synthetic data and evaluated on real test data. Privacy risk is quantified based on DCR, a widely used privacy metric for tabular synthetic data~\cite{zhao2021ctab,kotelnikov2023tabddpm,zhang2023mixed,shi2024tabdiff}. For a synthetic sample $x_{\text{syn}}$ and a reference set $D$, we denote its distance to the closest record as $\text{DCR}(x_{\text{syn}}, D)$. Let $p$ denote the proportion of synthetic samples whose DCR to the training set is smaller than DCR to the held-out validation set:
\begin{equation}
    p
    =
    \frac{1}{|D_{\text{syn}}|}
    \sum_{x_{\text{syn}} \in D_{\text{syn}}}
    \mathbb{I}
    \left[
    \text{DCR}(x_{\text{syn}}, D_{\text{train}})
    <
    \text{DCR}(x_{\text{syn}}, D_{\text{val}})
    \right].
\end{equation}
Under an ideal i.i.d. setting, $p$ should be close to $0.5$, since synthetic samples should not be systematically closer to training set than to held-out validation set. Values substantially above $0.5$ suggest overfitting of the training set. The DCROverfit score is then defined as
\begin{equation}
    \text{DCROverfit} = \min \left( 1,\; 2(1 - p) \right) \in [0, 1],
\end{equation}
where smaller values indicate greater privacy leakage risk.

As illustrated in first two rows of Figure~\ref{fig:training_dynamic}, for a fixed dataset size $N$, we observe a consistent training dynamic: quality improves steadily and eventually saturates, while privacy degrades over time. Following prior work~\cite{bonnaire2025diffusion}, this behavior can be interpreted through two characteristic timescales: $\tau_{\text{gen}}$, when useful generative capability is established, and $\tau_{\text{mem}}$, when memorization begins to emerge. Ideally, these two timescales are separated, creating a window in which models achieve high quality without memorization. However, this separation does not occur in the low-data regime ($N=200$ and $N=400$). Instead, memorization appears early in training, before sufficient generation quality is achieved, leaving no effective room for early stopping. 
This limitation is further reflected in the quality–privacy frontiers (last row of Figure~\ref{fig:training_dynamic}). Most trajectories exhibit a direct tradeoff: quality improves at the expense of privacy, preventing convergence to the desirable upper-right region. Ideally~\cite{bonnaire2025diffusion}, training would first exhibit a vertical phase (quality improvement with stable privacy), followed by a horizontal phase (privacy degradation after quality saturates). In practice, however, these two effects occur simultaneously. As the dataset size $N$ decreases, the frontiers contract inward, indicating a more severe quality–privacy tradeoff where both objectives cannot be simultaneously achieved.  
CTGAN and tabDDPM exhibit memorization early in training, suggesting that a dynamical regularization regime does not emerge in the small-data setting. At $N=600$, TVAE shows a nearly vertical trajectory, where quality improves without noticeable privacy degradation. However, this behavior does not imply the presence of dynamical regularization: if such a regime existed, it would be followed by a horizontal phase corresponding to delayed memorization. Instead, it reflects architectural regularization—TVAE lacks sufficient capacity to fit the empirical distribution and thus does not enter a memorization regime even with prolonged training. This benefit comes at a cost. While architectural regularization mitigates memorization, it also constrains model capacity, leading to lower peak quality compared to tabDDPM (e.g., 0.83 vs. 0.86 at $N=600$).  

Overall, our results show that existing generative models fail to achieve a favorable quality–privacy tradeoff in small-data tabular settings, where gains in quality are accompanied by increased memorization risk. We further find that this limitation cannot be mitigated through implicit training regularization, models must rely on architectural constraints to control memorization, which in turn restricts their achievable generation quality. This suggests that improving the quality–privacy tradeoff requires new mechanisms beyond those provided by current generative models.

\section{Methodology}
\label{sec:method}
\vspace{-0.8em}
\begin{figure}[t]
    \centering
    \begin{subfigure}{0.98\textwidth}
        \centering
        \includegraphics[width=\linewidth]{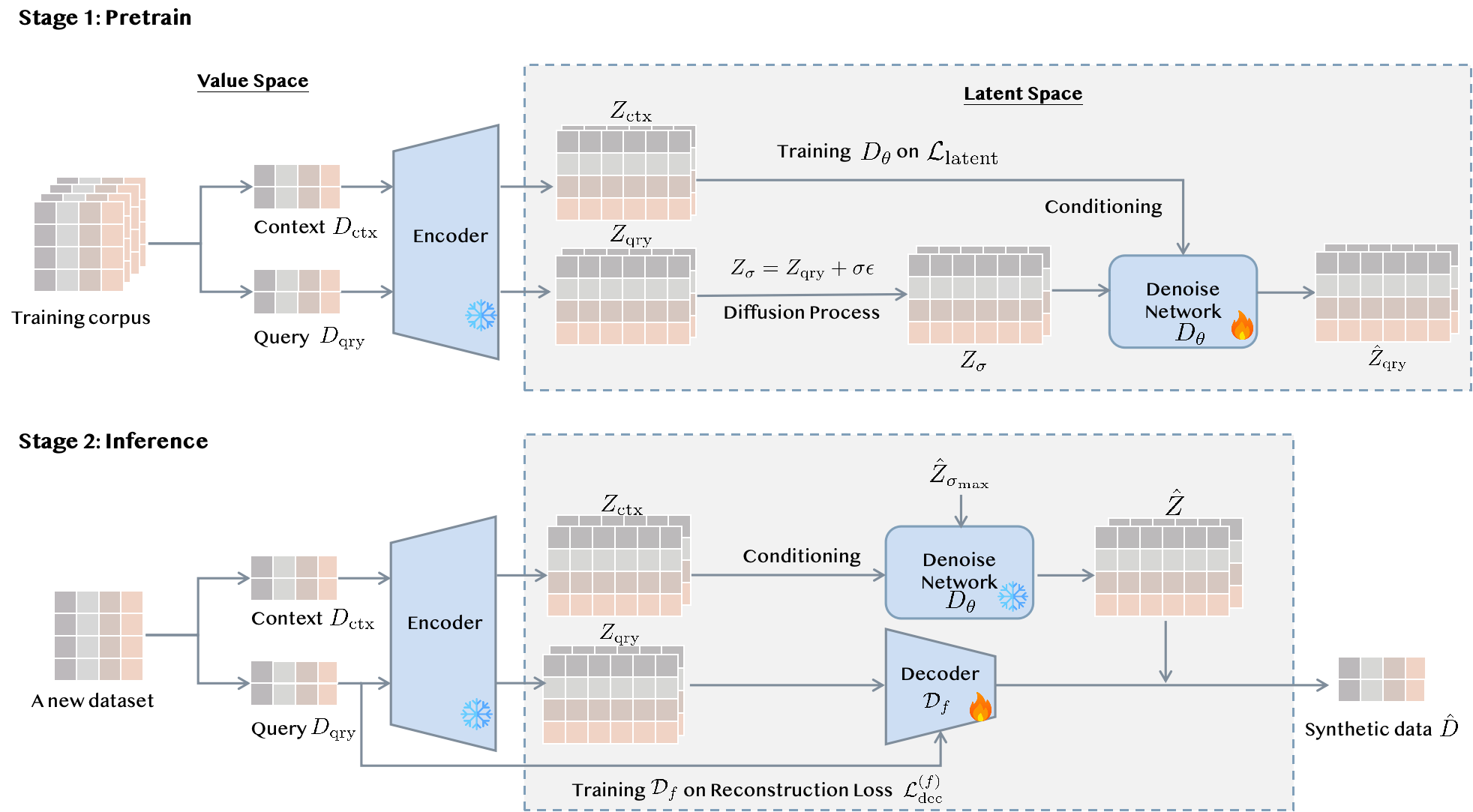}
    \end{subfigure}
    \caption{Illustration of DiffICL framework. \small{\textbf{Top}: At pretraining stage, tabular data are split into context and query sets and encoded into latent representations. A conditional diffusion model learns to denoise noisy query latents conditioning on context latents, capturing the dataset distribution in latent space. \textbf{Bottom}: At inference stage, the pretrained model generates synthetic latent samples from noise conditioned on the context set. These latents are then mapped back to raw features using feature-wise decoders trained on the query data.}}
    \vspace{-0.8em}
    \label{fig:framework}
\end{figure} 
\paragraph{Notations}To handle the heterogeneity of diverse tabular data, we map raw features to a continuous latent space with hidden dimension $d$. Throughout this section, uppercase letters denote sample sets: $D_{\text{ctx}}$ and $D_{\text{qry}}$ represent raw context and query sets, while $Z_{\text{ctx}}$ and $Z_{\text{qry}}$ represent their latent counterparts. $M_\text{ctx}$ and $M_\text{qry}$ denote the number of context and query samples, respectively. Synthetic sets are denoted with a hat (e.g., $\hat{D}$, $\hat{Z}$).
\vspace{-0.3em}
\paragraph{Method Overview}Our goal is to learn a conditional generative model that produces high-fidelity synthetic tabular samples $x$ based on a small set of context samples $D_{\text{ctx}}$ from a new dataset. Formally, we aim to approximate the conditional distribution $p_\theta(x|D_{\text{ctx}})$ for an arbitrary $D_{\text{ctx}}$. 
To prevent the model from trivially memorizing the training data, we formulate this through an ICL pretraining paradigm: For a given dataset $D \in \mathbb{R}^{N \times F}$ with $N$ samples and $F$ features, we randomly partition it into two disjoint subsets: a context set $D_{\text{ctx}} \in \mathbb{R}^{M_\text{ctx} \times F}$ and a query set $D_{\text{qry}} \in \mathbb{R}^{M_\text{qry} \times F}$, where $M_\text{ctx} + M_\text{qry} = N$. By fitting the conditional distribution of  $D_\text{qry}$ given $D_\text{ctx}$, i.e. $\hat{p}(x|D_{\text{ctx}}) = \frac{1}{M_\text{qry}} \sum_{x^{(i)} \in D_{\text{qry}}} \delta(x - x^{(i)})$, we force the model to infer the underlying true data distribution rather than copying the context. Building upon this ICL formulation, we propose a two-stage framework:
\textit{Stage 1: Pretraining.} We train a latent diffusion model on a large-scale corpus of over 800 real-world datasets. To promote permutation invariance and augment training diversity, for each dataset we randomly permute its rows and columns to obtain five structural variants, yielding a pretraining corpus five times the size of the original collection. The raw data $D_\text{ctx}$ and $D_\text{qry}$ are first projected into latent space through an encoder. The model then learns to generate the latent query set $Z_{\text{qry}}$ conditioned on the latent context set $Z_{\text{ctx}}$.
\textit{Stage 2: Inference.} For a new dataset, we use the pretrained latent diffusion model to sample a synthetic latent set $\hat{Z}_0$ conditioned on the provided context. These latents are then mapped back to the raw feature space via lightweight, dataset-specific decoders. We illustrate our method framework in Figure~\ref{fig:framework}.  
\vspace{-0.3em}
\subsection{Stage 1: Conditional Latent Diffusion Pretraining}
\vspace{-0.3em}
\paragraph{Latent Encoding.}
We adopt LimiX~\cite{zhang2025limix} as the shared encoder of context $D_\text{ctx}$ and query $D_\text{qry}$ because LimiX is among the strongest currently available tabular foundation models and supports missing-value imputation task, suggesting that its hidden representations preserve information about the input features rather than only label-predictive signals. This property is particularly important for DiffICL, since DiffICL requires to reconstruct raw feature values from the generated latent representations. The LimiX encoder is kept frozen throughout our framework. For any input $(D_{\text{ctx}}, D_{\text{qry}})$, we extract the final-layer hidden representations of LimiX to obtain the latent tensors:
\begin{equation}
    (Z_{\text{ctx}}, Z_{\text{qry}}) = \text{Encoder}(D_{\text{ctx}}, D_{\text{qry}})
\end{equation}
This operation projects the raw 2D matrices into 3D latent tensors, $Z_{\text{ctx}} \in \mathbb{R}^{M_\text{ctx} \times F \times d}$ and $Z_{\text{qry}} \in \mathbb{R}^{M_\text{qry} \times F \times d}$ (with $d=192$). \vspace{-0.8em}
\paragraph{In-Context Diffusion Process.}
We formulate generation as a continuous-time diffusion process~\cite{song2020score,karras2022elucidating} over the latent query set $Z_{\text{qry}}$. Given a clean latent set $Z_{\text{qry}}$, we construct a noisy latent set $Z_\sigma$ by adding independent Gaussian noise:
\begin{equation}
Z_\sigma = Z_{\text{qry}} + \sigma \mathcal{E}, \qquad
\mathcal{E} \in \mathbb{R}^{M_\text{qry} \times F \times d}
\text{ with } \mathcal{E}_{i,j,k} \overset{\text{i.i.d.}}{\sim} \mathcal{N}(0, 1),
\end{equation}
where the noise level $\sigma \in \mathbb{R}$ follows a log-normal distribution $\ln \sigma \sim \mathcal{N}(-1.2, 1.2^2)$. The denoising network $D_\theta$ is trained to reconstruct query latents $Z_{\text{qry}}$ from noisy latents $Z_\sigma$, conditioned on the context $Z_{\text{ctx}}$. The training objective is:
\begin{equation}
\mathcal{L}_{\text{latent}}(\theta) = \mathbb{E}_{{Z_{\text{qry}}, \sigma, \mathcal{E}}} \left[ \lambda(\sigma) \big( D_\theta(Z_\sigma; \sigma, Z_{\text{ctx}}) - Z_{\text{qry}} \big)^2 \right],
\end{equation}
where $\lambda(\sigma) = \frac{\sigma^2 + \sigma{\text{data}}^2}{(\sigma \sigma_{\text{data}})^2}$ and $\sigma_{\text{data}} = 0.5$. We provide more pretraining details and hyperparameter configuration in Appendix~\ref{sec:pretraining_details}.
\vspace{-0.8em}
\paragraph{Denoising Network Architecture.}
Since different datasets may contain different numbers of samples and features, the latent tensors produced by the frozen LimiX encoder have variable shapes:
$Z_{\mathrm{ctx}} \in \mathbb{R}^{M_\text{ctx} \times F \times d}$ and
$Z_{\mathrm{qry}} \in \mathbb{R}^{M_\text{qry} \times F \times d}$.
Therefore, the denoising network $D_\theta$ must operate on both variable sample size and variable feature dimension. Additionally, the denoising model should allow query latents to leverage the context for guidance while strictly preventing information exchange among individual query latents. To meet these requirements, we design the denoising network as a dual-axis attention Transformer. Within each layer, the model employs a decoupled, dual-axis attention mechanism: 
\textit{feature-wise attention} is applied over the feature dimension within each sample. This allows the model to capture dependencies among features and supports datasets with different numbers of features.
\textit{sample-wise attention} is applied over the sample dimension.
Context latents attend to each other, while each query latent attends only to the context latents. Query latents are not allowed to attend to other query latents, so the generated samples remain conditionally independent given the context.
Further architectural details are provided in Appendix~\ref{sec:denoise_network}.

\subsection{Stage 2: Inference and Feature Reconstruction}
At inference time, we are given a new dataset $D$, which may not have appeared during pretraining.
We split it into two disjoint subsets, $D_{\mathrm{ctx}}$ and $D_{\mathrm{qry}}$.
Following the same encoding procedure used in pretraining, we feed the pair
$(D_{\mathrm{ctx}}, D_{\mathrm{qry}})$ into the frozen LimiX encoder and obtain their latent representations 
$
    (Z_{\mathrm{ctx}}, Z_{\mathrm{qry}})
$
Here, $Z_{\mathrm{ctx}}$ is used to condition the pretrained diffusion model, while
$(Z_{\mathrm{qry}}, D_{\mathrm{qry}})$ is used to train feature decoders for the current dataset.
\vspace{-0.3em}
\paragraph{Latent Sampling.}
Given the context latent tensor $Z_{\mathrm{ctx}}$, let $K$ be the number of synthetic samples to generate.
We initialize a noisy latent tensor
$Z_{\sigma_{\max}} \in \mathbb{R}^{K \times F \times d}$,
where each entry is sampled independently from
$\mathcal{N}(0, \sigma_{\max}^2)$ with $\sigma_{\max}=80$.
Starting from this noise tensor, we run the reverse diffusion process for 50 steps using the EDM second-order Euler--Heun solver~\cite{karras2022elucidating}.
At each step, the denoising network $D_\theta$ removes noise from the latent tensor while conditioning on $Z_{\mathrm{ctx}}$.
This process iteratively produces clean synthetic latents
$\hat{Z}_0 \in \mathbb{R}^{K \times F \times d}$.
\vspace{-0.3em}
\paragraph{Feature Reconstruction.}
The generated latents $\hat{Z}_0$ need to be mapped back to the raw feature space of the current dataset. Because different datasets may have different feature types and value spaces, we train a separate lightweight decoder for each feature at inference time.
For feature $f$, we define an MLP decoder
$\mathcal{D}_f: \mathbb{R}^d \to \mathcal{X}_f$,
where $\mathcal{X}_f$ denotes the value space of feature $f$.
The decoder for feature $f$ is trained on the query set of the current dataset.
Let $Z_{\mathrm{qry},f} \in \mathbb{R}^{M_\text{qry} \times d}$ be the latent representations of feature $f$ for all query samples, and let $D_{\mathrm{qry},f}$ be the corresponding raw feature values.
We optimize $\mathcal{D}_f$ by minimizing
\begin{equation}
\mathcal{L}_{\mathrm{dec}}^{(f)}
=
\frac{1}{M_\text{qry}}
\sum_{i=1}^{M_\text{qry}}
l_f\bigl(
\mathcal{D}_f(Z_{\mathrm{qry},f}^{(i)}),
D_{\mathrm{qry},f}^{(i)}
\bigr),
\end{equation}
where $l_f$ is the mean squared error for numerical features and the cross-entropy loss for categorical features.
The superscript $(i)$ indexes query samples. After training the feature decoders, we apply them to the generated latents $\hat{Z}_0$. For each feature $f$, we take the corresponding latent slice
$\hat{Z}_{0,f} \in \mathbb{R}^{K \times d}$
and reconstruct the feature values by
$
\hat{D}_f = \mathcal{D}_f(\hat{Z}_{0,f}).
$
Finally, we concatenate the reconstructed feature columns
$\{\hat{D}_f\}_{f=1}^F$
to obtain the synthetic dataset
$\hat{D} \in \mathbb{R}^{K \times F}$.
Although the decoders are trained at inference time, they are small MLPs and add only a limited computational cost.
The overall inference time is comparable to single-stage generative models such as TVAE and tabDDPM, and is much lower than two-stage method TabSYN. We report the running time of different methods in Table~\ref{tab:running_time}.
More details about the decoder architecture and training configuration are provided in Appendix~\ref{sec:decoder_details}.
Any privacy risk introduced by the decoder training step is included in our end-to-end empirical privacy evaluation in Section~\ref{sec:exp2}. 

\vspace{-0.3em}
\section{Experiment}
\vspace{-0.3em}
\label{sec:exp}
\vspace{-0.3em}
\paragraph{Baselines.}
We compare our method with representative tabular data generation approaches across different modeling paradigms, including VAE-based (TVAE~\cite{xu2019modeling}), GAN-based (CTGAN~\cite{xu2019modeling} and CTABGAN+~\cite{zhao2024ctab}), diffusion-based (tabDDPM~\cite{kotelnikov2023tabddpm}, TabDiff~\cite{shi2024tabdiff} and TabSYN~\cite{zhang2023mixed}), and LLM-based (GReaT~\cite{borisov2022language}) methods. We provide a detailed introduction of these methods in Appendix~\ref{app:baselines}
\vspace{-0.3em}
\paragraph{Datasets.}
We evaluate DiffICL and the baseline methods on 7 classification datasets (kc2~\cite{kc2}, wdbc~\cite{wdbc}, ilpd~\cite{ilpd}, diabetes~\cite{smith1988using}, vehicle~\cite{Siebert1987VehicleRU}, mfeat-zernike~\cite{jain2000statistical}, kr-vs-kp~\cite{shapiro1987structured}) and 7 regression datasets (red wine~\cite{cortez2009modeling}, auction verification~\cite{ordoni2022analyzing}, abalone~\cite{nash1994population}, white wine~\cite{cortez2009modeling}, wind~\cite{liu2025talent}, cpu activity~\cite{rasmussen2003delve}, health insurance~\cite{olson1998comparison}). For each dataset, we randomly split the samples into equal halves for training and testing. The training set sizes range from 259 to 10,535. The datasets cover a diverse range of feature types, including fully numerical, fully categorical, and mixed-type tabular data. Detailed dataset metadata is provided in Table~\ref{tab:Dataset}.  
\vspace{-0.3em}
\subsection{On the Reliability of Quality Metrics}
\label{sec:exp1}
\begin{figure}[t]
    \centering
    \begin{subfigure}{0.98\textwidth}
        \centering
        \includegraphics[width=\linewidth]{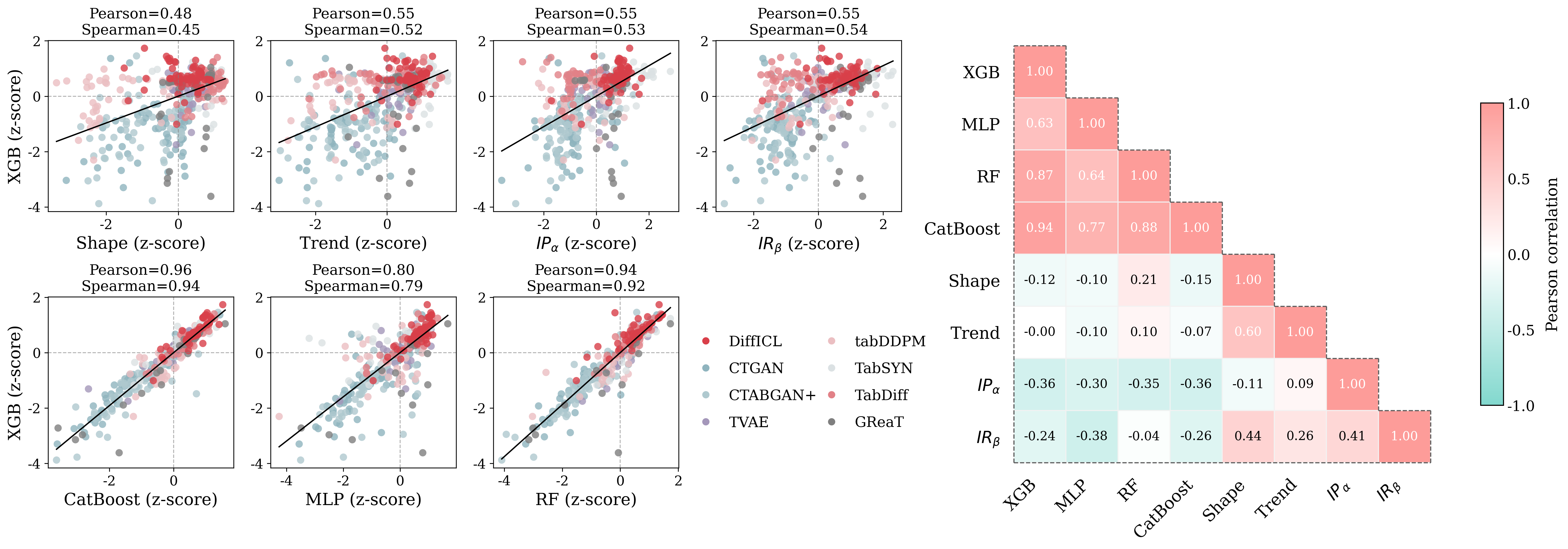}
    \end{subfigure}
    \vspace{-0.8em}
    \caption{
Alignment between different quality metrics.
\small{\textbf{Left}: pairwise relationships between downstream performance (XGBoost) and representative distribution-based metrics (Shape, Trend, $IP_\alpha$, $IR_\beta$), as well as other downstream learners, after z-score normalization and aggregation across all datasets and runs. Each point corresponds to a (dataset, method, seed) instance. 
\textbf{Right}: Pearson correlation matrix on the \textit{diabetes} dataset. 
Correlation matrices for additional datasets are provided in Figure~\ref{fig:correlation}.}
}
\vspace{-0.8em}
    \label{fig:correlation_metrics}
\end{figure} 
\vspace{-0.3em}
Existing quality metrics often yield inconsistent rankings of generative models, raising the question of which metrics truly reflect data quality. Since the primary goal of tabular data generation is to replace real data in downstream tasks, a reliable metric should measure how well models trained on synthetic data generalize to real data~\cite{alaa2022faithful,shi2024tabdiff}. We consider a set of widely used quality metrics and assess their alignment with downstream utility. 
\vspace{-0.3em}
\paragraph{Quality Metrics.} We consider two categories of quality metrics: downstream performance metrics and distributional similarity metrics. For downstream performance metrics, we assess performance using four widely adopted learners: (1) \textit{XGBoost}, (2) \textit{CatBoost}, (3) \textit{MLP}, and (4) \textit{Random Forest}. We report AUC for classification tasks and $R^2$ for regression tasks. For distributional similarity metrics, we include the followings: 
(1) \textit{Shape}~\cite{patki2016synthetic} measures column-wise similarity, with the final score averaged across all columns. 
(2) \textit{Trend}~\cite{patki2016synthetic} evaluates pairwise correlation between columns, with the final score averaged across all column pairs.  (3) $IP_\alpha$~\cite{alaa2022faithful} summarizes the fidelity of synthetic data by measuring the extent to which synthetic samples fall within high-density regions of the real distribution. 
(4) \textit{$IR_\beta$}~\cite{alaa2022faithful} captures the diversity of synthetic data by quantifying how well the synthetic distribution covers the support of the real data. Shape, Trend, $IP_\alpha$ and $IR_\beta$ metrics take values in $[0,1]$. For all quality metrics, higher values correspond to better generation quality.

\vspace{-0.3em}
\paragraph{Details}
For each real dataset $d$, we train each generative method with 5 random seeds and obtain multiple synthetic datasets. 
Each synthetic dataset is then evaluated by all quality metrics. 
For downstream utility, we train downstream learners (XGBoost, MLP, Random Forest, and CatBoost), on the synthetic data and evaluate it on the held-out real test data.  For distribution-based metrics, including Shape, Trend, $IP_\alpha$, and $IR_\beta$, we compare the synthetic data with the corresponding real training data using the standard metric definitions.
\vspace{-0.3em}
\paragraph{Distribution-Based Metrics Fail to Reflect Downstream Utility.}
We first examine metric agreement within each dataset. 
For a fixed dataset $d$, each synthetic dataset generated by one method and one random seed gives one evaluation point, and we compute pairwise Pearson correlations between metrics over these points. 
The right panel of Fig.~\ref{fig:correlation_metrics} shows the correlation matrix on the \textit{diabetes} dataset as a representative example, with additional datasets shown in Fig.~\ref{fig:correlation}. 
Downstream performance metrics evaluated by different learners are strongly correlated, indicating that they provide consistent assessments of synthetic data utility. 
In contrast, distribution-based metrics, including Shape, Trend, $IP_\alpha$, and $IR_\beta$, show weak correlations with downstream performance, suggesting that distributional similarity does not necessarily translate into downstream utility.

We further aggregate results across all datasets. 
Because different datasets may have different metric scales and variances, we normalize each metric within each dataset:
\begin{equation}
z_i^{(d,m)} = \frac{x_i^{(d,m)} - \mu^{(d,m)}}{\sigma^{(d,m)}},
\end{equation}
where $x_i^{(d,m)}$ denotes a valid metric value for metric $m$ on dataset $d$, and $\mu^{(d,m)}$ and $\sigma^{(d,m)}$ are computed over all methods and seeds within that dataset. 
We then concatenate the normalized values across datasets and compute pairwise Pearson and Spearman correlations. 
Consistent with the within-dataset results, distribution-based metrics show weak correlations with ML efficiency metrics, with Pearson correlations below $0.6$ in the left panel of Fig.~\ref{fig:correlation_metrics}. 
Shape, which mainly captures low-order marginal distributions, has the weakest correlation with ML efficiency, with Pearson correlation $0.48$. 
These results indicate that distribution-based metrics are poorly aligned with downstream utility and cannot faithfully reflect the practical quality of synthetic tabular data.

\begin{table*}[t]
\centering
\vspace{-0.3em}
\caption{Quality and Privacy comparison across 9 and 14 datasets. 
\small{The results represent the mean $\pm$ standard deviation (\%) over 5 independent runs. Data quality is measured by downstream task performance (using XGBoost, CatBoost, RandomForest, and MLP as learners), and privacy is quantified by DCROverfit. GReaT is included in the 9 Datasets group where it ran successfully, but excluded from the 14 Datasets group as it failed to generate valid samples across all subsets. Best results are highlighted in \textcolor[HTML]{D93F49}{red}.}}
\label{tab:syn_performance}
\vspace{-0.3em}
\setlength{\tabcolsep}{2.5pt} 
\resizebox{\textwidth}{!}{%
\begin{tabular}{lcccccccccc}
\toprule
\textbf{Method} & \multicolumn{5}{c}{\textbf{9 Datasets}} & \multicolumn{5}{c}{\textbf{14 Datasets}} \\
\cmidrule(r){2-6} \cmidrule(l){7-11}
& \textbf{XGB} & \textbf{CatB} & \textbf{RF} & \textbf{MLP} & \textbf{DCR} & \textbf{XGB} & \textbf{CatB} & \textbf{RF} & \textbf{MLP} & \textbf{DCR} \\
\midrule
CTGAN    & 32.4\std{1.6} & 36.8\std{1.5} & 34.5\std{1.8} & 36.7\std{1.6} & 69.0\std{2.0} & 42.8\std{1.1} & 46.1\std{1.0} & 44.6\std{1.0} & 45.6\std{1.1} & 70.3\std{1.7} \\
CTABGAN+ & 23.6\std{9.2} & 36.4\std{6.1} & 29.6\std{7.3} & 39.1\std{4.3} & 68.4\std{0.8} & 36.5\std{6.0} & 46.4\std{3.6} & 41.4\std{4.8} & 48.3\std{2.0} & 70.8\std{0.5} \\
GReaT    & 61.9\std{3.7} & 65.4\std{1.8} & 63.6\std{1.9} & 64.2\std{1.3} & 73.4\std{1.1} & - & - & - & - & - \\
TVAE     & 62.9\std{1.1} & 64.8\std{0.9} & 63.6\std{1.1} & 61.5\std{1.5} & 71.2\std{0.7} & 68.6\std{0.9} & 70.2\std{0.9} & 69.5\std{0.8} & 68.1\std{1.0} & 72.7\std{0.5} \\
tabDDPM  & 62.7\std{1.0} & 65.7\std{0.7} & 62.8\std{0.8} & 60.8\std{2.0} & 66.4\std{0.9} & 68.1\std{0.7} & 71.0\std{0.6} & 69.3\std{0.5} & 68.9\std{1.2} & 61.9\std{1.9} \\
TabSYN   & 66.3\std{1.2} & 68.2\std{1.3} & 66.8\std{1.2} & 62.4\std{5.6} & 66.8\std{1.0} & 71.4\std{0.8} & 72.8\std{0.9} & 72.0\std{0.8} & 67.8\std{3.5} & 66.3\std{0.7} \\
TabDiff  & 56.6\std{1.9} & 60.4\std{1.6} & 56.5\std{2.3} & 58.8\std{2.0} & 72.8\std{0.4} & 64.8\std{1.2} & 67.7\std{1.1} & 64.9\std{1.5} & 66.5\std{1.3} & 73.2\std{0.5} \\
DiffICL  & \textcolor[HTML]{D93F49}{70.2\std{0.7}} & \textcolor[HTML]{D93F49}{71.0\std{0.5}} & \textcolor[HTML]{D93F49}{69.6\std{0.8}} & \textcolor[HTML]{D93F49}{69.7\std{1.1}} & \textcolor[HTML]{D93F49}{74.4\std{0.5}} & \textcolor[HTML]{D93F49}{74.0\std{0.7}} & \textcolor[HTML]{D93F49}{74.6\std{0.4}} & \textcolor[HTML]{D93F49}{73.4\std{0.6}} & \textcolor[HTML]{D93F49}{73.5\std{0.9}} & \textcolor[HTML]{D93F49}{76.1\std{0.4}} \\
\bottomrule
\end{tabular}%
}
\end{table*}
\vspace{-0.3em}
\subsection{Bridging Quality and Privacy: Can DiffICL Reshape the Trade‑off Frontier? }
\vspace{-0.3em}
\label{sec:exp2}
While all generative models are fundamentally subject to a tradeoff between data utility and privacy risk, the attainable tradeoff frontier can vary significantly across different methods. In this experiment, we investigate a central question: can DiffICL push this frontier outward by simultaneously improving both data quality and privacy protection in tabular data generation?
We use downstream task performance as the quality metric, measured using four representative learners: XGBoost, MLP, CatBoost, and Random Forest. This choice is motivated by the observation that distribution-based metrics often fail to faithfully reflect the practical utility of synthetic data (Section~\ref{sec:exp1}). For privacy evaluation, we adopt the commonly used DCROverfit to quantify the extent of potential data leakage.
\vspace{-0.8em}
\paragraph{Details.}
Since data quality and privacy risk both vary during training, we select checkpoints for each baseline using the same balanced criterion. 
For each method, we evaluate multiple checkpoints and report the one that maximizes
$
\frac{1}{2}\cdot \texttt{XGBoost performance} + \frac{1}{2}\cdot \texttt{DCROverfit}
$, as detailed in Appendix~\ref{app:baselines}. 
Due to sample-generation failures of GReaT on several datasets, we report two evaluation settings: comparison with all methods on the datasets where GReaT successfully generates valid samples, and comparison of the consistently successful methods across all 14 datasets. 
\vspace{-0.4em}
\paragraph{DiffICL Improves Both Data Quality and Privacy Simultaneously.}
DiffICL achieves the best performance in terms of both data quality and privacy, as shown in Table~\ref{tab:syn_performance}. This result suggests that for a given dataset, the quality-privacy tradeoff is not a fixed limitation, but can be improved through more effective model architectures and training paradigms. 
\vspace{-0.3em}
\subsection{Data Augmentation for Downstream Tasks}
\vspace{-0.3em}
\label{sec:exp3}
While the experiment in Section~\ref{sec:exp2} demonstrates the utility of using synthetic data to completely replace real data for privacy protection purposes, another application of generative models is data augmentation. In this section, we evaluate whether our generated data can improve the performance of downstream tasks when combined with real data. We construct augmented training sets by combining the original real data with 2500 synthetic data. We then evaluate downstream models trained on these augmented sets against the models trained on real data, assessing whether the generated samples effectively improve model generalization. 
\begin{table*}[t]
\centering
\vspace{-0.3em}
\caption{Data augmentation performance comparison across 9 and 14 datasets.
\small{The results represent the mean $\pm$ standard deviation (\%) over 5 independent runs. We compare the downstream task performance of models trained solely on real data versus those trained on an augmented dataset (real + synthetic) generated by various methods. Best results in each column are highlighted in \textcolor[HTML]{D93F49}{red}, and the second-best in \textcolor[HTML]{499BC0}{blue}.}}
\vspace{-0.3em}
\label{tab:aug_performance}

\setlength{\tabcolsep}{5pt} 
\resizebox{\textwidth}{!}{%
\begin{tabular}{lcccccccc}
\toprule
\textbf{Method} & \multicolumn{4}{c}{\textbf{9 Datasets}} & \multicolumn{4}{c}{\textbf{14 Datasets}} \\
\cmidrule(r){2-5} \cmidrule(l){6-9}
& \textbf{XGB} & \textbf{CatB} & \textbf{RF} & \textbf{MLP} & \textbf{XGB} & \textbf{CatB} & \textbf{RF} & \textbf{MLP} \\
\midrule
Real Data &  \textcolor[HTML]{499BC0}{74.2\std{0.0}} & \textcolor[HTML]{499BC0}{74.2\std{0.1}} & \textcolor[HTML]{499BC0}{73.5\std{0.1}} &  \textcolor[HTML]{499BC0}{72.2\std{0.4}} &  \textcolor[HTML]{499BC0}{76.9\std{0.0}} & \textcolor[HTML]{499BC0}{76.9\std{0.1}} &  \textcolor[HTML]{499BC0}{76.8\std{0.1}} &  \textcolor[HTML]{499BC0}{75.6\std{0.3}} \\
\midrule
CTGAN    & 70.0\std{0.8} & 69.0\std{0.8} & 66.9\std{0.9} & 66.5\std{0.7} & 73.5\std{0.6} & 72.8\std{0.8} & 71.9\std{0.7} & 70.7\std{0.9} \\
CTABGAN+ & 73.0\std{0.3} & 71.8\std{0.4} & 72.4\std{0.2} & 69.6\std{0.3} & 75.9\std{0.3} & 75.2\std{0.4} & 75.8\std{0.2} & 73.5\std{0.3} \\
GReaT    & 65.8\std{1.1} & 66.2\std{0.8} & 64.3\std{1.4} & 65.0\std{1.2} & - & - & - & - \\
TVAE     & 72.5\std{0.5} & 72.0\std{0.5} & 70.9\std{0.5} & 71.0\std{0.3} & 75.8\std{0.4} & 75.7\std{0.4} & 75.2\std{0.4} & 74.7\std{0.2} \\
tabDDPM  & 72.6\std{0.4} & 72.0\std{0.4} & 71.1\std{0.4} & 69.2\std{0.3} & 75.7\std{0.2} & 75.3\std{0.2} & 75.3\std{0.2} & 73.8\std{0.2} \\
TabSYN   & 72.8\std{0.2} & 72.4\std{0.2} & 71.7\std{0.4} & 70.7\std{0.6} & 76.0\std{0.3} & 75.7\std{0.2} & 75.7\std{0.3} & 73.4\std{0.6} \\
TabDiff  & 71.6\std{0.3} & 70.4\std{0.3} & 69.2\std{0.3} & 66.7\std{0.2} & 75.3\std{0.1} & 74.7\std{0.1} & 74.0\std{0.2} & 72.1\std{0.2} \\
DiffICL  & \textcolor[HTML]{D93F49}{74.6\std{0.2}} &  \textcolor[HTML]{D93F49}{75.0\std{0.2}} & \textcolor[HTML]{D93F49}{73.8\std{0.2}} & \textcolor[HTML]{D93F49}{73.0\std{0.8}} & \textcolor[HTML]{D93F49}{79.8\std{0.1}} &  \textcolor[HTML]{D93F49}{77.1\std{0.2}} & \textcolor[HTML]{D93F49}{79.6\std{0.1}} & \textcolor[HTML]{D93F49}{78.8\std{0.5}} \\
\bottomrule
\end{tabular}%
}
\end{table*}
\vspace{-0.3em}
\begin{table*}[htbp]
\centering
\vspace{-0.3em}
\caption{Ablation study on the source of performance gains.
\small{We compare the full DiffICL model against two variants: \textbf{DiffICL(S)}, which restricts pre-training to the single target dataset (removing cross-dataset priors), and \textbf{DiffICL(N)}, which omits permutation-based corpus expansion during pre-training. Downstream performance is evaluated under two application scenarios: using purely synthetic data as a replacement for real data (\textbf{Syn.}), and using a mixture of real and synthetic data for data augmentation (\textbf{Aug.}). The results report the mean $\pm$ standard deviation (\%). The best results in each column are highlighted in \textcolor[HTML]{D93F49}{red}.}}
\vspace{-0.3em}
\label{tab:ablation}
\setlength{\tabcolsep}{4pt} 
\resizebox{\textwidth}{!}{
\begin{tabular}{lcccccccc}
\toprule
\multirow{2.5}{*}{\textbf{Method}} & \multicolumn{2}{c}{\textbf{XGB}} & \multicolumn{2}{c}{\textbf{CatB}} & \multicolumn{2}{c}{\textbf{RF}} & \multicolumn{2}{c}{\textbf{MLP}} \\
\cmidrule(lr){2-3} \cmidrule(lr){4-5} \cmidrule(lr){6-7} \cmidrule(lr){8-9}
& \textbf{Syn.} & \textbf{Aug.} & \textbf{Syn.} & \textbf{Aug.} & \textbf{Syn.} & \textbf{Aug.} & \textbf{Syn.} & \textbf{Aug.} \\
\midrule
DiffICL    & \textcolor[HTML]{D93F49}{74.0 \std{0.7}} & \textcolor[HTML]{D93F49}{79.8 \std{0.1}} & \textcolor[HTML]{D93F49}{74.6 \std{0.4}} & \textcolor[HTML]{D93F49}{77.1 \std{0.2}} & \textcolor[HTML]{D93F49}{73.4 \std{0.6}} & \textcolor[HTML]{D93F49}{79.6 \std{0.1}} & \textcolor[HTML]{D93F49}{73.5 \std{0.9}} & \textcolor[HTML]{D93F49}{78.8 \std{0.5}} \\
DiffICL(S) & 57.0 \std{1.2} & 76.3 \std{0.3} & 59.1 \std{1.1} & 75.6 \std{0.4} & 58.3 \std{1.2} & 76.0 \std{0.3} & 56.2 \std{2.6} & 73.6 \std{0.5} \\
DiffICL(N) & 70.8 \std{0.2} & 75.2 \std{0.1} & 71.5 \std{0.3} & 74.5 \std{0.1} & 70.3 \std{0.4} & 75.1 \std{0.2} & 70.1 \std{0.8} & 74.2 \std{0.4} \\
\bottomrule
\end{tabular}
}
\vspace{-1em}
\end{table*}
\vspace{-0.3em}
\paragraph{DiffICL Enables Effective Data Augmentation.} As demonstrated in Table~\ref{tab:aug_performance}, integrating synthetic samples generated by DiffICL with the real training set yields performance improvements for all downstream learners. Among all our evaluated deep generative models, DiffICL emerges as the sole method capable of effective data augmentation. Conversely, augmenting the training set with samples from other approaches universally degrades the predictive performance compared to the real-data baseline. We attribute the augmentation capability of DiffICL to the ICL pretraining paradigm. Standard tabular generative models learn the target distribution from scratch. Consequently, their hypothesis space is strictly bounded by the empirical distribution of the limited training samples. In contrast, DiffICL is pretrained across a vast, diverse corpus of tabular datasets, allowing it to acquire strong structural priors. By leveraging its extensive prior knowledge to complement the limited empirical data, DiffICL effectively mitigates the information bottleneck in finite training sets. 
\vspace{-0.5em}
\subsection{Disentangling the Sources of DiffICL’s Performance Gain}
\label{sec:exp4}
\vspace{-0.3em}
The preceding experiments show that DiffICL improves the quality--privacy frontier (Section~\ref{sec:exp2}) and enables effective data augmentation (Section~\ref{sec:exp3}). 
We next conduct ablation studies to identify the sources of these gains. This section focuses on the construction of the pretraining corpus, studying the contribution of multi-dataset pretraining and permutation-based corpus expansion. We defer inference-stage analyses to Appendix~\ref{app:ablation}, including the effect of the context ratio and the number of training samples.  

\textbf{The Role of Multi-Dataset Priors.}
We first evaluate DiffICL(S), a variant that uses the same encoder and latent diffusion architecture as DiffICL, but is pretrained only on the target training dataset. 
This variant isolates the contribution of multi-dataset ICL pretraining from that of the encoded tabular representations. 
As shown in Table~\ref{tab:ablation}, DiffICL substantially outperforms DiffICL(S), indicating that the performance gains cannot be explained by the LimiX representations alone. 
Instead, they rely on structural priors acquired through multi-dataset pretraining. Without these priors, DiffICL(S) remains constrained by the empirical distribution of a single dataset and generalizes less effectively beyond the observed samples. \textbf{The Impact of Corpus Expansion.} We then evaluate DiffICL(N), which is pretrained on the same datasets as DiffICL but without row and column permutations. 
Table~\ref{tab:ablation} shows that removing permutation-based corpus expansion consistently degrades performance, indicating that this expansion strategy is an important component of DiffICL's pretraining. 
\vspace{-0.3em}

\vspace{-0.6em}
\section{Conclusion}
\vspace{-0.6em}
We study the quality–privacy tradeoff in small-data tabular generation and proposes DiffICL, an ICL pretraining framework that uses multi-dataset structural priors to generate synthetic data from limited context. Through experiments, we then shows that commonly used distribution-based quality metrics are weakly aligned with downstream utility, motivating the use of predictive performance as the main quality measure. DiffICL achieves better utility and privacy than representative baselines. The data augmentation experiment further shows that DiffICL improves downstream performance when synthetic data are combined with real data. Finally, ablation studies confirm that these gains mainly come from multi-dataset pretraining. 

\bibliographystyle{plain} 
\bibliography{references} 
\medskip






\appendix

\section{Implementation Details}
\label{app:network}
\subsection{Pretraining Details}
\label{sec:pretraining_details}
We construct a pretraining corpus from real-world tabular datasets collected from Kaggle and the UCI Machine Learning Repository. We exclude datasets with fewer than 50 samples or more than 50 features, and further remove regression datasets with negative $R^2$ scores, which typically indicate poor data quality. This results in 826 datasets.
To increase diversity and encourage permutation invariance over rows and features, we perform data augmentation by randomly permuting the order of samples and features (while keeping the target column unchanged). For each dataset, we generate five augmented variants, resulting in a total of 4,130 datasets. For each dataset, we randomly split samples into a context set $D_{\text{ctx}}$ and a query set $D_{\text{qry}}$, where the context ratio $r$ is uniformly sampled from $[0.2, 0.5]$. Both subsets are encoded into latent representations, yielding training pairs $(Z_{\text{ctx}}, Z_{\text{qry}})$.

We pretrain the diffusion model for 20,000 epochs using a batch size of 128. When $|Z_{\text{qry}}| > 128$, we randomly subsample 128 query instances. Training is conducted on 32 NVIDIA RTX 4090 GPUs with a cosine learning rate schedule and linear warmup. The initial learning rate is set to $2 \times 10^{-4}$ with a warmup ratio of 0.05, and no weight decay is applied. The choice of hyperparameters is guided by a systematic validation protocol designed to assess \textit{cross-dataset generalization}. Specifically, from the collected dataset pool, we reserve a subset as a held-out validation set, which remains entirely unseen during pretraining. For each hyperparameter configuration, we evaluate model checkpoints by measuring the Fréchet Inception Distance (FID) between generated query latents $\hat{Z}_{\text{qry}}$ and real query latents $Z_{\text{qry}}$ on these validation datasets.
Since the validation datasets are not observed during pretraining, this FID score serves as a proxy for dataset-level generalization: it quantifies whether a conditional diffusion model trained on a collection of datasets can generate high-quality latent representations for previously unseen datasets under the in-context conditioning framework. We select hyperparameter configurations that achieve the best performance on this validation criterion. We take \textit{epoch} as a representative example: we observe that increasing the number of pretraining epochs initially improves generalization performance, but beyond a certain point leads to degradation. This behavior suggests that excessive training causes the model to overfit to the datasets seen during pretraining, thereby reducing its ability to generalize to new datasets. Similar selection principles are applied to other hyperparameters. 
\subsection{Denoising Network Architecture}
\label{sec:denoise_network}
\begin{figure}[htbp]
    \centering
    \begin{subfigure}{0.98\textwidth}
        \centering
        \includegraphics[width=\linewidth]{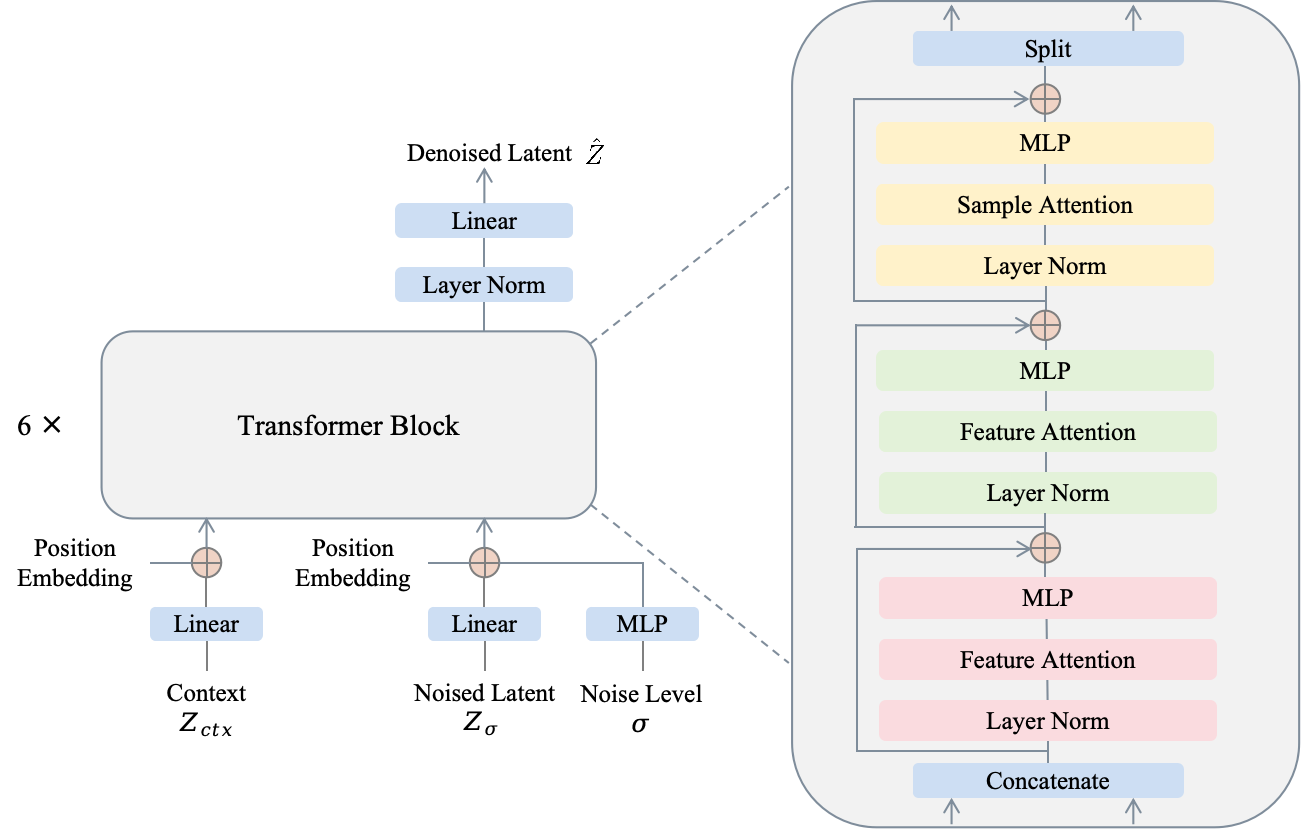}
    \end{subfigure}
    \caption{Denoise network architecture.}
    \label{fig:network}
\end{figure} 

Given a noised latent input $Z_\sigma \in \mathbb{R}^{M_{\text{qry}} \times F \times d}$, we first project it into the model dimension (512) via a linear layer. Context latents $Z_{\text{ctx}} \in \mathbb{R}^{M_{\text{ctx}} \times F \times d}$ are processed in parallel using a separate projection. To encode structural information, we add positional embeddings to both inputs: for context representations, we use 2D sinusoidal positional encodings over the sample and feature dimensions $(M_{\text{ctx}}, F)$, while for noised query representations, we apply 1D sinusoidal positional encodings along the feature dimension. In addition, diffusion timestep(noise level $\sigma$) embeddings are computed using a sinusoidal mapping followed by an MLP, and added to the query representations. The core of our denoising network consists of a stack of six transformer layers that jointly process context and query representations by concatenating them along the sample dimension, resulting in a unified representation $h\in \mathbb{R}^{(M_\text{ctx}+M_\text{qry})\times F \times 512}$ and enabling in-context learning-style conditioning. Each transformer layer comprises three attention sublayers: two feature-wise attention sublayers and one sample-wise attention sublayer. The feature-wise attention operates across feature dimensions within each sample, while the sample-wise attention captures interactions across different samples, allowing query tokens to attend to context tokens. All sublayers share the same architecture: Layer Normalization is first applied, followed by an 8-head attention module (implemented with FlashAttention for efficiency), a 2-layer MLP as feed-forward network, and a residual connection. Formally, for an input representation $h$, each sublayer computes
\[
h \leftarrow h + \mathrm{FFN}(\mathrm{Attn}(\mathrm{LayerNorm}(h))).
\]
After passing through all layers, the joint representation $h \in \mathbb{R}^{(M_{\text{ctx}} + M_{\text{qry}}) \times F \times 512}$ is split back into context and query parts. We retain only the query representations corresponding to $M_{\text{qry}}$ samples, apply a final Layer Normalization, and project them back to the latent dimension $d$ to obtain the denoised output $\hat{Z} \in \mathbb{R}^{M_{\text{qry}} \times F \times d}$. To stabilize training, we adopt an EDM-style preconditioning scheme in which the network predicts a scaled residual and the final output is formed via a noise-dependent linear combination of the input and network prediction , and the model is trained with the corresponding EDM loss across a range of noise levels.

\subsection{Decoder Architecture and Training}
\label{sec:decoder_details}
To reconstruct raw tabular data from latent representations, we train a lightweight embedding decoder for each evaluation dataset. The decoder takes per-feature latent embeddings as input and predicts the corresponding raw feature values and target variable. It assigns an independent MLP head to each predicted variable (features and target).
Each MLP head contains 2 hidden layers with hidden dimension 768, where each hidden layer is followed by ReLU activation and dropout with rate 0.1. A final linear layer maps the hidden representation to the required output dimension: feature heads output the corresponding reconstructed feature values, while the target head outputs a scalar for regression tasks or $C$ logits for classification tasks, where $C$ is the number of classes.

The decoder is trained on the query set using latent embeddings and corresponding raw values. We use mean squared error for feature reconstruction and for regression targets, and cross-entropy loss for classification targets. We optimize the decoder using Adam with a learning rate of $2\times 10^{-5}$.

\section{Related Works}
\paragraph{Diffusion Model} Building on the early diffusion probabilistic framework of Sohl-Dickstein et al.~\cite{sohl2015deep}, DDPM~\cite{ho2020denoising} reformulates reverse diffusion as noise prediction and trains the model with a simplified denoising objective, which substantially improves the practicality and sample quality of diffusion models. DDIM~\cite{song2020denoising} accelerates DDPM sampling by replacing the original Markovian diffusion process with a non-Markovian process that keeps the same training objective but allows deterministic, shorter-step sampling with much less loss in sample quality. Song et al.~\cite{song2020score} formulate score-based generative modeling in a continuous-time SDE framework, enabling flexible sampling through reverse-time SDEs, probability-flow ODEs, and predictor-corrector samplers. EDM~\cite{karras2022elucidating} improves diffusion-based generative models by preconditioning the denoising network, which rescales its inputs, outputs, skip connection, and loss across noise levels to make training more stable. It also introduces a more efficient sampler that combines improved noise schedules, second-order Heun integration, and controlled stochasticity. 

\paragraph{Tabular generative model} TableGAN~\cite{park2018data} uses a GAN-based architecture with an additional classifier, information loss, and classification loss to synthesize relational tables whose records are statistically similar to the original data and maintain semantic consistency among attributes. CTGAN~\cite{xu2019modeling} models continuous columns by decomposing each value into a mode indicator and a normalized value within that mode, while modeling discrete columns with one-hot representations and a conditional generator trained with sampling strategies that give rare categories more training opportunities. TVAE adapts the variational autoencoder framework to mixed-type tabular data by modeling continuous values with Gaussian distributions and discrete values with categorical distributions. CTAB-GAN~\cite{zhao2021ctab} extends CTGAN to more complex industrial tabular data by using a mixed-type encoder for variables that combine categorical and continuous values, and applying logarithm transformation to better handle long-tail continuous distributions. CTAB-GAN+\cite{zhao2024ctab} improves CTAB-GAN by combining more adaptive feature encoding and more stable Wasserstein training with downstream supervision, and further extends the framework to differentially private tabular data synthesis. GOGGLE~\cite{liu2023goggle} is a VAE-based tabular generative model that replaces the standard MLP decoder with a message-passing neural network operating on a learned relational structure between variables. As GAN-based tabular generators remain difficult to train and can suffer from mode collapse, recent studies have increasingly explored diffusion models for tabular synthesis. STaSy~\cite{kim2022stasy} adopts score-based generative modeling for tabular synthesis and introduces self-paced learning plus log-probability-based fine-tuning to improve the stability of denoising score matching. TabDDPM~\cite{kotelnikov2023tabddpm} proposes a DDPM framework that uses Gaussian diffusion for numerical features, multinomial diffusion for categorical features, and an MLP-based denoising network to generate mixed-type synthetic tables. CoDi~\cite{lee2023codi} uses two coupled diffusion models, one for continuous variables and one for discrete variables, and trains them to condition on each other with an additional contrastive loss. TabSyn~\cite{zhang2023mixed} first maps numerical and categorical columns into a unified VAE latent space with Transformer-based encoders and decoders, and then training a score-based diffusion model in this continuous latent space. TabDiff~\cite{shi2024tabdiff} directly models numerical columns with Gaussian diffusion and categorical columns with masked discrete diffusion, and learns a separate noise schedule for each column to handle the heterogeneous distributions of mixed-type tabular data. 

\paragraph{Tabular Foundation Model} TabPFN~\cite{hollmann2025accurate} is the first tabular foundation model pretrained on a large collection of synthetic tabular tasks, enabling it to perform classification and regression by conditioning on a labeled context set without task-specific training or hyperparameter tuning. TabICL~\cite{qu2025tabicl} scales tabular in-context learning to larger classification datasets by compressing each row into a fixed-dimensional embedding and then applying a Transformer over these row embeddings for efficient prediction from a labeled context set. TabDPT~\cite{ma2024tabdpt} trains tabular foundation model on real datasets with self-supervised learning and retrieval. Real-TabPFN~\cite{garg2025real} improves TabPFN by continuing pretraining on a small, curated set of large real-world tabular datasets, reducing the synthetic-to-real gap. Mitra~\cite{zhang2025mitra} argues that the key to improving synthetic-data-pretrained tabular foundation models lies in the design of synthetic priors, and trains a model on a carefully selected mixture of priors. Unlike tabular foundation models designed mainly for prediction, LimiX~\cite{zhang2025limix} pretrains one Transformer to handle partially observed tables, allowing the same model to perform prediction, regression and missing value imputation through a unified framework.

\section{Experimental Details}
\subsection{Dataset metadata}
We report the metadata of the datasets used in the experiments in Table~\ref{tab:Dataset}. The number of features excludes the target variable. \textit{We manually verify that there is no overlap between the evaluation datasets and the pretraining corpus.} The \texttt{wind} dataset is obtained from Hugging Face\footnote{\url{https://huggingface.co/datasets/LAMDA-Tabular/TALENT}}. All other datasets are obtained from OpenML\footnote{\url{https://www.openml.org}}.

\begin{table}[t]
\centering
\caption{Summary of the datasets used in the experiments.}
\label{tab:Dataset}
\begin{tabular}{l l c c c}
\toprule
\textbf{Task Type} & \textbf{Dataset Name} & \textbf{Training Size} & \textbf{Categorical} & \textbf{Numerical} \\
\midrule

\multirow{7}{*}{Classification}
& kc2              & 259  & 0  & 21 \\ 
& wdbc             & 282  & 0  & 30 \\ 
& ilpd             & 289  & 1  & 9  \\ 
& diabetes         & 382  & 0  & 8  \\ 
& vehicle          & 419  & 0  & 18 \\ 
& mfeat-zernike    & 990  & 0  & 47 \\ 
& kr-vs-kp         & 1596 & 36 & 0  \\ 

\midrule

\multirow{7}{*}{Regression}
& red wine            & 799   & 0 & 11 \\ 
& auction verification & 1021  & 1 & 7  \\ 
& abalone              & 2088  & 1 & 7  \\ 
& white wine          & 2449  & 0 & 11 \\ 
& wind                 & 3287  & 0 & 14 \\ 
& cpu activity        & 4096  & 0 & 21 \\ 
& health insurance    & 10535 & 7 & 5  \\ 

\bottomrule
\end{tabular}
\end{table}

\subsection{Baseline Methods}
\label{app:baselines}

We implement all baseline methods based on their respective open-source repositories, as provided in the footnotes.

\begin{itemize}

\item \textbf{TVAE.} 
TVAE models tabular data using a variational autoencoder, where each feature is encoded into a latent space and reconstructed via a decoder. It models numerical features using Gaussian distributions and categorical features using discrete distributions.\footnote{\url{https://github.com/sdv-dev/CTGAN}}

\item \textbf{CTGAN.} 
CTGAN employs a conditional generative adversarial network designed for tabular data, using conditional vectors to handle imbalanced discrete columns and mode-specific normalization to better model continuous variables.\footnote{\url{https://github.com/sdv-dev/CTGAN}}

\item \textbf{CTABGAN+.} 
CTABGAN+ extends CTGAN by introducing improved data preprocessing and training strategies, including better handling of mixed-type columns (continuous, categorical, and mixed distributions).\footnote{\url{https://github.com/Team-TUD/CTAB-GAN-Plus}}

\item \textbf{tabDDPM.} 
To the best of our knowledge, tabDDPM is the first work to apply diffusion models to tabular data generation. It adopts a discrete-time diffusion formulation, modeling the data distribution via a forward noising process and a learned reverse denoising process.\footnote{\url{https://github.com/yandex-research/tab-ddpm}}

\item \textbf{TabDiff.} 
TabDiff proposes a unified mixed-type diffusion framework. It adopts a continuous-time diffusion formulation and introduces a stochastic sampler to reduce accumulated decoding errors during sampling.\footnote{\url{https://github.com/MinkaiXu/TabDiff}}

\item \textbf{TabSYN.} 
TabSYN proposes a latent diffusion framework for tabular data generation. It first trains a variational autoencoder to project mixed-type tabular data into a continuous latent space, and then learns a diffusion model over the latent representations.\footnote{\url{https://github.com/amazon-science/tabsyn}}

\item \textbf{GReaT.} 
GReaT formulates tabular data generation as a conditional language modeling problem, where each row is serialized into a sequence and generated autoregressively using a pre-trained large language model.\footnote{\url{https://github.com/tabularis-ai/be_great}}

\end{itemize}

\paragraph{Baseline Checkpoint Selection.}
During training, generative models often improve data quality while also increasing privacy risk, making checkpoint selection important for fair comparison. 
For each baseline, we therefore evaluate multiple checkpoints and select the checkpoint that maximizes a balanced objective:
\[
\frac{1}{2}\cdot \text{XGBoost performance} + \frac{1}{2}\cdot \text{DCROverfit}.
\]
Here, XGBoost performance measures downstream utility, while DCROverfit measures privacy protection. 
For TVAE, CTGAN, CTABGAN+, TabSYN, and TabDiff, we evaluate checkpoints at training steps
$\{500, 1000, 2000, 3000, 4000, 5000\}$.
For tabDDPM, which generally requires longer training to converge, we use
$\{1000, 2000, 3000, 4000, 5000, 8000, 10000\}$.
For GReaT, we consider checkpoints at
$\{1000, 2000, 5000, 10000\}$. For all baseline methods other than GReaT, the learning rate is set to $2\times 10^{-4}$ with a training batch size of 500. For GReaT, we retain its default learning rate and reduce the batch size to 8 due to memory constraints.  

\paragraph{Handling GReaT Generation Failures.}
We observe that GReaT fails to generate valid samples on 9 datasets, including red wine, mfeat-zernike, vehicle, wdbc, kc2, kr-vs-kp, white wine, wind, and cpu activity, even after 10,000 training steps. 
Since training GReaT for 10,000 steps already requires substantial computational cost, approximately 10 hours in our setting, we do not further increase its training budget. 
Instead, following the official recommendation, we enable guided sampling, which allows GReaT to successfully generate valid samples on 4 of these 9 datasets: kr-vs-kp, white wine, wind, and cpu activity. 
Based on these observations, we report results under two settings. 
First, we compare all methods on the 9 datasets where GReaT successfully produces valid samples. 
Second, we compare the 7 consistently successful methods across all 14 datasets.

\subsection{Running Time}
Table~\ref{tab:running_time} presents the running times of the evaluated generation methods across the Diabetes, Health Insurance, kr-vs-kp, and CPU Activity datasets. For the baseline models, the execution time accounts for training model and generating 2,500 synthetic samples. The training phase consists of 5,000 steps for tabDDPM and 2,000 steps for the remaining baselines. 
For DiffICL, the running time accounts for training the dataset-specific decoders and sampling 2500 samples. DiffICL's pre-training time is excluded from this analysis. In practice, generating new samples requires only inference with the pre-trained model weights, which we intend to release upon acceptance of this paper. Notably, DiffICL demonstrates top-tier computational efficiency, ranking among the most efficient methods alongside tabDDPM and TVAE. We also observe that GReaT incurs a substantially higher computational cost when the guided-samples=True setting is required to generate valid samples (as seen in the kr-vs-kp and CPU Activity datasets).

\begin{table}[htbp]
\centering
\caption{Running Time Comparison of Different Methods(Mean $\pm$ Std)}
\label{tab:running_time}
\begin{tabular}{lcccc}
\toprule
\textbf{Method} & \textbf{diabetes} & \textbf{cpu activity} & \textbf{kr-vs-kp} & \textbf{health insurance} \\
\midrule
CTABGAN+ & 52.12 $\pm$ 3.04 & 21.48 $\pm$ 0.63 & 21.88 $\pm$ 0.81 & 215.16 $\pm$ 19.16 \\
CTGAN    & 40.37 $\pm$ 0.25 & 57.10 $\pm$ 1.75 & 55.65 $\pm$ 0.70 & 266.32 $\pm$ 10.60 \\
tabDDPM  & 13.50 $\pm$ 0.21 & 12.05 $\pm$ 0.18 & 101.47 $\pm$ 4.11 & 43.27 $\pm$ 0.94 \\
TabDiff  & 686.19 $\pm$ 26.97 & 113.96 $\pm$ 17.37 & 385.78 $\pm$ 20.38 & 101.30 $\pm$ 12.35 \\
TabSYN   & 628.47 $\pm$ 11.53 & 104.40 $\pm$ 11.49 & 272.33 $\pm$ 28.64 & 82.56 $\pm$ 12.45 \\
TVAE     & 21.12 $\pm$ 0.31 & 46.61 $\pm$ 1.29 & 21.55 $\pm$ 0.99 & 32.63 $\pm$ 1.14 \\
GReaT    & 169.86 $\pm$ 0.57 & 16537.65 $\pm$ 677.19 & 26759.59 $\pm$ 917.63 & 119.11 $\pm$ 4.35 \\
DiffICL  & 18.17 $\pm$ 2.08 & 41.37 $\pm$ 9.99 & 40.80 $\pm$ 2.91 & 69.16 $\pm$ 19.29 \\
\bottomrule
\end{tabular}
\end{table}

\subsection{Correlation between quality metrics on each evaluation dataset}
\begin{figure}[htbp]
    \centering
    
    \begin{subfigure}[b]{0.32\linewidth}
        \centering
        \includegraphics[width=\linewidth]{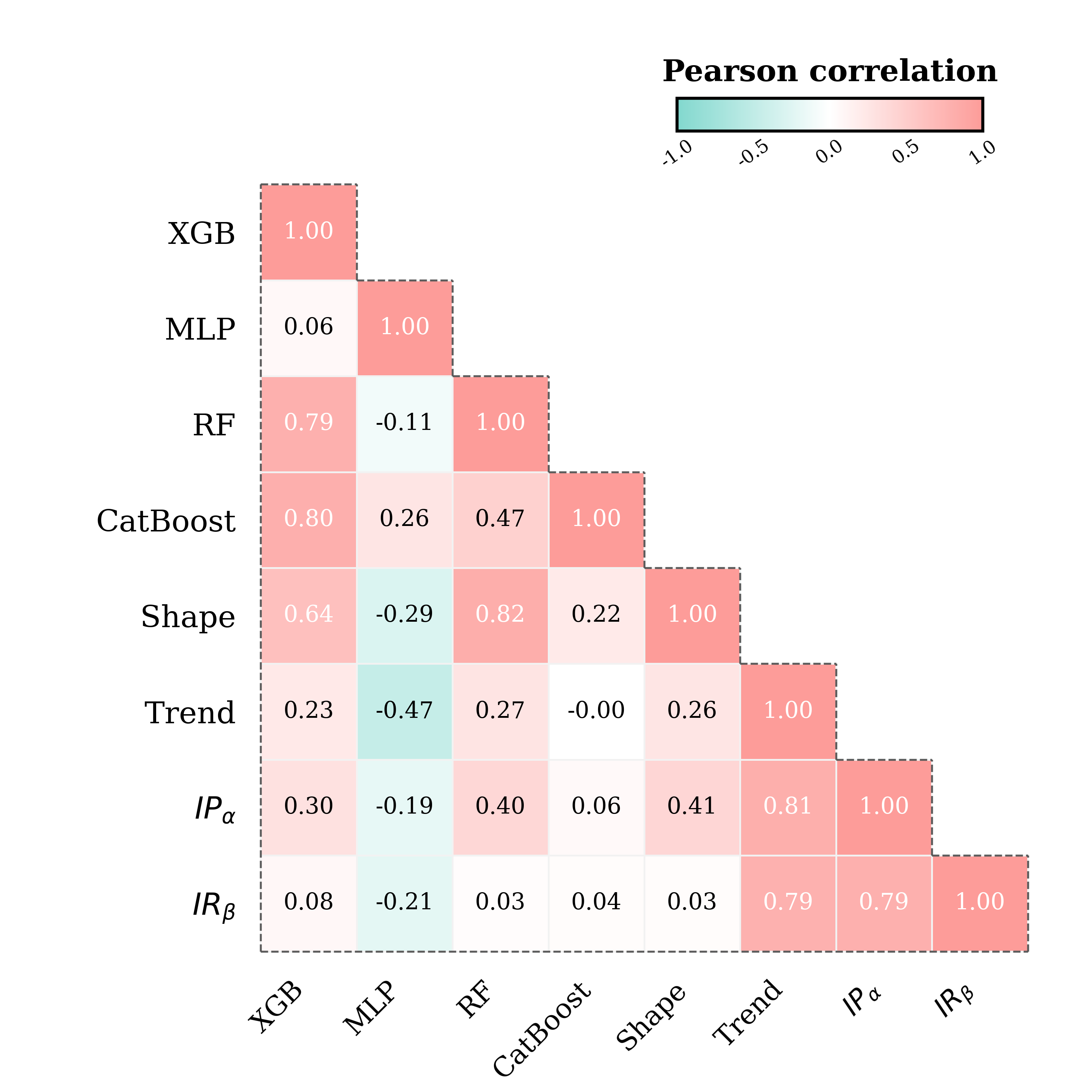}

    \end{subfigure}
    \hfill
    \begin{subfigure}[b]{0.32\linewidth}
        \centering
        \includegraphics[width=\linewidth]{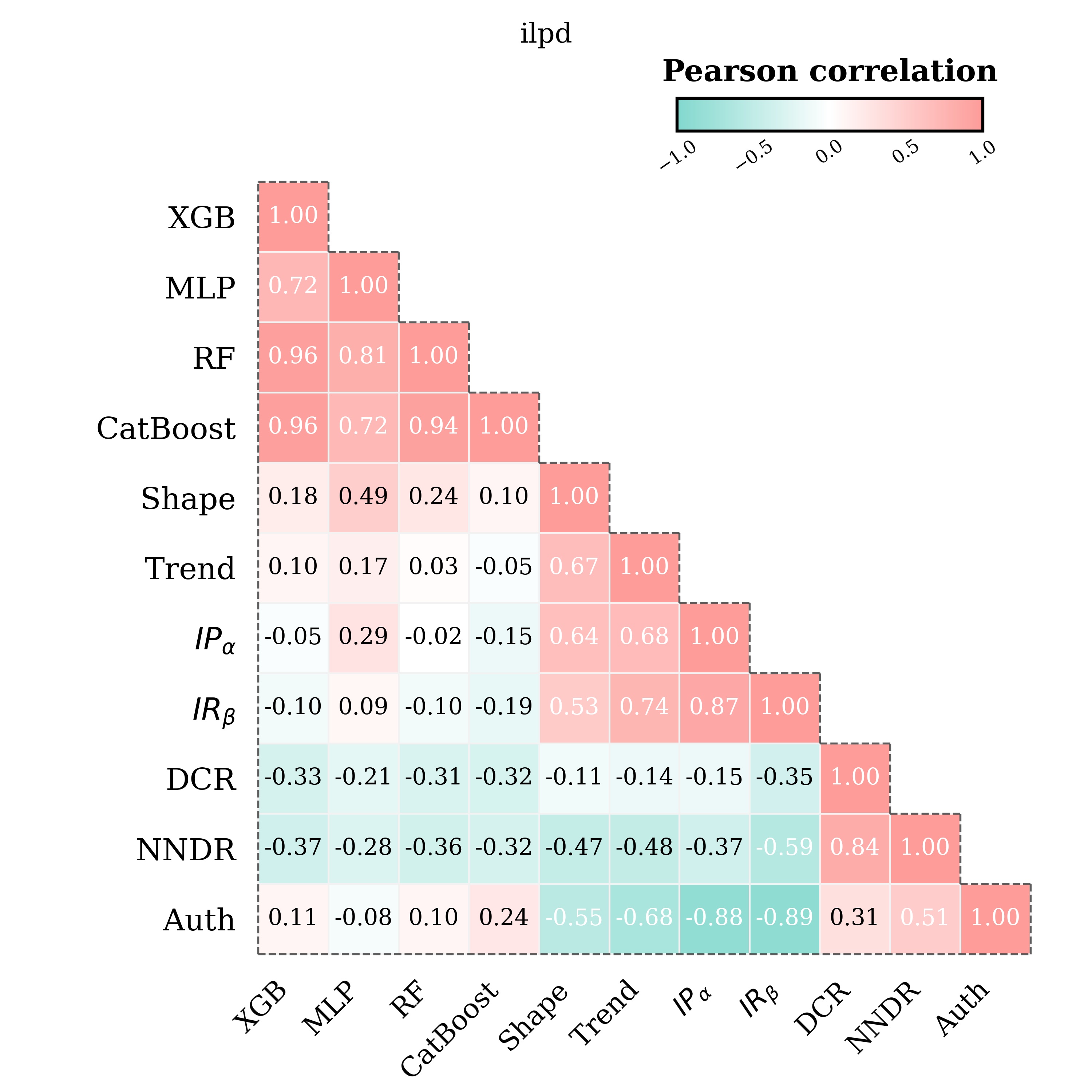}

    \end{subfigure}
    \hfill
    \begin{subfigure}[b]{0.32\linewidth}
        \centering
        \includegraphics[width=\linewidth]{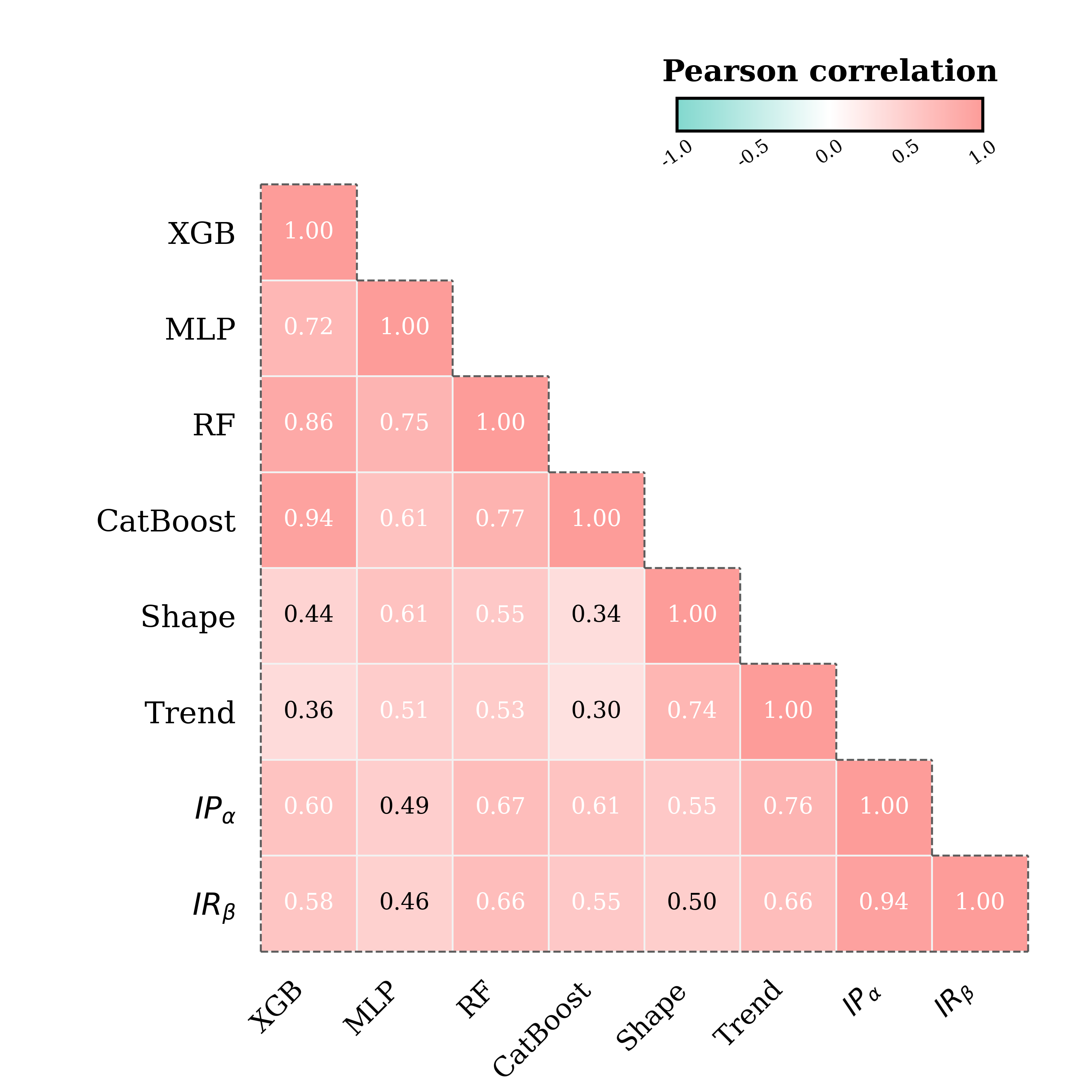}

    \end{subfigure}

    \begin{subfigure}[b]{0.32\linewidth}
        \centering
        \includegraphics[width=\linewidth]{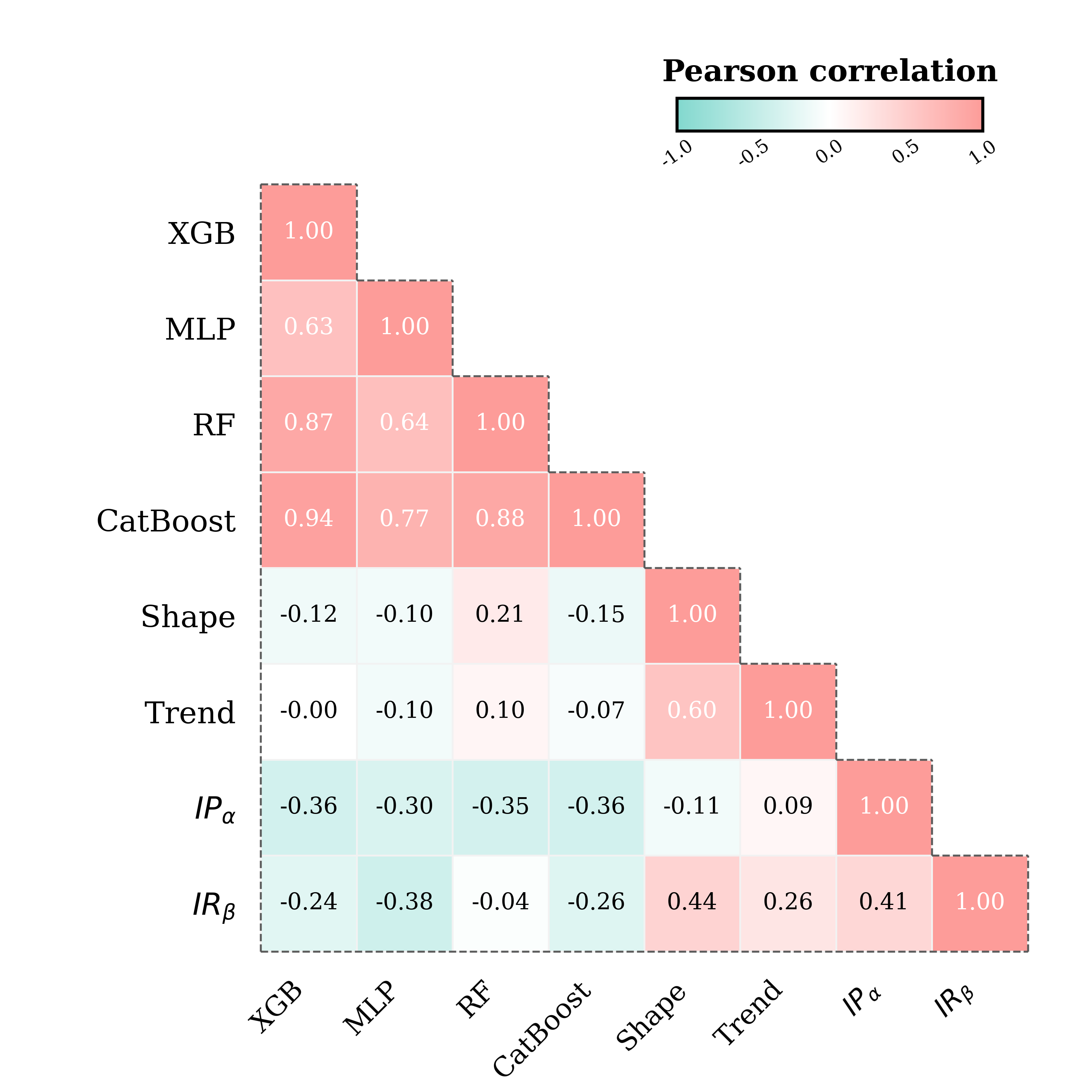}

    \end{subfigure}
    \hfill
    \begin{subfigure}[b]{0.32\linewidth}
        \centering
        \includegraphics[width=\linewidth]{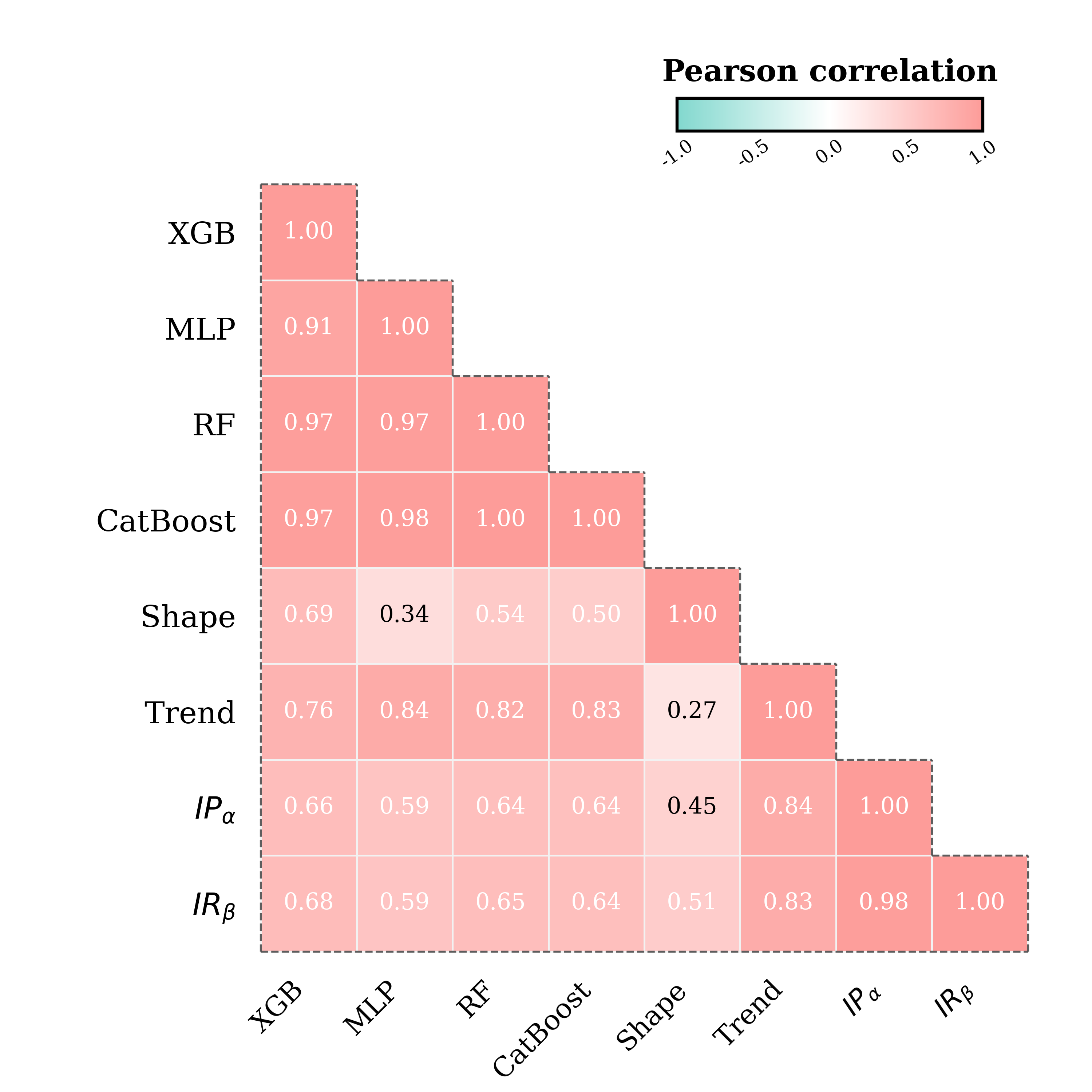}

    \end{subfigure}
    \hfill
    \begin{subfigure}[b]{0.32\linewidth}
        \centering
        \includegraphics[width=\linewidth]{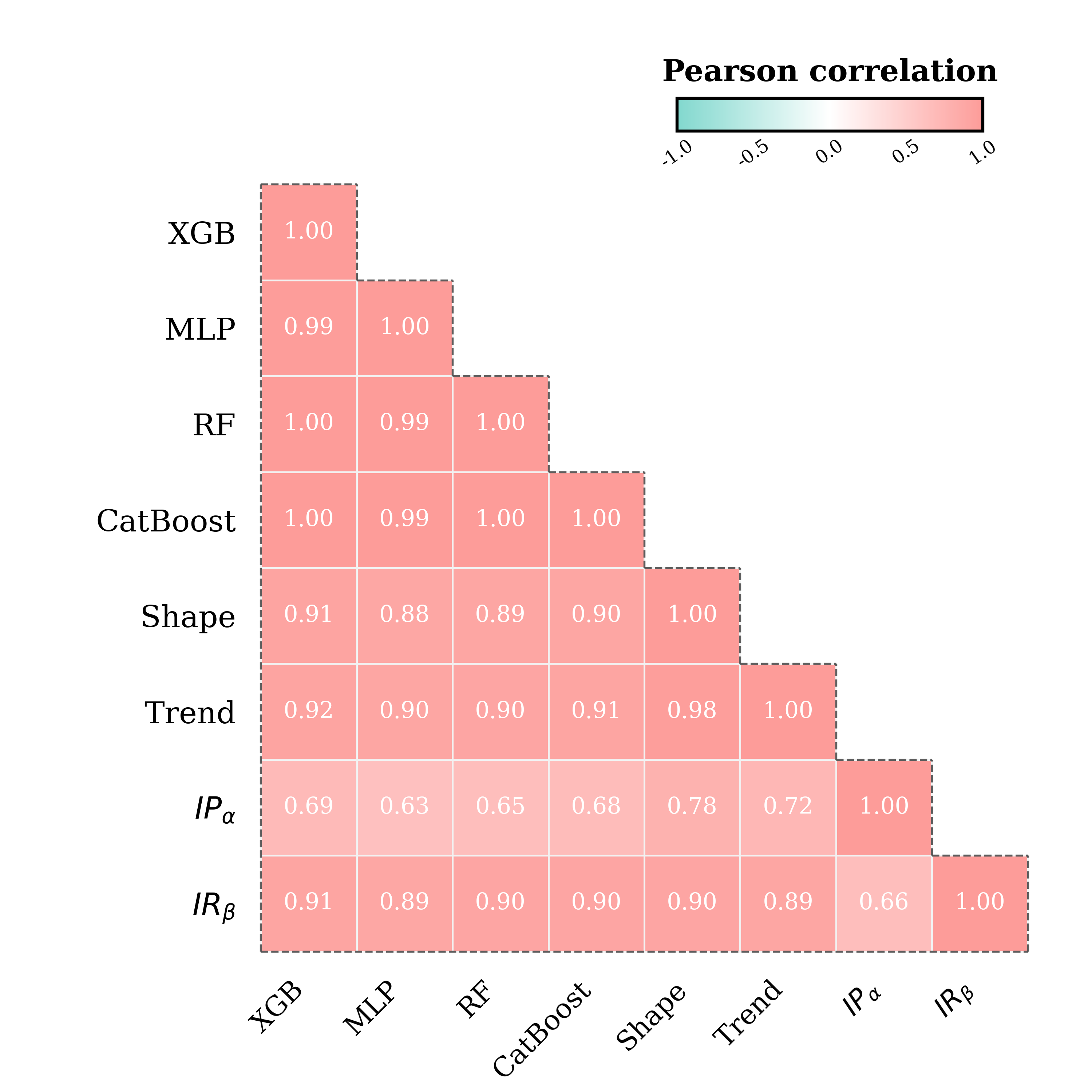}

    \end{subfigure}

    \begin{subfigure}[b]{0.32\linewidth}
        \centering
        \includegraphics[width=\linewidth]{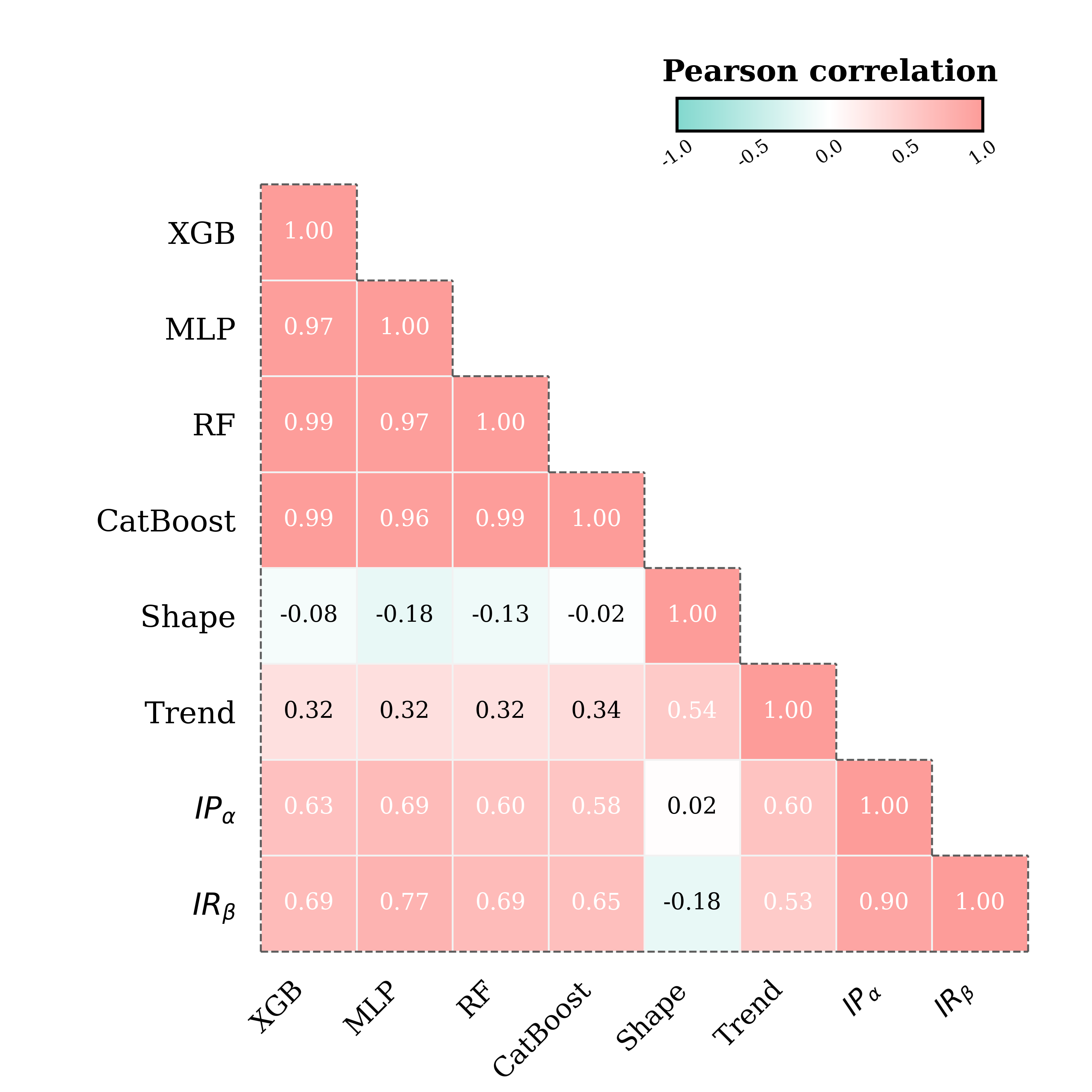}

    \end{subfigure}
    \hfill
    \begin{subfigure}[b]{0.32\linewidth}
        \centering
        \includegraphics[width=\linewidth]{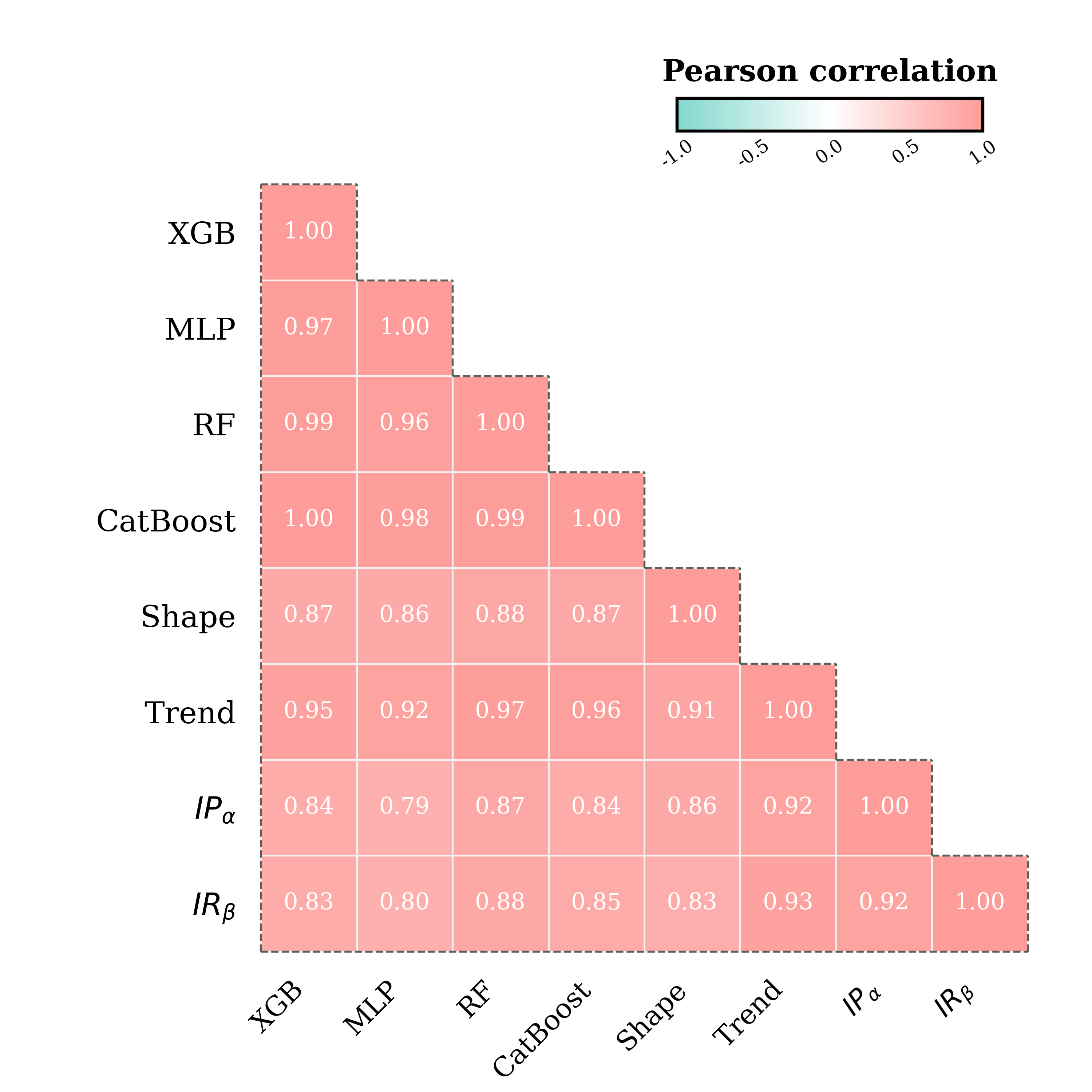}

    \end{subfigure}
    \hfill
    \begin{subfigure}[b]{0.32\linewidth}
        \centering
        \includegraphics[width=\linewidth]{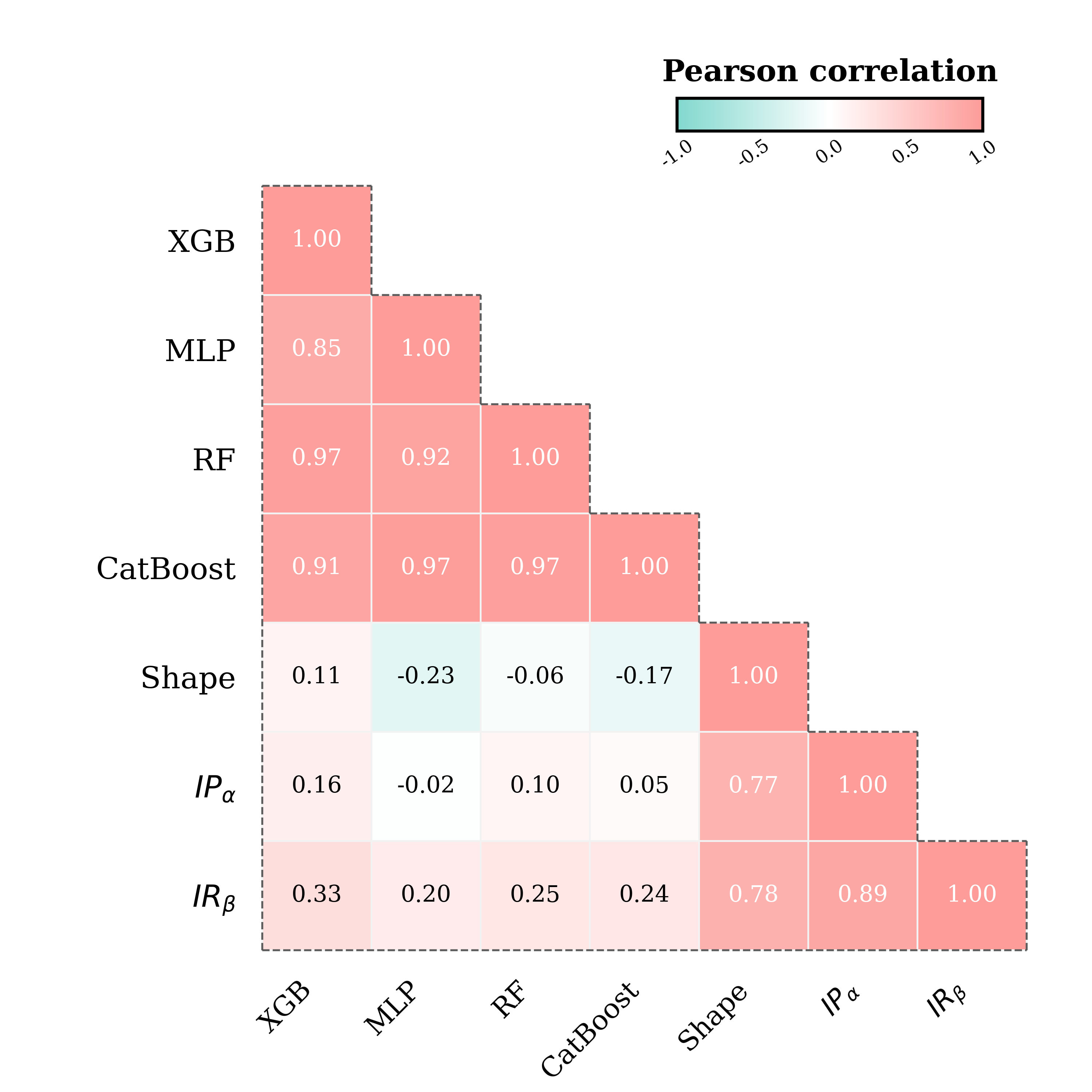}

    \end{subfigure} 
    
    \begin{subfigure}[b]{0.32\linewidth}
        \centering
        \includegraphics[width=\linewidth]{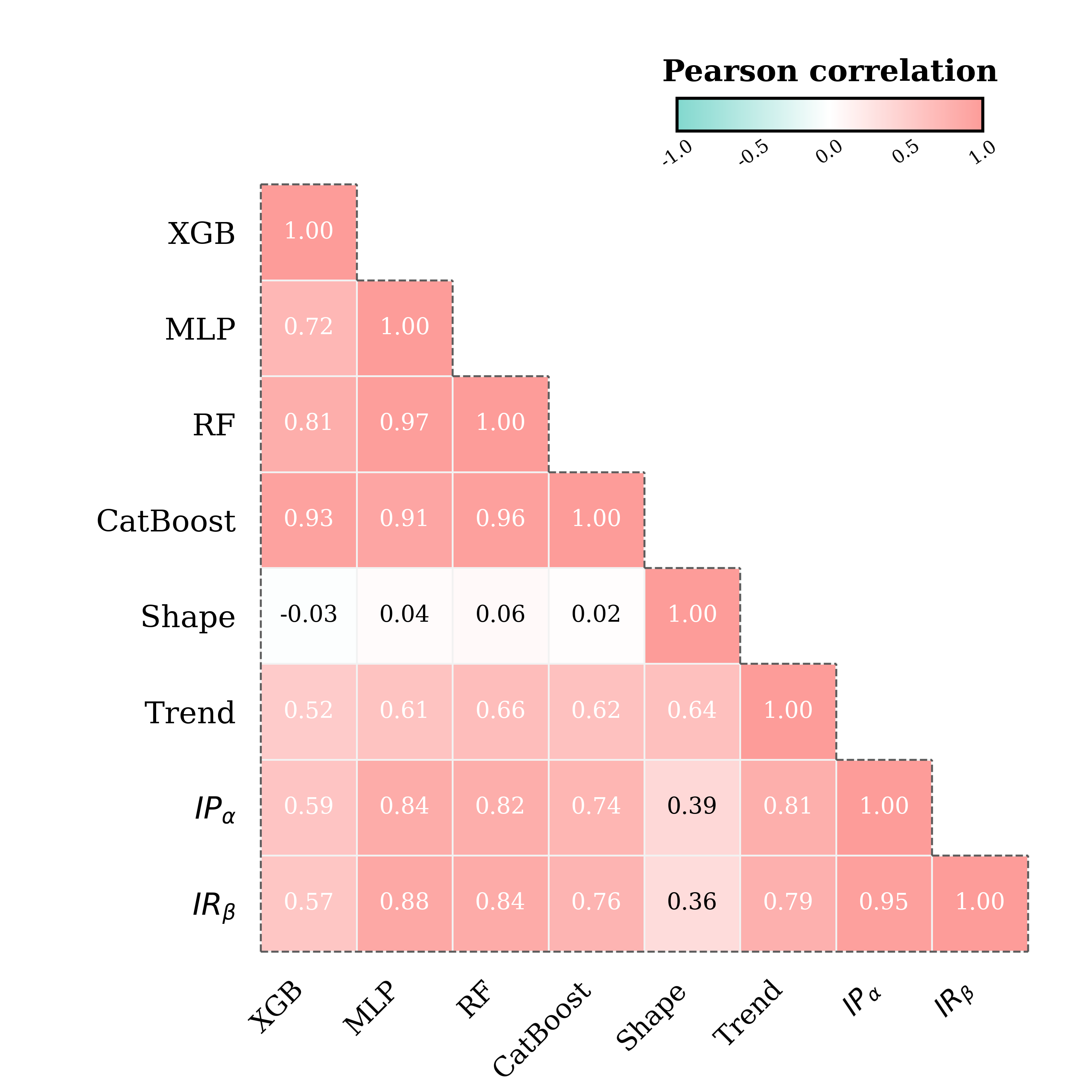}

    \end{subfigure}
    \hfill
    \begin{subfigure}[b]{0.32\linewidth}
        \centering
        \includegraphics[width=\linewidth]{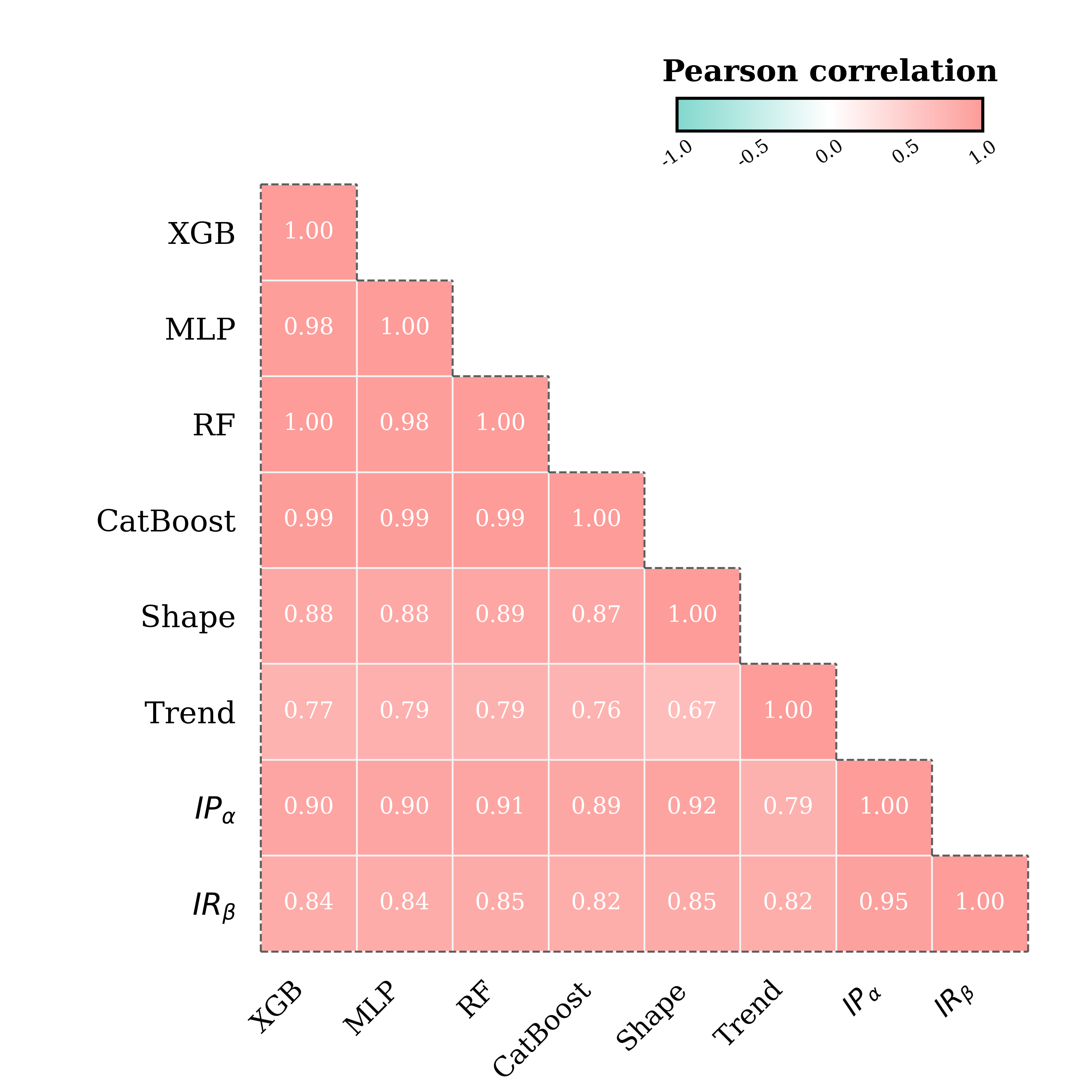}

    \end{subfigure}
    \hfill
    \begin{subfigure}[b]{0.32\linewidth}
        \centering
        \includegraphics[width=\linewidth]{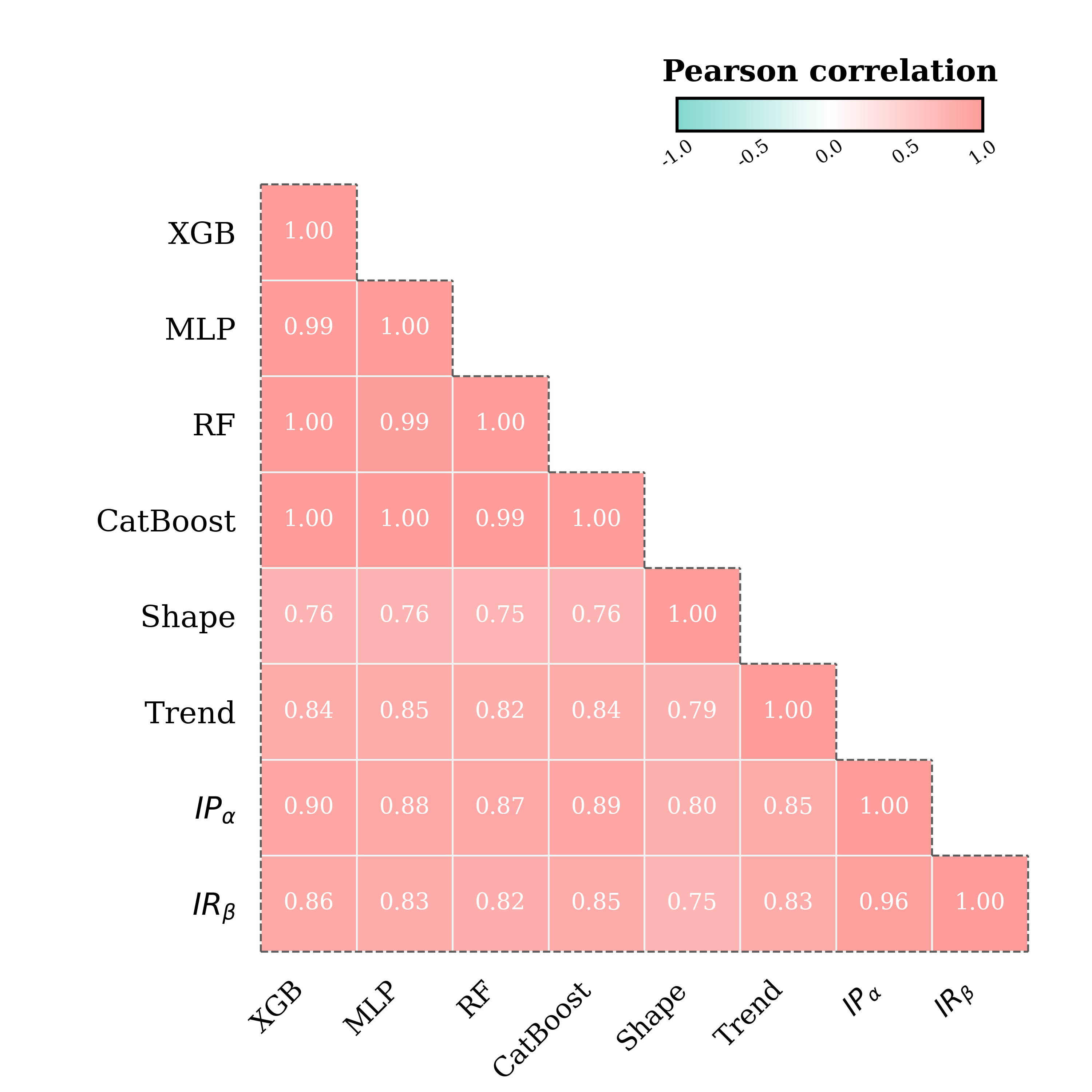}

    \end{subfigure} 

    \begin{subfigure}[b]{0.32\linewidth}
        \centering
        \includegraphics[width=\linewidth]{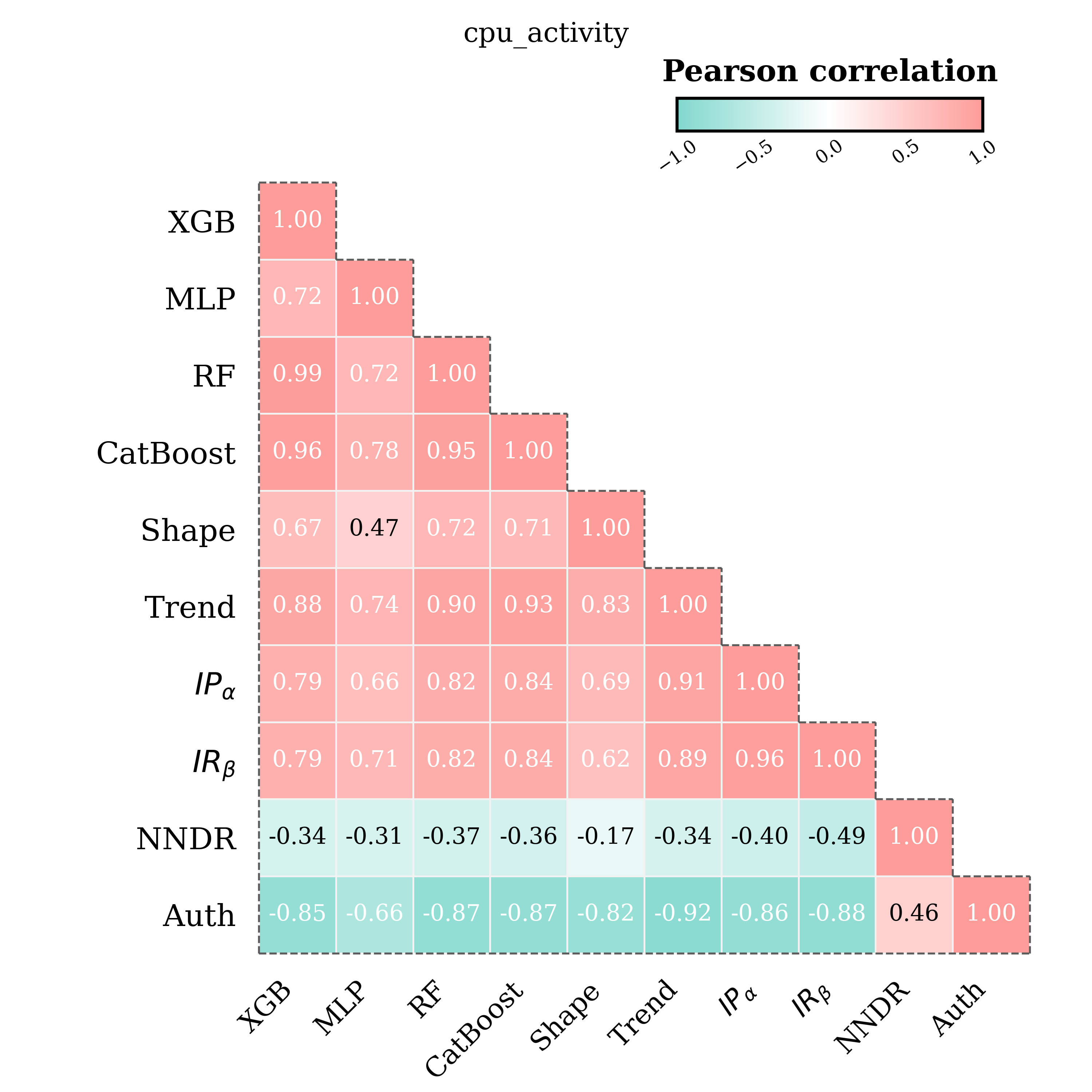}

    \end{subfigure}
    \hspace{2em}
    \begin{subfigure}[b]{0.32\linewidth}
        \centering
        \includegraphics[width=\linewidth]{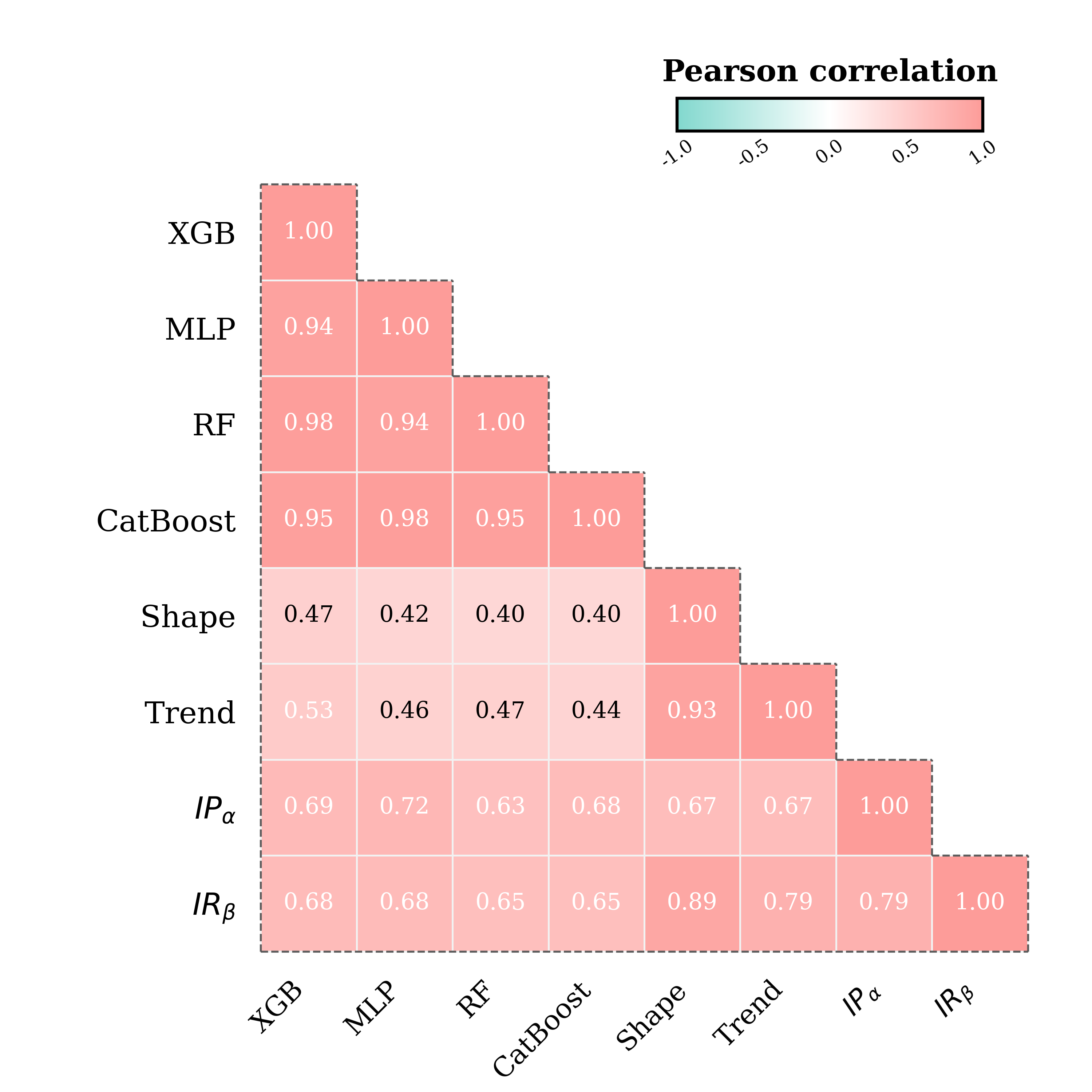}

    \end{subfigure}
    \caption{Correlation between quality metrics on each evaluation dataset.}
    \label{fig:correlation}
\end{figure}

The right panel of Figure~\ref{fig:correlation_metrics} shows the correlation matrix of quality metrics on the diabetes dataset. Here, we further report the corresponding correlation matrices for all 14 datasets in Figure~\ref{fig:correlation}. These results show that the disagreement between distribution-based quality metrics and downstream task performance is common across datasets. 

\subsection{Per-Dataset Results}
\label{app:per_dataset_results}

In the main experiments, Sections~\ref{sec:exp2}, \ref{sec:exp3}, and \ref{sec:exp4} report the average performance across multiple datasets to provide an overall comparison. In this subsection, we provide the detailed per-dataset results for all evaluated methods. Specifically, we include the baseline methods compared in Sections~\ref{sec:exp2} and \ref{sec:exp3}, as well as the ablation variants introduced in Section~\ref{sec:exp4}, namely DiffICL(S) and DiffICL(N). These results provide a more fine-grained view of the behavior of each method on individual datasets and complement the averaged results reported in the main text.
\begin{table*}[htbp]
\centering
\caption{Detailed performance comparison on the \textit{auction verification} dataset.
\small{We report the downstream task performance of baselines alongside the DiffICL and its two ablation variants: DiffICL(S) and DiffICL(N). Performance is evaluated across four learners under two scenarios: utilizing synthetic data as a replacement for real data (\textbf{Syn.}), and utilizing a combination of real and synthetic data for data augmentation (\textbf{Aug.}). The best results  are highlighted in \textcolor[HTML]{D93F49}{red}.}}
\label{tab:auction_performance}
\setlength{\tabcolsep}{4pt} 
\resizebox{\textwidth}{!}{
\begin{tabular}{lcccccccc}
\toprule
\multirow{2.5}{*}{\textbf{Method}} & \multicolumn{2}{c}{\textbf{XGB}} & \multicolumn{2}{c}{\textbf{CatB}} & \multicolumn{2}{c}{\textbf{RF}} & \multicolumn{2}{c}{\textbf{MLP}} \\
\cmidrule(lr){2-3} \cmidrule(lr){4-5} \cmidrule(lr){6-7} \cmidrule(lr){8-9}
& \textbf{Syn.} & \textbf{Aug.} & \textbf{Syn.} & \textbf{Aug.} & \textbf{Syn.} & \textbf{Aug.} & \textbf{Syn.} & \textbf{Aug.} \\
\midrule
CTGAN      & 15.0 \std{8.4}  & 71.6 \std{5.1}  & 21.2 \std{8.6}  & 72.6 \std{5.5}  & 7.9 \std{9.9}   & 50.5 \std{7.1}  & 11.8 \std{6.7}  & 55.3 \std{5.1}  \\
CTABGAN+   & 23.8 \std{17.7} & 96.0 \std{0.2}  & 58.7 \std{11.2} & 93.7 \std{0.9}  & 33.3 \std{10.5} & 95.2 \std{0.2}  & 58.2 \std{8.7}  & 79.5 \std{2.1}  \\
TVAE       & 69.5 \std{7.5}  & 87.1 \std{3.4}  & 74.8 \std{6.2}  & 87.6 \std{2.9}  & 65.8 \std{8.3}  & 77.7 \std{4.1}  & 62.5 \std{12.2} & 81.6 \std{3.4}  \\
tabDDPM    & 67.6 \std{4.7}  & 86.0 \std{2.8}  & 69.3 \std{4.0}  & 82.7 \std{2.6}  & 61.3 \std{3.8}  & 77.0 \std{3.2}  & 60.6 \std{5.0}  & 73.8 \std{1.8}  \\
GReaT      & 69.5 \std{10.9} & 89.7 \std{5.9}  & 79.1 \std{6.2}  & 92.3 \std{2.3}  & 64.3 \std{12.3} & 80.1 \std{9.8}  & 78.0 \std{2.2}  & 86.5 \std{0.7}  \\
TabSYN     & 74.2 \std{10.8} & 90.0 \std{2.4}  & 76.2 \std{12.0} & 88.2 \std{2.4}  & 71.7 \std{12.4} & 82.3 \std{3.8}  & 70.3 \std{12.6} & 80.6 \std{5.3}  \\
TabDiff    & 19.1 \std{4.8}  & 81.0 \std{1.6}  & 28.6 \std{4.6}  & 74.4 \std{1.4}  & 8.1 \std{9.3}   & 63.1 \std{2.2}  & 28.0 \std{2.4}  & 44.6 \std{1.8}  \\
\midrule
DiffICL    & \textcolor[HTML]{D93F49}{83.2 \std{3.4}} & 98.1 \std{0.2} & \textcolor[HTML]{D93F49}{88.6 \std{1.8}} & \textcolor[HTML]{D93F49}{96.9 \std{0.4}} & \textcolor[HTML]{D93F49}{83.2 \std{4.3}} & \textcolor[HTML]{D93F49}{95.8 \std{0.2}} & \textcolor[HTML]{D93F49}{83.7 \std{5.9}} & \textcolor[HTML]{D93F49}{92.1 \std{0.5}} \\
DiffICL(S) & 35.9 \std{7.1}  & 97.3 \std{0.1}  & 38.4 \std{5.9}  & 94.1 \std{0.4}  & 38.2 \std{6.4}  & 95.6 \std{0.3}  & 38.2 \std{6.2}  & 79.3 \std{5.5}  \\
DiffICL(N) & 74.3 \std{3.1}  & \textcolor[HTML]{D93F49}{98.2 \std{0.2}} & 83.1 \std{3.4} & \textcolor[HTML]{D93F49}{96.9 \std{0.3}} & 77.5 \std{4.1}  & 95.6 \std{0.4}  & 74.3 \std{15.2} & 91.3 \std{1.3}  \\
\bottomrule
\end{tabular}
}
\end{table*}
\vspace{-5pt}

\vspace{-5pt}
\begin{table*}[htbp]
\centering
\caption{Detailed performance comparison on the \textit{health insurance} dataset.
\small{We report the downstream task performance of baselines alongside the DiffICL and its two ablation variants: DiffICL(S) and DiffICL(N). Performance is evaluated across four learners under two scenarios: utilizing synthetic data as a replacement for real data (\textbf{Syn.}), and utilizing a combination of real and synthetic data for data augmentation (\textbf{Aug.}). The best results  are highlighted in \textcolor[HTML]{D93F49}{red}.}}
\label{tab:health_insurance_performance}
\setlength{\tabcolsep}{4pt} 
\resizebox{\textwidth}{!}{
\begin{tabular}{lcccccccc}
\toprule
\multirow{2.5}{*}{\textbf{Method}} & \multicolumn{2}{c}{\textbf{XGB}} & \multicolumn{2}{c}{\textbf{CatB}} & \multicolumn{2}{c}{\textbf{RF}} & \multicolumn{2}{c}{\textbf{MLP}} \\
\cmidrule(lr){2-3} \cmidrule(lr){4-5} \cmidrule(lr){6-7} \cmidrule(lr){8-9}
& \textbf{Syn.} & \textbf{Aug.} & \textbf{Syn.} & \textbf{Aug.} & \textbf{Syn.} & \textbf{Aug.} & \textbf{Syn.} & \textbf{Aug.} \\
\midrule
CTGAN      & 4.6 \std{6.5}   & 39.3 \std{0.1} & 16.9 \std{3.4} & 39.9 \std{0.1} & 9.3 \std{4.6}   & 31.6 \std{0.3} & 15.6 \std{4.4} & 38.7 \std{0.3} \\
CTABGAN+   & -16.2 \std{9.1} & \textcolor[HTML]{D93F49}{39.8 \std{0.1}} & -7.4 \std{8.7} & 40.3 \std{0.1} & -3.1 \std{4.4}  & 32.3 \std{0.1} & -6.8 \std{6.4} & 38.1 \std{0.3} \\
TVAE       & 19.2 \std{1.5}  & 39.1 \std{0.1} & 25.9 \std{2.4} & 40.5 \std{0.2} & 23.5 \std{2.2}  & 32.5 \std{0.1} & 18.0 \std{5.0} & 39.2 \std{0.3} \\
tabDDPM    & 4.0 \std{9.5}   & 39.4 \std{0.1} & 20.8 \std{8.2} & 40.6 \std{0.1} & 13.7 \std{6.5}  & 32.0 \std{0.3} & 18.6 \std{8.1} & 39.7 \std{0.2} \\
GReaT      & 27.0 \std{1.6}  & 39.0 \std{0.1} & 35.8 \std{0.9} & 40.4 \std{0.0} & 29.0 \std{0.7}  & 31.2 \std{0.4} & 34.9 \std{1.1} & 39.4 \std{0.2} \\
TabSYN     & 28.1 \std{1.5}  & 39.2 \std{0.1} & 35.6 \std{1.5} & 40.6 \std{0.2} & 28.5 \std{1.6}  & 31.8 \std{0.3} & 34.4 \std{1.6} & 39.6 \std{0.3} \\
TabDiff    & 24.4 \std{3.7}  & 39.3 \std{0.2} & 33.1 \std{2.7} & 40.4 \std{0.2} & 25.7 \std{3.4}  & 31.6 \std{0.5} & 31.7 \std{2.9} & 39.5 \std{0.3} \\
\midrule
DiffICL    & 38.5 \std{1.0} & 39.5 \std{0.1} & \textcolor[HTML]{D93F49}{39.1 \std{0.7}} & \textcolor[HTML]{D93F49}{40.8 \std{0.1}} & \textcolor[HTML]{D93F49}{38.1 \std{0.8}} & \textcolor[HTML]{D93F49}{33.0 \std{0.1}} & \textcolor[HTML]{D93F49}{38.5 \std{0.8}} & 39.8 \std{0.1} \\
DiffICL(S) & 21.5 \std{8.3} & \textcolor[HTML]{D93F49}{39.8 \std{0.2}} & 26.4 \std{4.2} & 40.5 \std{0.1} & 24.4 \std{5.6}  & 32.9 \std{0.1} & 26.2 \std{4.1} & 39.5 \std{0.1} \\
DiffICL(N) & \textcolor[HTML]{D93F49}{38.7 \std{0.5}} & 39.5 \std{0.1} & 38.9 \std{0.4} & \textcolor[HTML]{D93F49}{40.8 \std{0.1}} & 37.9 \std{0.3} & \textcolor[HTML]{D93F49}{33.0 \std{0.0}} & \textcolor[HTML]{D93F49}{38.5 \std{0.8}} & \textcolor[HTML]{D93F49}{39.9 \std{0.2}} \\
\bottomrule
\end{tabular}
}
\end{table*}

\begin{table*}[htbp]
\centering
\caption{Detailed performance comparison on the \textit{cpu activity} dataset.
\small{We report the downstream task performance of baselines alongside the DiffICL and its two ablation variants: DiffICL(S) and DiffICL(N). Performance is evaluated across four learners under two scenarios: utilizing synthetic data as a replacement for real data (\textbf{Syn.}), and utilizing a combination of real and synthetic data for data augmentation (\textbf{Aug.}). The best results  are highlighted in \textcolor[HTML]{D93F49}{red}.}}
\label{tab:cpu_activity_performance}
\setlength{\tabcolsep}{4pt} 
\resizebox{\textwidth}{!}{
\begin{tabular}{lcccccccc}
\toprule
\multirow{2.5}{*}{\textbf{Method}} & \multicolumn{2}{c}{\textbf{XGB}} & \multicolumn{2}{c}{\textbf{CatB}} & \multicolumn{2}{c}{\textbf{RF}} & \multicolumn{2}{c}{\textbf{MLP}} \\
\cmidrule(lr){2-3} \cmidrule(lr){4-5} \cmidrule(lr){6-7} \cmidrule(lr){8-9}
& \textbf{Syn.} & \textbf{Aug.} & \textbf{Syn.} & \textbf{Aug.} & \textbf{Syn.} & \textbf{Aug.} & \textbf{Syn.} & \textbf{Aug.} \\
\midrule
CTGAN      & 18.1 \std{4.5}  & 97.6 \std{0.2} & 18.1 \std{3.5} & 96.7 \std{0.4} & 21.6 \std{2.4}  & 97.5 \std{0.1} & 17.8 \std{5.3}  & 94.0 \std{0.6} \\
CTABGAN+   & -45.8 \std{64.9}& 98.1 \std{0.0} & -2.3 \std{24.4}& 97.4 \std{0.1} & -34.2 \std{54.4}& 97.8 \std{0.1} & 11.7 \std{12.5} & 95.8 \std{0.6} \\
TVAE       & 93.3 \std{1.9}  & 98.1 \std{0.1} & 93.9 \std{1.6} & 97.6 \std{0.1} & 93.7 \std{2.0}  & 97.9 \std{0.0} & 83.5 \std{2.6}  & 96.6 \std{0.3} \\
tabDDPM    & 90.3 \std{1.5}  & 97.4 \std{0.1} & 90.6 \std{0.7} & 97.1 \std{0.2} & 86.7 \std{3.1}  & 97.6 \std{0.0} & 73.8 \std{8.4}  & 95.4 \std{0.3} \\
GReaT      & 86.2 \std{6.9}  & 97.4 \std{0.4} & 84.3 \std{5.5} & 97.5 \std{0.1} & 91.4 \std{3.2}  & 97.5 \std{0.1} & 78.2 \std{2.3}  & 96.5 \std{0.5} \\
TabSYN     & 89.2 \std{1.9}  & 96.9 \std{0.7} & 89.1 \std{2.7} & 96.6 \std{0.6} & 91.1 \std{1.8}  & 97.3 \std{0.4} & 48.6 \std{57.8} & 95.1 \std{0.9} \\
TabDiff    & 82.4 \std{13.6} & 96.9 \std{0.3} & 86.3 \std{10.0}& 96.8 \std{0.3} & 83.5 \std{14.5} & 97.4 \std{0.2} & 75.2 \std{12.7} & 94.2 \std{1.0} \\
\midrule
DiffICL    & \textcolor[HTML]{D93F49}{96.8 \std{0.9}} & \textcolor[HTML]{D93F49}{98.5 \std{0.0}} & \textcolor[HTML]{D93F49}{96.0 \std{1.2}} & \textcolor[HTML]{D93F49}{98.1 \std{0.2}} & \textcolor[HTML]{D93F49}{96.0 \std{0.5}} & \textcolor[HTML]{D93F49}{98.0 \std{0.0}} & \textcolor[HTML]{D93F49}{90.3 \std{2.9}} & \textcolor[HTML]{D93F49}{97.4 \std{0.3}} \\
DiffICL(S) & 76.0 \std{19.7} & 98.2 \std{0.2} & 76.7 \std{17.6}& 97.6 \std{0.4} & 70.3 \std{15.4} & 97.9 \std{0.1} & 48.8 \std{27.7} & 96.7 \std{0.1} \\
DiffICL(N) & 96.2 \std{0.6}  & 98.3 \std{0.2} & 95.1 \std{0.9} & \textcolor[HTML]{D93F49}{98.1 \std{0.1}} & 94.4 \std{2.8}  & \textcolor[HTML]{D93F49}{98.0 \std{0.1}} & 88.6 \std{2.2}  & 97.3 \std{0.3} \\
\bottomrule
\end{tabular}
}
\end{table*}

\begin{table*}[htbp]
\centering
\caption{Detailed performance comparison on the \textit{wind} dataset.
\small{We report the downstream task performance of baselines alongside the DiffICL and its two ablation variants: DiffICL(S) and DiffICL(N). Performance is evaluated across four learners under two scenarios: utilizing synthetic data as a replacement for real data (\textbf{Syn.}), and utilizing a combination of real and synthetic data for data augmentation (\textbf{Aug.}). The best results  are highlighted in \textcolor[HTML]{D93F49}{red}.}}
\label{tab:wind_performance}
\setlength{\tabcolsep}{4pt} 
\resizebox{\textwidth}{!}{
\begin{tabular}{lcccccccc}
\toprule
\multirow{2.5}{*}{\textbf{Method}} & \multicolumn{2}{c}{\textbf{XGB}} & \multicolumn{2}{c}{\textbf{CatB}} & \multicolumn{2}{c}{\textbf{RF}} & \multicolumn{2}{c}{\textbf{MLP}} \\
\cmidrule(lr){2-3} \cmidrule(lr){4-5} \cmidrule(lr){6-7} \cmidrule(lr){8-9}
& \textbf{Syn.} & \textbf{Aug.} & \textbf{Syn.} & \textbf{Aug.} & \textbf{Syn.} & \textbf{Aug.} & \textbf{Syn.} & \textbf{Aug.} \\
\midrule
CTGAN      & 41.7 \std{18.6} & 76.0 \std{0.9} & 44.6 \std{18.7} & 74.8 \std{1.6} & 44.1 \std{20.0} & 72.8 \std{2.1} & 47.4 \std{20.4} & 75.4 \std{1.1} \\
CTABGAN+   & 8.3 \std{27.1}  & 76.0 \std{0.7} & 15.6 \std{23.9} & 74.8 \std{1.1} & 10.9 \std{30.3} & 73.6 \std{0.7} & 17.7 \std{22.8} & 75.8 \std{0.9} \\
TVAE       & 65.0 \std{1.8}  & 77.2 \std{0.3} & 66.8 \std{1.6}  & 77.1 \std{0.2} & 65.5 \std{1.6}  & 74.8 \std{0.2} & 67.0 \std{2.4}  & 77.1 \std{0.1} \\
tabDDPM    & 74.9 \std{0.7}  & 78.0 \std{0.2} & 75.9 \std{0.3}  & 77.9 \std{0.1} & \textcolor[HTML]{D93F49}{73.9 \std{0.5}} & 76.2 \std{0.1} & 75.6 \std{0.6}  & 77.8 \std{0.4} \\
GReaT      & 73.6 \std{0.4}  & 76.2 \std{0.1} & 74.9 \std{0.3}  & 76.3 \std{0.1} & 71.9 \std{0.3}  & 74.0 \std{0.1} & 75.4 \std{0.4}  & 77.1 \std{0.2} \\
TabSYN     & 74.6 \std{0.4}  & 77.8 \std{0.2} & 75.8 \std{0.5}  & 78.1 \std{0.1} & 73.7 \std{0.3}  & 76.0 \std{0.3} & 76.2 \std{0.5}  & 78.5 \std{0.2} \\
TabDiff    & 74.1 \std{0.6}  & 77.5 \std{0.1} & 75.6 \std{0.5}  & 77.8 \std{0.1} & 73.5 \std{0.3}  & 76.0 \std{0.2} & 76.1 \std{0.5}  & 78.0 \std{0.2} \\
\midrule
DiffICL    & \textcolor[HTML]{D93F49}{75.7 \std{0.5}} & \textcolor[HTML]{D93F49}{78.4 \std{0.2}} & 76.1 \std{0.8} & \textcolor[HTML]{D93F49}{78.3 \std{0.1}} & 73.7 \std{0.7} & \textcolor[HTML]{D93F49}{76.5 \std{0.3}} & \textcolor[HTML]{D93F49}{76.6 \std{0.5}} & 78.6 \std{0.1} \\
DiffICL(S) & 65.0 \std{6.3}  & 77.4 \std{0.3} & 66.9 \std{5.7}  & 77.1 \std{0.5} & 65.3 \std{5.8}  & 75.1 \std{0.3} & 67.3 \std{4.7}  & 77.3 \std{0.6} \\
DiffICL(N) & \textcolor[HTML]{D93F49}{75.7 \std{0.7}} & 78.2 \std{0.1} & \textcolor[HTML]{D93F49}{76.5 \std{0.6}} & \textcolor[HTML]{D93F49}{78.3 \std{0.1}} & 73.8 \std{0.7} & 76.3 \std{0.2} & 76.3 \std{0.7} & \textcolor[HTML]{D93F49}{78.7 \std{0.2}} \\
\bottomrule
\end{tabular}
}
\end{table*}

\begin{table*}[htbp]
\centering
\caption{Detailed performance comparison on the \textit{white wine} dataset. 
\small{We report the downstream task performance of baselines alongside the DiffICL and its two ablation variants: DiffICL(S) and DiffICL(N). Performance is evaluated across four learners under two scenarios: utilizing synthetic data as a replacement for real data (\textbf{Syn.}), and utilizing a combination of real and synthetic data for data augmentation (\textbf{Aug.}). The best results  are highlighted in \textcolor[HTML]{D93F49}{red}.}}
\label{tab:white_wine_performance}
\setlength{\tabcolsep}{4pt} 
\resizebox{\textwidth}{!}{
\begin{tabular}{lcccccccc}
\toprule
\multirow{2.5}{*}{\textbf{Method}} & \multicolumn{2}{c}{\textbf{XGB}} & \multicolumn{2}{c}{\textbf{CatB}} & \multicolumn{2}{c}{\textbf{RF}} & \multicolumn{2}{c}{\textbf{MLP}} \\
\cmidrule(lr){2-3} \cmidrule(lr){4-5} \cmidrule(lr){6-7} \cmidrule(lr){8-9}
& \textbf{Syn.} & \textbf{Aug.} & \textbf{Syn.} & \textbf{Aug.} & \textbf{Syn.} & \textbf{Aug.} & \textbf{Syn.} & \textbf{Aug.} \\
\midrule
CTGAN      & -21.7 \std{21.0} & 41.4 \std{1.4} & -13.5 \std{20.2} & 36.9 \std{1.6} & -12.2 \std{16.2} & 44.8 \std{1.6} & -13.0 \std{15.9} & 34.1 \std{1.6} \\
CTABGAN+   & -9.2 \std{11.1}  & 41.7 \std{1.0} & -1.3 \std{13.2}  & 37.3 \std{1.3} & 0.1 \std{10.2}   & 45.2 \std{0.9} & 1.4 \std{8.6}    & 32.0 \std{1.5} \\
TVAE       & 26.5 \std{1.8}   & 43.5 \std{0.4} & 27.2 \std{1.4}   & 38.3 \std{0.3} & 29.2 \std{1.3}   & 46.5 \std{0.3} & 24.3 \std{1.8}   & 35.1 \std{0.6} \\
tabDDPM    & 29.5 \std{0.8}   & 43.5 \std{0.6} & \textcolor[HTML]{D93F49}{32.7 \std{1.0}} & 39.1 \std{0.3} & 31.3 \std{0.8} & 46.8 \std{0.3} & \textcolor[HTML]{D93F49}{31.1 \std{0.7}} & 36.0 \std{0.5} \\
GReaT      & 26.1 \std{2.0}   & 41.9 \std{0.2} & 29.4 \std{1.2}   & 37.3 \std{0.2} & 28.4 \std{1.0}   & 44.7 \std{0.2} & 28.0 \std{1.5}   & 33.7 \std{0.9} \\
TabSYN     & 29.2 \std{1.8}   & 42.9 \std{0.4} & 32.2 \std{1.3}   & 38.8 \std{0.1} & \textcolor[HTML]{D93F49}{31.6 \std{0.9}} & 46.5 \std{0.5} & 30.7 \std{1.2}   & 35.6 \std{0.3} \\
TabDiff    & 26.9 \std{1.9}   & 42.8 \std{0.9} & 30.6 \std{1.2}   & 38.6 \std{0.7} & 29.4 \std{1.3}   & 46.6 \std{0.5} & 30.5 \std{1.3}   & 35.9 \std{0.5} \\
\midrule
DiffICL    & \textcolor[HTML]{D93F49}{30.6 \std{2.4}} & 44.3 \std{0.4} & 31.6 \std{2.0} & 39.6 \std{0.4} & 31.3 \std{1.7} & \textcolor[HTML]{D93F49}{47.2 \std{0.3}} & 30.2 \std{2.0} & \textcolor[HTML]{D93F49}{37.8 \std{0.6}} \\
DiffICL(S) & 6.7 \std{4.6}    & 41.9 \std{0.7} & 13.9 \std{3.3}   & 38.1 \std{0.9} & 14.8 \std{2.1}   & 46.0 \std{0.4} & 14.2 \std{2.4}   & 34.4 \std{0.6} \\
DiffICL(N) & 29.7 \std{2.9}   & \textcolor[HTML]{D93F49}{44.5 \std{0.5}} & 31.2 \std{2.5} & \textcolor[HTML]{D93F49}{39.9 \std{0.4}} & 30.4 \std{2.5} & 47.1 \std{0.3} & 29.6 \std{2.4} & \textcolor[HTML]{D93F49}{37.8 \std{0.6}} \\
\bottomrule
\end{tabular}
}
\end{table*}
\begin{table*}[htbp]
\centering
\caption{Detailed performance comparison on the \textit{abalone} dataset. 
\small{We report the downstream task performance of baselines alongside the DiffICL and its two ablation variants: DiffICL(S) and DiffICL(N). Performance is evaluated across four learners under two scenarios: utilizing synthetic data as a replacement for real data (\textbf{Syn.}), and utilizing a combination of real and synthetic data for data augmentation (\textbf{Aug.}). The best results  are highlighted in \textcolor[HTML]{D93F49}{red}.}}
\label{tab:abalone_performance}
\setlength{\tabcolsep}{4pt} 
\resizebox{\textwidth}{!}{
\begin{tabular}{lcccccccc}
\toprule
\multirow{2.5}{*}{\textbf{Method}} & \multicolumn{2}{c}{\textbf{XGB}} & \multicolumn{2}{c}{\textbf{CatB}} & \multicolumn{2}{c}{\textbf{RF}} & \multicolumn{2}{c}{\textbf{MLP}} \\
\cmidrule(lr){2-3} \cmidrule(lr){4-5} \cmidrule(lr){6-7} \cmidrule(lr){8-9}
& \textbf{Syn.} & \textbf{Aug.} & \textbf{Syn.} & \textbf{Aug.} & \textbf{Syn.} & \textbf{Aug.} & \textbf{Syn.} & \textbf{Aug.} \\
\midrule
CTGAN      & 16.9 \std{10.5} & 52.3 \std{0.2} & 21.2 \std{8.8} & 50.0 \std{0.9} & 19.9 \std{10.1} & 51.9 \std{0.5} & 26.4 \std{8.1} & 53.1 \std{1.6} \\
CTABGAN+   & 16.0 \std{4.8}  & 52.6 \std{0.3} & 25.1 \std{2.5} & 50.3 \std{0.7} & 22.0 \std{2.8}  & 52.9 \std{0.4} & 28.5 \std{1.0} & 52.4 \std{0.8} \\
TVAE       & 43.3 \std{2.3}  & 53.0 \std{0.4} & 45.2 \std{2.1} & 53.6 \std{0.5} & 44.6 \std{2.1}  & 53.6 \std{0.3} & 45.7 \std{2.0} & 54.7 \std{0.2} \\
tabDDPM    & 47.7 \std{1.6}  & 53.3 \std{0.3} & 50.1 \std{0.8} & 54.9 \std{0.4} & 49.5 \std{1.5}  & 54.6 \std{0.4} & 52.1 \std{1.4} & 55.8 \std{0.6} \\
GReaT      & 34.3 \std{37.1} & 52.0 \std{0.6} & 44.4 \std{14.4}& 53.2 \std{0.5} & 48.5 \std{4.8}  & 52.9 \std{0.8} & 53.7 \std{0.8} & 55.3 \std{0.4} \\
TabSYN     & 50.7 \std{2.1}  & 54.1 \std{1.2} & 53.4 \std{2.1} & 55.9 \std{0.8} & 53.0 \std{2.0}  & 55.7 \std{0.4} & 52.8 \std{2.1} & 56.9 \std{0.5} \\
TabDiff    & 29.0 \std{8.5}  & 52.9 \std{0.7} & 35.4 \std{6.2} & 51.8 \std{1.4} & 34.3 \std{6.2}  & 53.0 \std{0.5} & 34.8 \std{7.7} & 53.2 \std{0.9} \\
\midrule
DiffICL    & \textcolor[HTML]{D93F49}{56.0 \std{1.3}} & \textcolor[HTML]{D93F49}{56.1 \std{0.6}} & \textcolor[HTML]{D93F49}{55.2 \std{1.1}} & \textcolor[HTML]{D93F49}{56.3 \std{0.3}} & \textcolor[HTML]{D93F49}{54.8 \std{0.6}} & \textcolor[HTML]{D93F49}{56.3 \std{0.4}} & \textcolor[HTML]{D93F49}{56.6 \std{1.4}} & \textcolor[HTML]{D93F49}{58.2 \std{0.3}} \\
DiffICL(S) & 27.0 \std{14.8} & 51.7 \std{1.5} & 29.7 \std{15.8}& 51.8 \std{2.1} & 28.2 \std{15.9} & 51.9 \std{1.8} & 30.4 \std{16.8}& 53.8 \std{2.3} \\
DiffICL(N) & 53.9 \std{1.1}  & 55.7 \std{0.4} & 53.5 \std{1.1} & 56.0 \std{0.5} & 53.4 \std{1.1}  & 55.9 \std{0.5} & 54.4 \std{1.3} & 57.4 \std{0.5} \\
\bottomrule
\end{tabular}
}
\end{table*}
\begin{table*}[htbp]
\centering
\caption{Detailed performance comparison on the \textit{red wine} dataset. 
\small{We report the downstream task performance of baselines alongside the DiffICL and its two ablation variants: DiffICL(S) and DiffICL(N). Performance is evaluated across four learners under two scenarios: utilizing synthetic data as a replacement for real data (\textbf{Syn.}), and utilizing a combination of real and synthetic data for data augmentation (\textbf{Aug.}). The best results  are highlighted in \textcolor[HTML]{D93F49}{red}.}}
\label{tab:red_wine_performance}
\setlength{\tabcolsep}{4pt} 
\resizebox{\textwidth}{!}{
\begin{tabular}{lcccccccc}
\toprule
\multirow{2.5}{*}{\textbf{Method}} & \multicolumn{2}{c}{\textbf{XGB}} & \multicolumn{2}{c}{\textbf{CatB}} & \multicolumn{2}{c}{\textbf{RF}} & \multicolumn{2}{c}{\textbf{MLP}} \\
\cmidrule(lr){2-3} \cmidrule(lr){4-5} \cmidrule(lr){6-7} \cmidrule(lr){8-9}
& \textbf{Syn.} & \textbf{Aug.} & \textbf{Syn.} & \textbf{Aug.} & \textbf{Syn.} & \textbf{Aug.} & \textbf{Syn.} & \textbf{Aug.} \\
\midrule
CTGAN      & -3.4 \std{4.0}   & 37.2 \std{2.6} & 1.6 \std{3.9}  & 34.3 \std{3.9} & 0.8 \std{2.0}   & 38.3 \std{2.8} & 1.7 \std{3.3}  & 29.2 \std{5.6} \\
CTABGAN+   & -24.9 \std{5.9}  & 39.9 \std{1.6} & -15.8 \std{4.1}& 38.0 \std{0.9} & -9.4 \std{3.9}  & 41.5 \std{1.3} & -18.1 \std{10.4}& 32.9 \std{1.0} \\
TVAE       & 32.0 \std{1.3}   & 43.5 \std{1.2} & 35.1 \std{1.4} & 41.6 \std{0.9} & 35.4 \std{1.3}  & 44.9 \std{0.9} & 32.1 \std{2.1} & 35.0 \std{1.1} \\
tabDDPM    & 33.6 \std{1.4}   & 42.2 \std{1.3} & \textcolor[HTML]{D93F49}{37.3 \std{0.9}} & 40.5 \std{1.1} & 36.1 \std{1.5} & 44.9 \std{1.1} & \textcolor[HTML]{D93F49}{33.8 \std{1.1}} & 36.1 \std{1.5} \\
TabSYN     & 34.3 \std{1.7}   & 42.6 \std{1.9} & 36.4 \std{1.5} & 40.7 \std{1.7} & \textcolor[HTML]{D93F49}{37.1 \std{1.4}} & 44.9 \std{1.9} & 26.8 \std{2.6} & 29.4 \std{2.8} \\
TabDiff    & 31.2 \std{0.5}   & 41.5 \std{1.9} & 35.0 \std{0.5} & 39.8 \std{1.1} & 34.4 \std{0.8}  & 43.4 \std{1.4} & 33.2 \std{1.1} & 34.7 \std{1.1} \\
\midrule
DiffICL    & 34.0 \std{2.3}   & \textcolor[HTML]{D93F49}{44.8 \std{0.8}} & 35.1 \std{2.1} & \textcolor[HTML]{D93F49}{41.9 \std{0.6}} & 34.5 \std{1.7} & \textcolor[HTML]{D93F49}{45.2 \std{0.8}} & 29.4 \std{2.5} & 35.9 \std{1.3} \\
DiffICL(S) & 7.1 \std{13.0}   & 38.7 \std{4.3} & 11.0 \std{11.0}& 36.4 \std{5.4} & 13.0 \std{7.0}  & 40.4 \std{4.4} & 1.5 \std{20.1} & 29.4 \std{6.4} \\
DiffICL(N) & \textcolor[HTML]{D93F49}{35.0 \std{4.2}} & \textcolor[HTML]{D93F49}{44.8 \std{1.6}} & 35.7 \std{3.7} & 41.4 \std{1.9} & 34.3 \std{3.1} & 44.8 \std{1.3} & 33.4 \std{4.2} & \textcolor[HTML]{D93F49}{36.5 \std{2.3}} \\
\bottomrule
\end{tabular}
}
\end{table*}
\begin{table*}[htbp]
\centering
\caption{Detailed performance comparison on the \textit{mfeat-zernike} dataset.
\small{We report the downstream task performance of baselines alongside the DiffICL and its two ablation variants: DiffICL(S) and DiffICL(N). Performance is evaluated across four learners under two scenarios: utilizing synthetic data as a replacement for real data (\textbf{Syn.}), and utilizing a combination of real and synthetic data for data augmentation (\textbf{Aug.}). The best results  are highlighted in \textcolor[HTML]{D93F49}{red}.}}
\label{tab:mfeat_zernike_performance}
\setlength{\tabcolsep}{4pt} 
\resizebox{\textwidth}{!}{
\begin{tabular}{lcccccccc}
\toprule
\multirow{2.5}{*}{\textbf{Method}} & \multicolumn{2}{c}{\textbf{XGB}} & \multicolumn{2}{c}{\textbf{CatB}} & \multicolumn{2}{c}{\textbf{RF}} & \multicolumn{2}{c}{\textbf{MLP}} \\
\cmidrule(lr){2-3} \cmidrule(lr){4-5} \cmidrule(lr){6-7} \cmidrule(lr){8-9}
& \textbf{Syn.} & \textbf{Aug.} & \textbf{Syn.} & \textbf{Aug.} & \textbf{Syn.} & \textbf{Aug.} & \textbf{Syn.} & \textbf{Aug.} \\
\midrule
CTGAN      & 56.6 \std{3.5} & 96.7 \std{0.2} & 57.4 \std{3.4} & 97.2 \std{0.1} & 55.2 \std{3.8} & 96.5 \std{0.1} & 56.9 \std{3.3} & 97.6 \std{0.2} \\
CTABGAN+   & 66.7 \std{8.5} & 96.5 \std{0.1} & 71.2 \std{7.7} & 97.2 \std{0.1} & 70.9 \std{7.6} & 96.2 \std{0.2} & 72.6 \std{8.1} & 97.5 \std{0.3} \\
TVAE       & 96.7 \std{0.3} & 96.9 \std{0.1} & 96.8 \std{0.1} & 97.4 \std{0.1} & 96.3 \std{0.2} & 96.4 \std{0.1} & 97.0 \std{0.2} & 97.6 \std{0.1} \\
tabDDPM    & 65.6 \std{2.4} & 96.5 \std{0.1} & 78.4 \std{2.1} & 97.4 \std{0.1} & 75.6 \std{2.4} & 96.4 \std{0.0} & 86.3 \std{1.1} & 97.7 \std{0.1} \\
TabSYN     & \textcolor[HTML]{D93F49}{97.3 \std{0.5}} & \textcolor[HTML]{D93F49}{97.2 \std{0.1}} & \textcolor[HTML]{D93F49}{97.4 \std{0.4}} & \textcolor[HTML]{D93F49}{97.6 \std{0.1}} & \textcolor[HTML]{D93F49}{97.1 \std{0.3}} & \textcolor[HTML]{D93F49}{96.9 \std{0.1}} & \textcolor[HTML]{D93F49}{97.3 \std{0.5}} & \textcolor[HTML]{D93F49}{97.8 \std{0.1}} \\
TabDiff    & 95.7 \std{0.8} & 96.9 \std{0.1} & 95.9 \std{0.6} & 97.5 \std{0.1} & 95.3 \std{0.8} & 96.6 \std{0.0} & 95.8 \std{0.7} & 97.7 \std{0.2} \\
\midrule
DiffICL    & 96.4 \std{0.2} & 97.0 \std{0.1} & 96.8 \std{0.2} & 97.4 \std{0.1} & 96.4 \std{0.3} & 96.6 \std{0.1} & 97.0 \std{0.3} & \textcolor[HTML]{D93F49}{97.8 \std{0.1}} \\
DiffICL(S) & 94.4 \std{1.1} & 96.8 \std{0.1} & 95.2 \std{0.7} & 97.4 \std{0.1} & 94.7 \std{0.6} & 96.4 \std{0.2} & 95.0 \std{0.7} & \textcolor[HTML]{D93F49}{97.8 \std{0.1}} \\
DiffICL(N) & 96.1 \std{0.5} & 96.9 \std{0.1} & 96.3 \std{0.4} & 97.4 \std{0.1} & 96.0 \std{0.3} & 96.6 \std{0.1} & 96.6 \std{0.5} & \textcolor[HTML]{D93F49}{97.8 \std{0.1}} \\
\bottomrule
\end{tabular}
}
\end{table*}
\begin{table*}[htbp]
\centering
\caption{Detailed performance comparison on the \textit{diabetes} dataset. 
\small{We report the downstream task performance of baselines alongside the DiffICL and its two ablation variants: DiffICL(S) and DiffICL(N). Performance is evaluated across four learners under two scenarios: utilizing synthetic data as a replacement for real data (\textbf{Syn.}), and utilizing a combination of real and synthetic data for data augmentation (\textbf{Aug.}). The best results  are highlighted in \textcolor[HTML]{D93F49}{red}.}}
\label{tab:diabetes_performance}
\setlength{\tabcolsep}{4pt} 
\resizebox{\textwidth}{!}{
\begin{tabular}{lcccccccc}
\toprule
\multirow{2.5}{*}{\textbf{Method}} & \multicolumn{2}{c}{\textbf{XGB}} & \multicolumn{2}{c}{\textbf{CatB}} & \multicolumn{2}{c}{\textbf{RF}} & \multicolumn{2}{c}{\textbf{MLP}} \\
\cmidrule(lr){2-3} \cmidrule(lr){4-5} \cmidrule(lr){6-7} \cmidrule(lr){8-9}
& \textbf{Syn.} & \textbf{Aug.} & \textbf{Syn.} & \textbf{Aug.} & \textbf{Syn.} & \textbf{Aug.} & \textbf{Syn.} & \textbf{Aug.} \\
\midrule
CTGAN      & 80.2 \std{2.9} & 81.4 \std{2.4} & 81.1 \std{2.8} & 81.5 \std{2.5} & 80.4 \std{2.6} & 81.2 \std{2.2} & 81.5 \std{1.9} & 81.8 \std{1.9} \\
CTABGAN+   & 80.9 \std{1.4} & \textcolor[HTML]{D93F49}{81.8 \std{0.9}} & 81.7 \std{1.3} & \textcolor[HTML]{D93F49}{82.9 \std{0.7}} & 80.5 \std{1.2} & \textcolor[HTML]{D93F49}{82.3 \std{0.6}} & 82.2 \std{1.2} & 83.2 \std{0.6} \\
TVAE       & 80.6 \std{2.0} & 81.3 \std{1.6} & 81.3 \std{1.7} & 81.7 \std{1.9} & 81.3 \std{1.5} & 81.5 \std{1.1} & 81.6 \std{2.5} & 82.5 \std{1.4} \\
tabDDPM    & 80.6 \std{0.6} & 81.5 \std{1.4} & 80.6 \std{0.8} & 81.6 \std{1.3} & 81.2 \std{0.6} & 81.7 \std{1.2} & 80.5 \std{1.4} & 81.0 \std{1.0} \\
GReaT      & 79.5 \std{2.2} & 79.7 \std{1.6} & 81.2 \std{2.1} & 81.4 \std{1.8} & 81.2 \std{2.1} & 81.4 \std{1.6} & 83.5 \std{1.1} & \textcolor[HTML]{D93F49}{83.5 \std{0.9}} \\
TabSYN     & 79.7 \std{1.1} & 80.1 \std{1.2} & 80.2 \std{0.9} & 80.5 \std{1.3} & 81.0 \std{0.8} & 81.5 \std{0.8} & 79.8 \std{0.8} & 79.4 \std{1.8} \\
TabDiff    & \textcolor[HTML]{D93F49}{82.5 \std{0.9}} & 81.3 \std{0.5} & \textcolor[HTML]{D93F49}{83.2 \std{0.5}} & 82.2 \std{0.5} & \textcolor[HTML]{D93F49}{82.9 \std{0.6}} & 81.9 \std{0.3} & \textcolor[HTML]{D93F49}{83.8 \std{0.5}} & 83.4 \std{0.6} \\
\midrule
DiffICL    & 82.0 \std{1.3} & 81.7 \std{0.8} & 82.5 \std{1.5} & 82.3 \std{1.1} & 81.4 \std{1.6} & 81.7 \std{1.0} & 82.8 \std{1.4} & 82.9 \std{1.5} \\
DiffICL(S) & 75.4 \std{9.6} & 80.5 \std{1.4} & 75.9 \std{9.4} & 80.7 \std{1.7} & 75.4 \std{7.8} & 80.6 \std{1.0} & 79.8 \std{3.1} & 81.5 \std{1.6} \\
DiffICL(N) & 79.0 \std{3.5} & 79.6 \std{2.1} & 79.2 \std{3.4} & 79.8 \std{2.2} & 78.5 \std{2.7} & 80.2 \std{1.6} & 80.7 \std{3.2} & 81.5 \std{2.3} \\
\bottomrule
\end{tabular}
}
\end{table*}
\begin{table*}[htbp]
\centering
\caption{Detailed performance comparison on the \textit{vehicle} dataset. 
\small{We report the downstream task performance of baselines alongside the DiffICL and its two ablation variants: DiffICL(S) and DiffICL(N). Performance is evaluated across four learners under two scenarios: utilizing synthetic data as a replacement for real data (\textbf{Syn.}), and utilizing a combination of real and synthetic data for data augmentation (\textbf{Aug.}). The best results  are highlighted in \textcolor[HTML]{D93F49}{red}.}}
\label{tab:vehicle_performance}
\setlength{\tabcolsep}{4pt} 
\resizebox{\textwidth}{!}{
\begin{tabular}{lcccccccc}
\toprule
\multirow{2.5}{*}{\textbf{Method}} & \multicolumn{2}{c}{\textbf{XGB}} & \multicolumn{2}{c}{\textbf{CatB}} & \multicolumn{2}{c}{\textbf{RF}} & \multicolumn{2}{c}{\textbf{MLP}} \\
\cmidrule(lr){2-3} \cmidrule(lr){4-5} \cmidrule(lr){6-7} \cmidrule(lr){8-9}
& \textbf{Syn.} & \textbf{Aug.} & \textbf{Syn.} & \textbf{Aug.} & \textbf{Syn.} & \textbf{Aug.} & \textbf{Syn.} & \textbf{Aug.} \\
\midrule
CTGAN      & 81.6 \std{1.7} & 91.2 \std{0.7} & 81.3 \std{2.0} & 89.4 \std{1.3} & 80.7 \std{2.4} & 90.5 \std{0.7} & 82.8 \std{1.9} & 91.1 \std{1.3} \\
CTABGAN+   & 89.9 \std{0.4} & 92.9 \std{0.4} & 90.7 \std{0.4} & 92.3 \std{0.5} & 89.7 \std{0.5} & 92.3 \std{0.3} & 90.3 \std{0.4} & 92.8 \std{0.7} \\
TVAE       & 90.6 \std{1.0} & 92.7 \std{0.4} & 91.2 \std{0.6} & 92.7 \std{0.4} & 90.3 \std{0.7} & 92.4 \std{0.2} & 91.5 \std{1.1} & 93.0 \std{0.5} \\
tabDDPM    & 92.7 \std{0.5} & 94.1 \std{0.3} & 92.4 \std{0.7} & 93.6 \std{0.4} & 92.5 \std{0.6} & 93.8 \std{0.3} & 94.5 \std{0.1} & 95.3 \std{0.3} \\
TabSYN     & \textcolor[HTML]{D93F49}{94.3 \std{0.6}} & \textcolor[HTML]{D93F49}{94.7 \std{0.4}} & \textcolor[HTML]{D93F49}{94.0 \std{0.4}} & \textcolor[HTML]{D93F49}{94.5 \std{0.4}} & \textcolor[HTML]{D93F49}{93.6 \std{0.6}} & \textcolor[HTML]{D93F49}{93.9 \std{0.4}} & \textcolor[HTML]{D93F49}{95.4 \std{0.3}} & \textcolor[HTML]{D93F49}{95.9 \std{0.5}} \\
TabDiff    & 91.1 \std{1.0} & 93.7 \std{0.4} & 91.4 \std{1.0} & 93.4 \std{0.4} & 89.9 \std{1.1} & 93.1 \std{0.3} & 90.7 \std{1.4} & 94.4 \std{0.4} \\
\midrule
DiffICL    & 93.2 \std{0.6} & 94.5 \std{0.6} & 93.0 \std{0.7} & 93.9 \std{0.6} & 92.1 \std{0.9} & 93.5 \std{0.3} & 94.4 \std{0.9} & 95.4 \std{0.9} \\
DiffICL(S) & 67.6 \std{13.4} & 93.1 \std{0.4} & 68.5 \std{12.3} & 91.8 \std{0.4} & 69.5 \std{11.1} & 92.2 \std{0.4} & 65.3 \std{12.9} & 93.5 \std{0.2} \\
DiffICL(N) & 91.4 \std{1.8} & 94.2 \std{0.5} & 91.3 \std{1.7} & 93.5 \std{0.8} & 90.6 \std{1.7} & 93.2 \std{0.5} & 92.2 \std{1.9} & 95.0 \std{0.8} \\
\bottomrule
\end{tabular}
}
\end{table*}
\begin{table*}[htbp]
\centering
\caption{Detailed performance comparison on the \textit{kr-vs-kp} dataset. 
\small{We report the downstream task performance of baselines alongside the DiffICL and its two ablation variants: DiffICL(S) and DiffICL(N). Performance is evaluated across four learners under two scenarios: utilizing synthetic data as a replacement for real data (\textbf{Syn.}), and utilizing a combination of real and synthetic data for data augmentation (\textbf{Aug.}). The best results  are highlighted in \textcolor[HTML]{D93F49}{red}.}}
\label{tab:kr_vs_kp_performance}
\setlength{\tabcolsep}{4pt} 
\resizebox{\textwidth}{!}{
\begin{tabular}{lcccccccc}
\toprule
\multirow{2.5}{*}{\textbf{Method}} & \multicolumn{2}{c}{\textbf{XGB}} & \multicolumn{2}{c}{\textbf{CatB}} & \multicolumn{2}{c}{\textbf{RF}} & \multicolumn{2}{c}{\textbf{MLP}} \\
\cmidrule(lr){2-3} \cmidrule(lr){4-5} \cmidrule(lr){6-7} \cmidrule(lr){8-9}
& \textbf{Syn.} & \textbf{Aug.} & \textbf{Syn.} & \textbf{Aug.} & \textbf{Syn.} & \textbf{Aug.} & \textbf{Syn.} & \textbf{Aug.} \\
\midrule
CTGAN      & 69.5 \std{5.5} & 99.6 \std{0.2} & 72.7 \std{4.3} & 99.5 \std{0.1} & 69.5 \std{4.6} & 99.3 \std{0.4} & 73.6 \std{3.4} & 98.9 \std{0.5} \\
CTABGAN+   & 85.8 \std{1.7} & 99.5 \std{0.2} & 87.4 \std{2.3} & 99.3 \std{0.2} & 85.0 \std{1.8} & 99.2 \std{0.2} & 88.1 \std{2.3} & 98.8 \std{0.3} \\
TVAE       & 98.2 \std{0.4} & 99.7 \std{0.1} & 98.5 \std{0.2} & 99.7 \std{0.1} & 97.4 \std{0.4} & 99.1 \std{0.2} & 98.2 \std{0.4} & 99.5 \std{0.1} \\
tabDDPM    & 99.0 \std{0.4} & 99.8 \std{0.1} & 99.1 \std{0.2} & 99.7 \std{0.1} & 98.3 \std{0.6} & 99.3 \std{0.2} & 98.7 \std{0.3} & 99.5 \std{0.1} \\
GReaT      & 99.6 \std{0.1} & 99.4 \std{0.2} & 99.6 \std{0.1} & 99.3 \std{0.2} & 99.2 \std{0.2} & 97.8 \std{0.5} & 99.4 \std{0.1} & 98.7 \std{0.4} \\
TabSYN     & 98.5 \std{0.5} & 99.7 \std{0.0} & 98.8 \std{0.3} & 99.6 \std{0.0} & 97.3 \std{0.6} & 99.1 \std{0.1} & 98.8 \std{0.3} & 99.5 \std{0.1} \\
TabDiff    & 99.3 \std{0.1} & 99.8 \std{0.1} & 99.3 \std{0.1} & 99.8 \std{0.1} & 98.8 \std{0.3} & 99.6 \std{0.1} & 99.1 \std{0.2} & 99.8 \std{0.1} \\
\midrule
DiffICL    & \textcolor[HTML]{D93F49}{99.7 \std{0.1}} & \textcolor[HTML]{D93F49}{99.9 \std{0.0}} & \textcolor[HTML]{D93F49}{99.7 \std{0.2}} & \textcolor[HTML]{D93F49}{99.9 \std{0.0}} & \textcolor[HTML]{D93F49}{99.5 \std{0.3}} & \textcolor[HTML]{D93F49}{99.9 \std{0.1}} & \textcolor[HTML]{D93F49}{99.6 \std{0.2}} & \textcolor[HTML]{D93F49}{99.9 \std{0.0}} \\
DiffICL(S) & 89.2 \std{1.7} & \textcolor[HTML]{D93F49}{99.9 \std{0.0}} & 90.2 \std{1.8} & 99.8 \std{0.1} & 87.8 \std{2.9} & 99.6 \std{0.1} & 89.5 \std{3.0} & 99.6 \std{0.1} \\
DiffICL(N) & 99.5 \std{0.3} & \textcolor[HTML]{D93F49}{99.9 \std{0.0}} & 99.4 \std{0.3} & \textcolor[HTML]{D93F49}{99.9 \std{0.0}} & 99.2 \std{0.4} & \textcolor[HTML]{D93F49}{99.9 \std{0.0}} & 98.9 \std{0.7} & \textcolor[HTML]{D93F49}{99.9 \std{0.0}} \\
\bottomrule
\end{tabular}
}
\end{table*}
\begin{table*}[htbp]
\centering
\caption{Detailed performance comparison on the \textit{ilpd} dataset.
\small{We report the downstream task performance of baselines alongside the DiffICL and its two ablation variants: DiffICL(S) and DiffICL(N). Performance is evaluated across four learners under two scenarios: utilizing synthetic data as a replacement for real data (\textbf{Syn.}), and utilizing a combination of real and synthetic data for data augmentation (\textbf{Aug.}). The best results  are highlighted in \textcolor[HTML]{D93F49}{red}.}}
\label{tab:ilpd_performance}
\setlength{\tabcolsep}{4pt} 
\resizebox{\textwidth}{!}{
\begin{tabular}{lcccccccc}
\toprule
\multirow{2.5}{*}{\textbf{Method}} & \multicolumn{2}{c}{\textbf{XGB}} & \multicolumn{2}{c}{\textbf{CatB}} & \multicolumn{2}{c}{\textbf{RF}} & \multicolumn{2}{c}{\textbf{MLP}} \\
\cmidrule(lr){2-3} \cmidrule(lr){4-5} \cmidrule(lr){6-7} \cmidrule(lr){8-9}
& \textbf{Syn.} & \textbf{Aug.} & \textbf{Syn.} & \textbf{Aug.} & \textbf{Syn.} & \textbf{Aug.} & \textbf{Syn.} & \textbf{Aug.} \\
\midrule
CTGAN      & 67.1 \std{3.6} & 70.6 \std{3.1} & 68.5 \std{2.3} & 69.0 \std{3.1} & 69.4 \std{3.1} & 71.9 \std{2.3} & 69.1 \std{2.2} & 67.3 \std{2.8} \\
CTABGAN+   & 69.0 \std{3.3} & 71.6 \std{1.2} & 70.0 \std{2.2} & 70.4 \std{2.3} & 71.3 \std{1.9} & 72.7 \std{1.0} & 70.7 \std{1.2} & 70.7 \std{1.7} \\
TVAE       & 70.7 \std{0.9} & 73.1 \std{1.2} & 69.7 \std{1.0} & 72.2 \std{1.4} & 71.5 \std{0.9} & 74.4 \std{1.5} & \textcolor[HTML]{D93F49}{72.7 \std{1.0}} & \textcolor[HTML]{D93F49}{72.7 \std{2.2}} \\
tabDDPM    & 70.5 \std{2.3} & 74.2 \std{2.1} & 71.7 \std{0.8} & 73.9 \std{1.6} & 69.0 \std{3.1} & 74.6 \std{1.5} & 55.6 \std{6.9} & 64.1 \std{3.2} \\
GReaT      & 56.8 \std{7.1} & 63.9 \std{4.8} & 57.2 \std{8.4} & 64.3 \std{5.0} & 55.4 \std{8.4} & 64.3 \std{4.6} & 49.9 \std{12.8} & 57.3 \std{10.4} \\
TabSYN     & \textcolor[HTML]{D93F49}{72.2 \std{2.0}} & 74.2 \std{1.0} & \textcolor[HTML]{D93F49}{72.2 \std{2.0}} & 73.2 \std{1.5} & \textcolor[HTML]{D93F49}{73.1 \std{1.8}} & \textcolor[HTML]{D93F49}{75.3 \std{1.2}} & 69.8 \std{1.8} & 70.8 \std{1.3} \\
TabDiff    & 71.4 \std{1.4} & 72.7 \std{1.8} & 71.9 \std{1.6} & 72.2 \std{1.4} & 72.5 \std{0.8} & 74.0 \std{1.6} & 69.9 \std{2.0} & 71.9 \std{1.4} \\
\midrule
DiffICL    & 69.4 \std{4.9} & 74.8 \std{1.8} & 71.0 \std{3.9} & 72.8 \std{1.7} & 68.2 \std{5.9} & 74.4 \std{2.5} & 69.2 \std{7.8} & 70.1 \std{7.6} \\
DiffICL(S) & 61.3 \std{6.5} & \textcolor[HTML]{D93F49}{75.9 \std{0.3}} & 60.9 \std{4.2} & \textcolor[HTML]{D93F49}{74.3 \std{0.4}} & 63.4 \std{3.7} & 74.9 \std{0.2} & 58.0 \std{4.9} & 66.8 \std{0.6} \\
DiffICL(N) & 71.5 \std{5.5} & 74.6 \std{2.3} & 71.1 \std{5.4} & 73.5 \std{2.7} & 70.8 \std{6.0} & 74.6 \std{2.2} & 71.2 \std{9.9} & 70.9 \std{9.4} \\
\bottomrule
\end{tabular}
}
\end{table*}
\begin{table*}[htbp]
\centering
\caption{Detailed performance comparison on the \textit{wdbc} dataset.
\small{We report the downstream task performance of baselines alongside the DiffICL and its two ablation variants: \textbf{DiffICL(S)} and \textbf{DiffICL(N)}. Performance is evaluated across four learners under two scenarios: utilizing synthetic data as a replacement for real data (\textbf{Syn.}), and utilizing a combination of real and synthetic data for data augmentation (\textbf{Aug.}). The best results  are highlighted in \textcolor[HTML]{D93F49}{red}.}}
\label{tab:wdbc_performance}
\setlength{\tabcolsep}{4pt} 
\resizebox{\textwidth}{!}{
\begin{tabular}{lcccccccc}
\toprule
\multirow{2.5}{*}{\textbf{Method}} & \multicolumn{2}{c}{\textbf{XGB}} & \multicolumn{2}{c}{\textbf{CatB}} & \multicolumn{2}{c}{\textbf{RF}} & \multicolumn{2}{c}{\textbf{MLP}} \\
\cmidrule(lr){2-3} \cmidrule(lr){4-5} \cmidrule(lr){6-7} \cmidrule(lr){8-9}
& \textbf{Syn.} & \textbf{Aug.} & \textbf{Syn.} & \textbf{Aug.} & \textbf{Syn.} & \textbf{Aug.} & \textbf{Syn.} & \textbf{Aug.} \\
\midrule
CTGAN      & 97.5 \std{0.4} & 98.8 \std{0.2} & 97.7 \std{0.4} & 98.7 \std{0.2} & 97.5 \std{0.3} & 98.3 \std{0.2} & 98.4 \std{0.7} & 99.0 \std{0.2} \\
CTABGAN+   & 98.8 \std{0.2} & 98.8 \std{0.4} & 98.8 \std{0.2} & 98.9 \std{0.4} & 98.3 \std{0.2} & 98.5 \std{0.4} & 98.8 \std{0.3} & 99.0 \std{0.5} \\
TVAE       & 98.7 \std{0.3} & 98.9 \std{0.2} & 98.7 \std{0.3} & 98.8 \std{0.2} & 98.5 \std{0.3} & 98.6 \std{0.1} & 99.1 \std{0.1} & \textcolor[HTML]{D93F49}{99.3 \std{0.2}} \\
tabDDPM    & 98.0 \std{0.5} & 99.1 \std{0.1} & 98.3 \std{0.3} & 99.1 \std{0.1} & 97.5 \std{0.9} & 98.7 \std{0.2} & 98.0 \std{1.1} & 99.0 \std{0.2} \\
TabSYN     & \textcolor[HTML]{D93F49}{99.1 \std{0.1}} & \textcolor[HTML]{D93F49}{99.2 \std{0.1}} & \textcolor[HTML]{D93F49}{99.2 \std{0.1}} & \textcolor[HTML]{D93F49}{99.2 \std{0.1}} & \textcolor[HTML]{D93F49}{98.9 \std{0.2}} & \textcolor[HTML]{D93F49}{98.9 \std{0.1}} & \textcolor[HTML]{D93F49}{99.2 \std{0.1}} & \textcolor[HTML]{D93F49}{99.3 \std{0.0}} \\
TabDiff    & 99.0 \std{0.1} & \textcolor[HTML]{D93F49}{99.2 \std{0.2}} & 99.0 \std{0.1} & 99.1 \std{0.1} & 98.4 \std{0.3} & 98.5 \std{0.2} & \textcolor[HTML]{D93F49}{99.2 \std{0.1}} & 99.2 \std{0.2} \\
\midrule
DiffICL    & 98.8 \std{0.2} & 99.1 \std{0.2} & 98.7 \std{0.2} & 99.0 \std{0.1} & 98.4 \std{0.1} & 98.5 \std{0.3} & 99.1 \std{0.3} & \textcolor[HTML]{D93F49}{99.3 \std{0.2}} \\
DiffICL(S) & 98.0 \std{0.4} & 98.8 \std{0.2} & 98.1 \std{0.4} & 98.9 \std{0.1} & 98.3 \std{0.2} & 98.5 \std{0.2} & 98.3 \std{0.8} & 99.0 \std{0.2} \\
DiffICL(N) & 98.6 \std{0.4} & 99.0 \std{0.3} & 98.6 \std{0.4} & 98.9 \std{0.3} & 98.5 \std{0.4} & 98.7 \std{0.3} & 99.1 \std{0.2} & \textcolor[HTML]{D93F49}{99.3 \std{0.1}} \\
\bottomrule
\end{tabular}
}
\end{table*}
\begin{table*}[htbp]
\centering
\caption{Detailed performance comparison on the \textit{kc2} dataset.
\small{We report the downstream task performance of baselines alongside the DiffICL and its two ablation variants: DiffICL(S) and DiffICL(N). Performance is evaluated across four learners under two scenarios: utilizing synthetic data as a replacement for real data (\textbf{Syn.}), and utilizing a combination of real and synthetic data for data augmentation (\textbf{Aug.}). The best results  are highlighted in \textcolor[HTML]{D93F49}{red}.}}
\label{tab:kc2_performance}
\setlength{\tabcolsep}{4pt} 
\resizebox{\textwidth}{!}{
\begin{tabular}{lcccccccc}
\toprule
\multirow{2.5}{*}{\textbf{Method}} & \multicolumn{2}{c}{\textbf{XGB}} & \multicolumn{2}{c}{\textbf{CatB}} & \multicolumn{2}{c}{\textbf{RF}} & \multicolumn{2}{c}{\textbf{MLP}} \\
\cmidrule(lr){2-3} \cmidrule(lr){4-5} \cmidrule(lr){6-7} \cmidrule(lr){8-9}
& \textbf{Syn.} & \textbf{Aug.} & \textbf{Syn.} & \textbf{Aug.} & \textbf{Syn.} & \textbf{Aug.} & \textbf{Syn.} & \textbf{Aug.} \\
\midrule
CTGAN      & 75.6 \std{4.8} & 75.6 \std{1.9} & 77.2 \std{4.0} & 79.0 \std{1.5} & 80.5 \std{1.6} & 81.0 \std{0.4} & 68.9 \std{1.7} & 74.6 \std{4.1} \\
CTABGAN+   & 68.0 \std{6.1} & 76.8 \std{2.3} & 77.0 \std{5.3} & 79.4 \std{1.8} & 64.2 \std{6.9} & 81.6 \std{0.7} & 80.4 \std{1.3} & 80.2 \std{1.5} \\
TVAE       & 76.2 \std{5.0} & 77.5 \std{1.8} & 77.3 \std{5.4} & 81.1 \std{1.8} & 79.4 \std{2.7} & 81.8 \std{1.0} & 80.1 \std{1.0} & 81.5 \std{1.1} \\
tabDDPM    & 75.2 \std{3.2} & 75.0 \std{1.7} & 77.4 \std{2.4} & 76.4 \std{0.9} & 81.1 \std{1.5} & 80.4 \std{1.1} & 82.1 \std{0.8} & 82.4 \std{0.8} \\
TabSYN     & 78.9 \std{2.7} & 74.9 \std{1.9} & 78.5 \std{2.4} & 75.9 \std{2.3} & 80.6 \std{1.3} & 79.9 \std{1.3} & 69.5 \std{4.8} & 68.6 \std{2.6} \\
TabDiff    & 80.9 \std{0.4} & \textcolor[HTML]{D93F49}{78.8 \std{1.3}} & 82.0 \std{1.2} & \textcolor[HTML]{D93F49}{81.7 \std{0.9}} & \textcolor[HTML]{D93F49}{82.5 \std{0.7}} & 81.0 \std{0.2} & \textcolor[HTML]{D93F49}{83.6 \std{0.3}} & \textcolor[HTML]{D93F49}{83.3 \std{0.2}} \\
\midrule
DiffICL    & \textcolor[HTML]{D93F49}{81.0 \std{3.2}} & 71.7 \std{3.0} & \textcolor[HTML]{D93F49}{82.4 \std{1.1}} & 82.2 \std{2.3} & 79.9 \std{2.3} & 80.2 \std{1.1} & 82.2 \std{2.4} & 76.6 \std{2.5} \\
DiffICL(S) & 78.7 \std{1.4} & 78.1 \std{1.6} & 80.4 \std{4.2} & 79.9 \std{1.7} & 79.4 \std{3.4} & \textcolor[HTML]{D93F49}{82.9 \std{0.4}} & 79.8 \std{6.4} & 80.4 \std{2.8} \\
DiffICL(N) & 79.6 \std{2.3} & 74.0 \std{0.9} & 80.6 \std{1.3} & 73.9 \std{0.8} & 77.8 \std{3.5} & 80.8 \std{0.8} & 83.1 \std{0.5} & 80.0 \std{2.4} \\
\bottomrule
\end{tabular}
}
\end{table*}
\section{Limitations}
\label{sec:limitation}
A current limitation of DiffICL is that its inference stage requires training decoders for each new dataset. 
We split the available data into a context set and a query set: the context set is used to condition latent generation, while the query set is used to train feature-wise decoders that map generated latent representations back to raw tabular values. This design improves reconstruction accuracy, but it also introduces an additional data requirement, since part of the available data must be reserved for decoder training.

A more data-efficient alternative would be to learn a universal decoder during pretraining, so that the same decoder could map query embeddings to raw feature values across different datasets and columns. 
We explored this direction, but found that a simple shared MLP decoder, which maps an arbitrary 192-dimensional feature embedding to a scalar value, leads to noticeable reconstruction errors across heterogeneous datasets and feature types. This suggests that universal decoding for tabular foundation models is non-trivial, because raw feature spaces differ substantially across datasets.

Therefore, we adopt dataset-specific lightweight decoders in the current implementation. Although this choice sacrifices a small amount of sample efficiency, we find empirically that the decoder can be trained accurately with as few as several dozen samples. This allows DiffICL to achieve more reliable feature reconstruction while keeping the inference procedure practical.

\section{Additional Ablation Studies}
\label{app:ablation}
\subsection{The Impact of Context Ratio}
\begin{figure}[htbp]
    \centering
    \begin{subfigure}{0.32\textwidth}
        \centering
        \includegraphics[width=\linewidth]{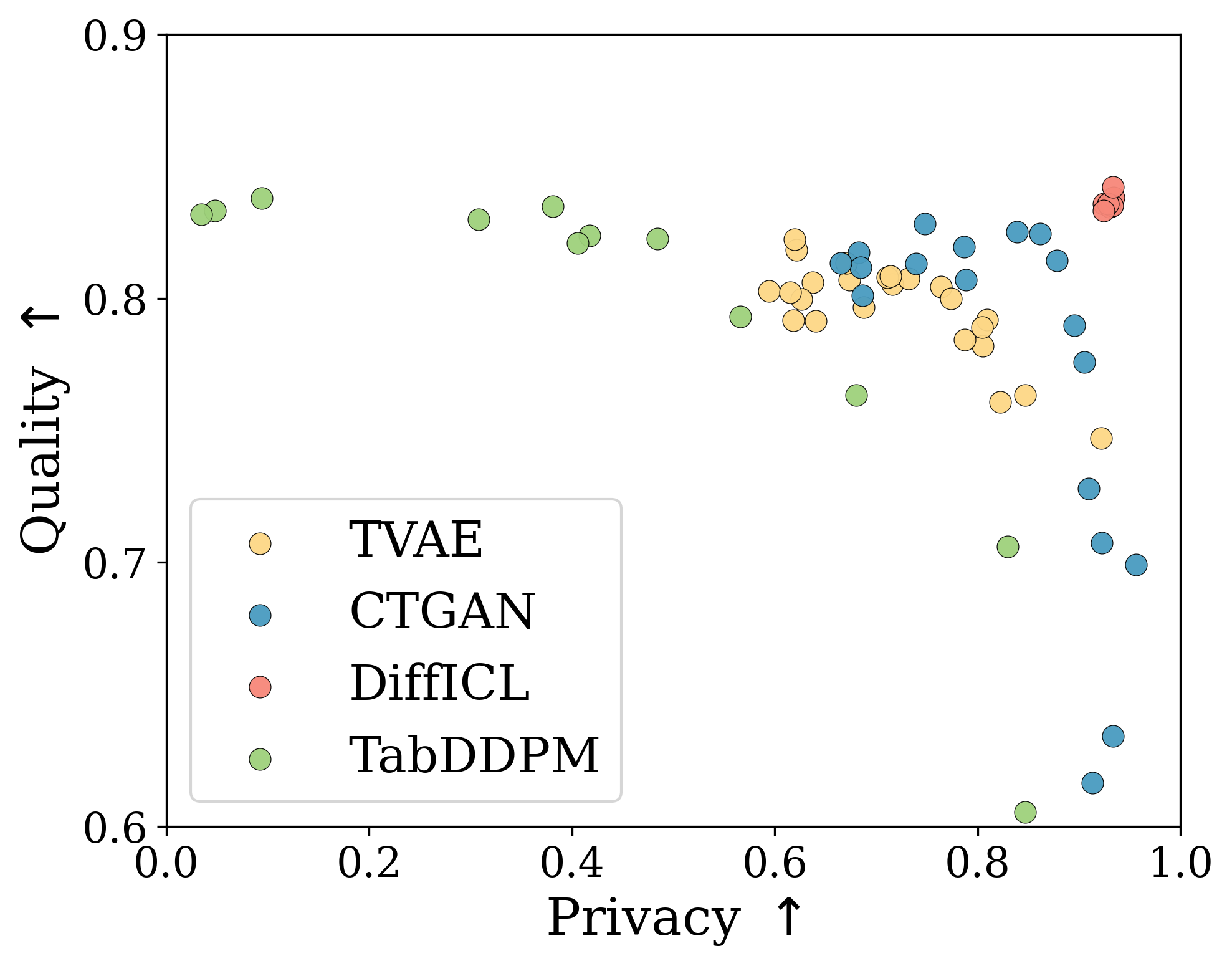}
        
    \end{subfigure}
    \hfill
    \begin{subfigure}{0.32\textwidth}
        \centering
        \includegraphics[width=\linewidth]{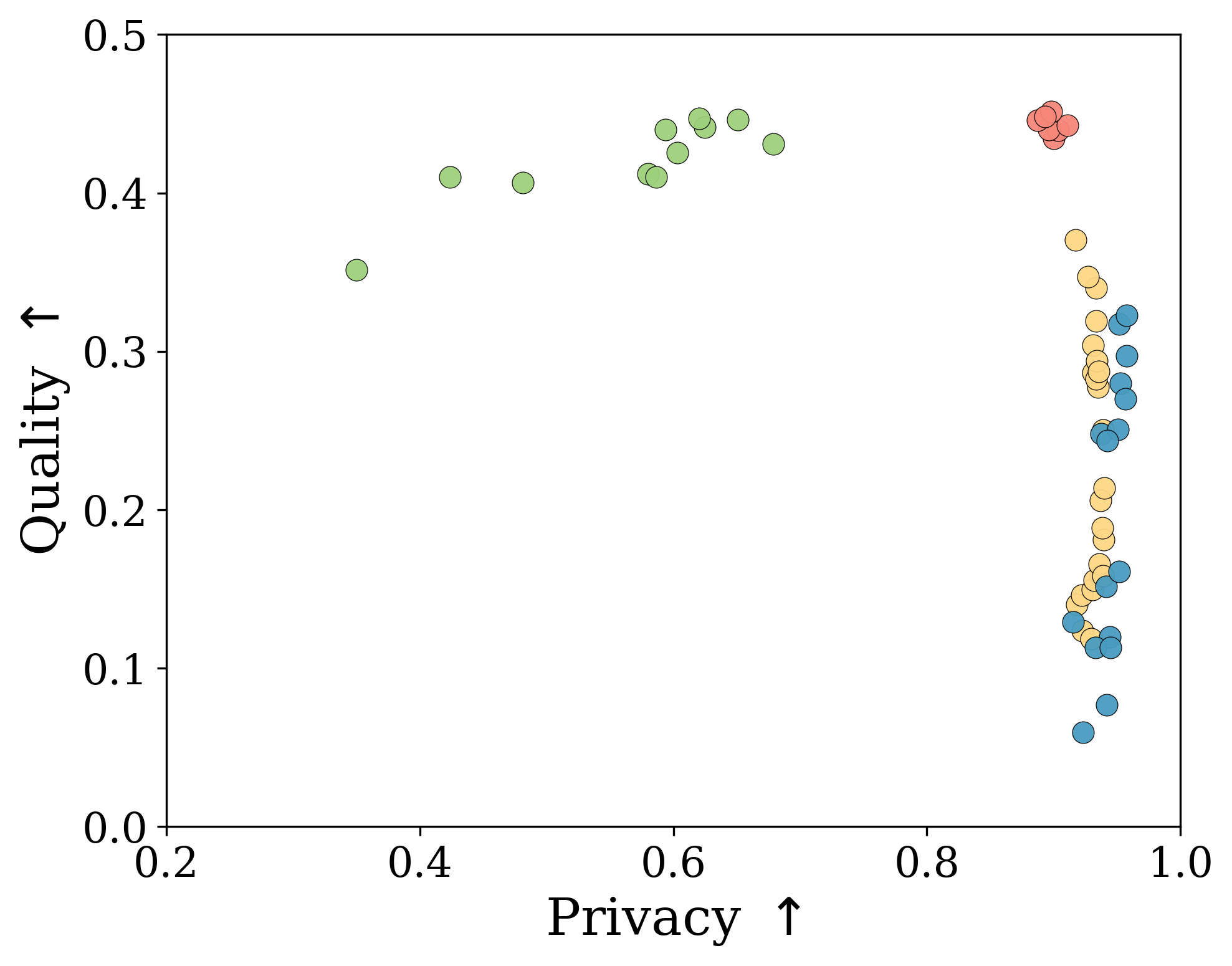}
        
    \end{subfigure}
    \hfill
    \begin{subfigure}{0.32\textwidth}
        \centering
        \includegraphics[width=\linewidth]{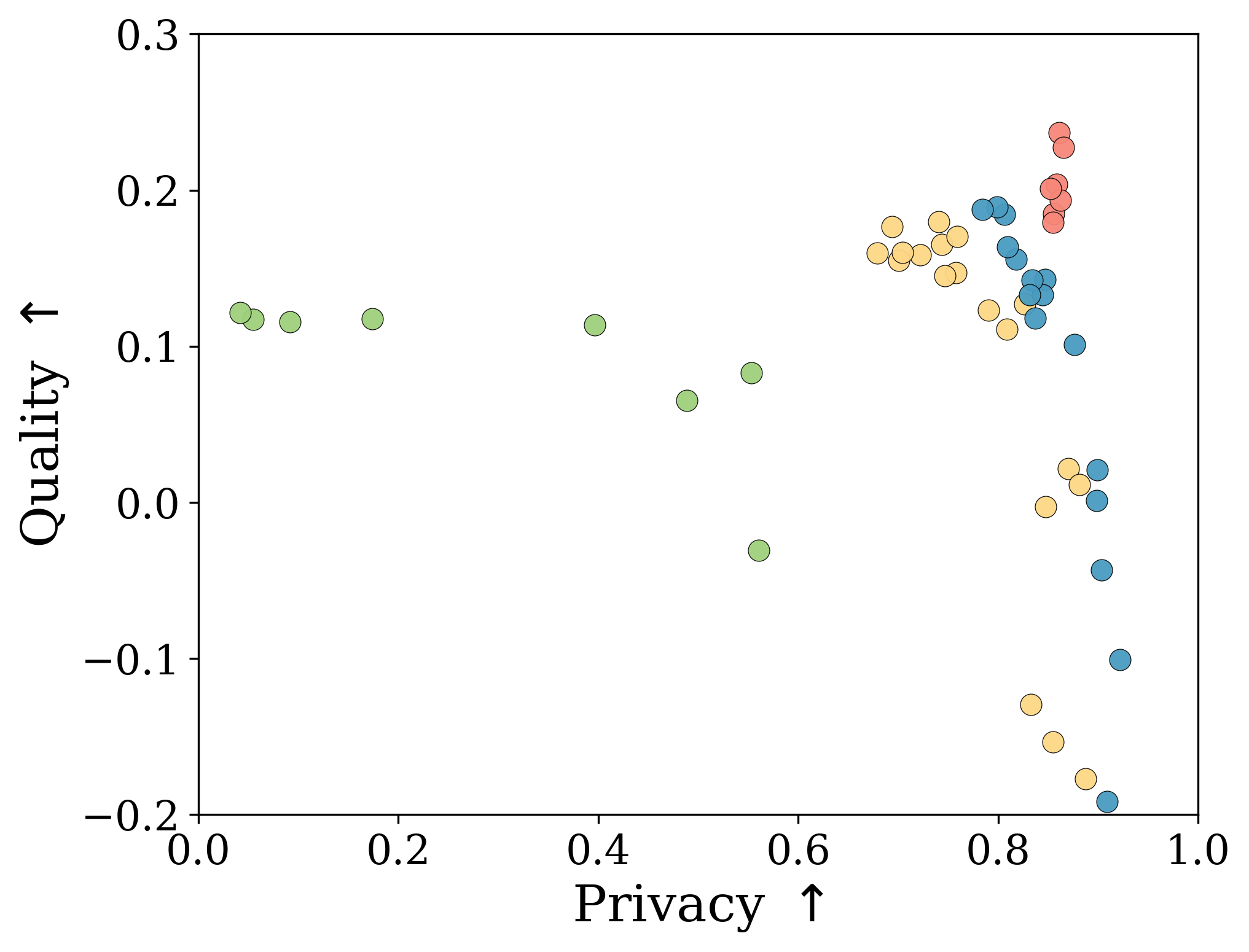}
       
    \end{subfigure}
    \caption{Quality--privacy tradeoffs under different training configurations.\small{Each point corresponds to a model checkpoint under different training configurations. For tabDDPM, TVAE, and CTGAN, the points are obtained by varying the number of training steps from 500 to 30000. For DiffICL, the points are obtained by varying the context ratio from 0.2 to 0.8. The three subplots correspond to the Adult, Abalone, and Health Insurance datasets, respectively, all with $N=200$ training samples.}}
    \label{fig:context_ratio}
\end{figure}

Context ratio is the ratio we split a new dataset $D$ into $D_\text{ctx}$ and $D_\text{qry}$. In all experiments in Section~\ref{sec:exp}, we set the context ratio of DiffICL to 0.3. We further study the sensitivity of DiffICL to this hyperparameter. 
As shown in Figure~\ref{fig:context_ratio}, we evaluate DiffICL on each dataset with context ratios ranging from 0.2 to 0.8. Each context ratio is a red point. 
The results show that DiffICL is robust to the choice of context ratio: the performance variation caused by this hyperparameter is small compared with the gap between DiffICL and the baseline methods.

\subsection{Effect of the number of training samples.}
We further study how different methods behave when the target dataset contains fewer training samples. For each dataset, we randomly subsample the original training set without replacement using ratios in $\{0.2, 0.4, 0.6, 0.8, 1.0\}$. Each generative model is then trained on the subsampled training set and used to generate 2,500 synthetic samples. 
The ratio $1.0$ corresponds to the full training set used in the main experiments.

\begin{figure}[htbp]
    \centering
    \begin{subfigure}{0.99\textwidth}
        \centering
        \includegraphics[width=\linewidth]{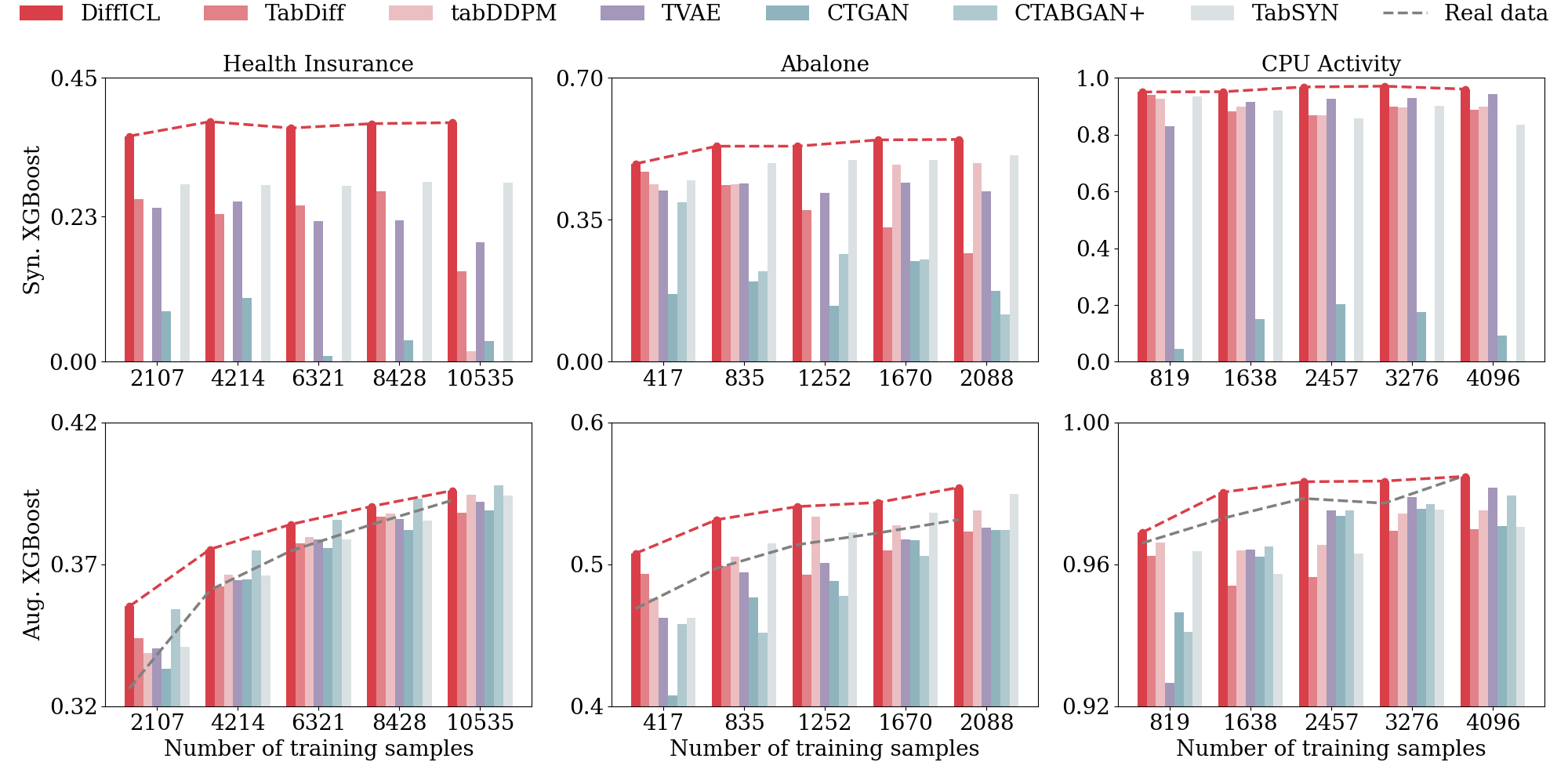}
    \end{subfigure}
    
    \caption{
Effect of the number of training samples on synthetic-data quality and data augmentation performance.
\small{\textbf{Top}: XGBoost is trained  on synthetic data generated by each method and evaluated on the held-out test set.
\textbf{Bottom}: XGBoost is trained on the mixture of real training data and synthetic data; the gray dashed curve reports the performance of XGBoost trained only on real data, and the red dashed curve highlights the augmentation performance of DiffICL.}
}
    \label{fig:num_training_samples}
\end{figure}

The first row of Figure~\ref{fig:num_training_samples} reports the performance of an XGBoost model trained on synthetic data and evaluated on the held-out test set. 
DiffICL remains effective even when only a small fraction (e.g. 0.2) of the training data is available. 
For example, on Abalone, DiffICL achieves an $R^2$ of 0.49 using only 417 training samples. 
On Healthcare, with 2,107 training samples, DiffICL reaches an $R^2$ of 0.358, outperforming the second-best method, TabDiff, which obtains 0.257. 
These results suggest that DiffICL is less sensitive to limited target-dataset size, consistent with its use of transferable priors learned during cross-dataset pretraining.

The second row of Figure~\ref{fig:num_training_samples} shows the effect of using synthetic data for augmentation, where the learner is trained on the mixture of real and synthetic samples. 
We find that adding low-quality synthetic data often does not severely degrade the learner, even when the synthetic-only performance is poor or even negative in terms of $R^2$. 
In contrast, adding high-quality samples generated by DiffICL improves downstream prediction: the red dashed curve, which represents XGBoost trained with real and DiffICL-generated data, lies above the gray dashed curve, which represents XGBoost trained only on real data.

\end{document}